\documentclass{article}

\usepackage[preprint]{neurips_2026}

\makeatletter
\renewcommand{\@notice}{}
\makeatother

\usepackage[utf8]{inputenc}
\usepackage[T1]{fontenc}
\usepackage{hyperref}
\usepackage{url}
\usepackage{booktabs}
\usepackage{amsfonts}
\usepackage{amsmath}
\usepackage{nicefrac}
\usepackage{microtype}
\usepackage{xcolor}

\usepackage{graphicx}
\usepackage{adjustbox}
\usepackage{makecell}
\usepackage{caption,subcaption}

\newcommand{\onescore}[1]{\makecell[c]{#1}}
\newcommand{\compacttablesetup}{%
  \setlength{\tabcolsep}{3.5pt}%
  \renewcommand{\arraystretch}{1.05}%
}

\usepackage{multirow}
\usepackage{tabularx}
\usepackage{placeins}
\usepackage{float}
\usepackage{titletoc}
\usepackage{enumitem}
\usepackage{wrapfig}
\usepackage{amsthm}

\usepackage{flafter}
\usepackage{cleveref}
\usepackage{pgffor}

\newcommand{\TODO}[1]{{}}

\title{PairIT: Autoregressive Transformers for Low-Data Molecule Optimization}

\author{%
Hannah Lawrence$^{1,2}$ \And
Bodhi P. Vani$^{1}$ \And
Ji Won Park$^{1}$ \And
Natasa Tagasovska$^{3}$ \And
Samuel Don Stanton$^{4}$ \And
Andrew Martin Watkins$^{1}$ \And
Michael Maser$^{1}$ \And
Ewa M. Nowara$^{1,*}$ \\[0.5em]
$^{1}$Genentech, South San Francisco, CA, USA \\
$^{2}$Massachusetts Institute of Technology, Cambridge, MA, USA \\
$^{3}$Prescient Design, Genentech, Basel, Switzerland \\
$^{4}$Prescient Design, Genentech, New York, NY, USA \\
$^*$Corresponding author: \texttt{nowara.ewa@gene.com}
}
\begin{document}

\maketitle
\begingroup
\renewcommand{\thefootnote}{}
\footnotetext{Generative AI tools were used to assist with manuscript editing and formatting, and with code development and debugging. All AI-assisted outputs were reviewed and verified by the authors.

\smallskip
Code is available at \url{https://github.com/prescient-design/pairit}.}
\addtocounter{footnote}{-1}
\endgroup

\begin{abstract}
  Generating novel molecules with desired properties is a central challenge in drug discovery. Although foundational work focused on \textit{de novo} generation, practical use cases require a more complex setting: conditioning on an initial seed molecule, optimizing multiple properties, and operating in data-sparse regimes. We introduce \textit{PairIT}, a flexible training framework for autoregressive transformers over SMILES sequences that enables \textit{seeded} generation by learning from pairs of structurally similar molecules. By leveraging matched pairs, the model learns to generate local, chemically meaningful edits that improve a property of the seed molecule. Prompt tokens extend this framework by specifying the target property, desired direction, and approximate magnitude of change. Multiple prompt tokens can be concatenated to enable multi-property optimization at inference time. For properties where pair supervision alone is insufficient, reinforcement learning post-training can further improve optimization, providing a tunable trade-off between proximity to the seed and the optimization objective. PairIT naturally supports pretraining on a large dataset and fine-tuning on a smaller dataset, which is critical for performance in data-sparse regimes. Surprisingly, we observe diminishing returns to the scale of the pretraining dataset. We demonstrate empirically on MoleculeNet and Therapeutics Data Commons datasets that, with appropriate paired-data curation, a standard transformer model of modest size (68M parameters) can achieve robust seeded molecular optimization in low-data regimes.
  \end{abstract}

\suppressfloats[t]

\begin{figure}[t]
    \centering
    \includegraphics[width=0.85\textwidth]{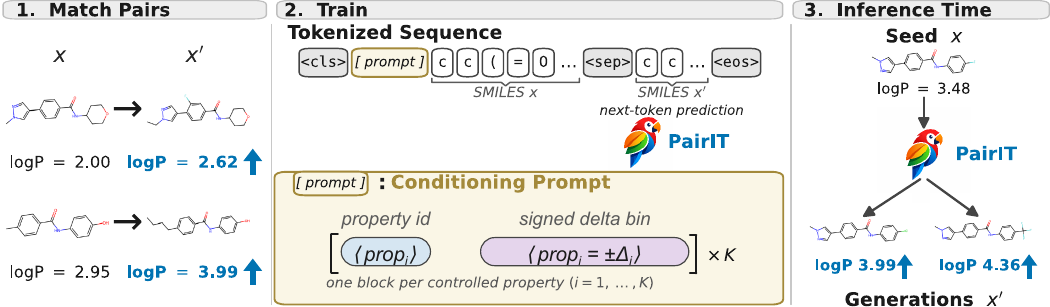}
    \caption{\small Overview of PairIT. We (1) construct training pairs of structurally similar molecules with differing property values and (2) train an autoregressive transformer to predict the target SMILES given the seed SMILES and structured property-control prompt tokens. (3) At inference time, the model generates property-improved candidates from a new seed and prompt. Multiple prompt blocks can be concatenated for multi-property control.}
    \label{fig:overview}
\end{figure}

\section{Introduction}
\label{sec:intro}

Generative models enable efficient exploration of chemical space, making them increasingly important tools for molecular discovery~\citep{bohacek1996art}. In practical drug discovery, an important task is \emph{molecular optimization}. The goal is to propose small, local modifications to a known seed molecule that improve one or more (bio)chemical properties. Maintaining structural similarity to a seed molecule increases the likelihood of preserving biological function~\citep{Bender2004MolecularSA}. However, structure--function response surfaces are notoriously jagged ~\citep{aldeghi2022roughness}; molecules differing by even a single atom may have drastically different properties, making optimization challenging. In practice, computational and medicinal chemists often approach this problem through local molecular edits informed by cheminformatics tools and domain expertise~\citep{kramer2014matched}. A full discovery campaign has an ``outer loop'' in which candidates are synthesized and experimentally measured and an ``inner loop'' in which candidates are proposed and scored against a surrogate model. We focus on seeded generation\footnote{In this work, we use the terms \emph{molecular optimization} and \emph{seeded generation} interchangeably, and likewise for \emph{seed molecule} and \emph{input molecule}.} in this ``inner loop'', in the scarce-label regime common in early-stage discovery, where experimental measurements are expensive and datasets often contain only hundreds to thousands of labeled molecules~\citep{wu2018moleculenet}.

While substantial prior literature exists on molecular optimization, the small-dataset regime remains under-addressed, with few works evaluating seeded optimization on biological properties. Many recent models are developed and evaluated either in \emph{de novo} settings, where there is no seed molecule~\citep{pinheiro20233d,joshi2025all,xu2023geometric}, or with far more labeled data than is typically available in early-stage discovery~\citep{dey2025large,lee2025latent,jin2020hierarchical}. In contrast, we focus on seeded optimization in the low-data regime. Generated molecules must remain close enough to the seed to preserve relevant structure while introducing enough novelty to improve the desired property. Moreover, constraining generation to a local search space is restrictive, by definition: the model must introduce enough novelty to improve the requested properties while still remaining close to the seed, and such conditioning can collapse toward near-reconstruction rather than diverse one-to-many outputs \citep{kumar2024conditional}. In small labeled datasets, limited examples further increase the risk of overfitting and memorized edits \citep{wu2018moleculenet}. Yet even small datasets can contain useful local structure--property relationships, motivating training objectives that expose these transformations directly.

One way to expose such local transformations directly is to match structurally similar molecules from a dataset and train the model to ``translate'' from one molecule to the other. If the pairs are directed so that the target molecule has an improved value for a property of interest, the model learns local, property-improving edits. For example, \citet{damani2019black} apply this sequence-to-sequence approach to small-molecule optimization, while \citet{tagasovska2024implicitly, pinheiroimplicit, chen2024generalists} use related pair-based approaches for protein sequence optimization. Recent work also applies pair-based training to large language models for controllable molecular optimization~\citep{dey2025large}. 

We propose PairIT (Figure~\ref{fig:overview}), a decoder-only autoregressive transformer over SMILES\footnote{SMILES strings are text representations of molecular structure~\citep{weininger1988smiles}.} trained on matched seed--target pairs of structurally similar molecules with differing properties. The base model learns to generate local, property-improving edits directly from pair supervision. To extend optimization beyond the property changes learned in a single generation step, we also consider iterative re-seeding, in which high-scoring generations are fed back to the model as new seeds for further optimization. We refer to this variant as PairIT-Beam. For properties that remain difficult to optimize from pair supervision alone, we further explore reinforcement learning (RL) post-training with a property-model reward when a reliable property predictor is available, which we refer to as PairIT-RL.

However, pair-only approaches tie the optimization objective to the constructed pair set. Changing the target property or direction, or composing multiple objectives, can require constructing new pairs and retraining. This becomes increasingly restrictive in small datasets, where suitable pairs for specific combinations of property changes may be sparse or absent. To provide finer control over how much a seed property is changed, we additionally introduce structured control tokens specifying the property, direction, and approximate magnitude of change. This also extends naturally to
multi-property optimization, where constructing similar pairs that improve several properties simultaneously can be especially difficult in very small
datasets. Multiple property prompts can instead be composed at inference time without constructing a separate pair set and retraining for each combination.

Because the downstream pair sets are very small, we first pretrain PairIT on ZINC-250K~\citep{irwin2012zinc} before fine-tuning on each target dataset.
This raises two practical questions, whether the gains come simply from molecular pretraining or specifically from pretraining on directed pairs, and
how much paired pretraining is needed in terms of pair diversity and training duration. We therefore compare no pretraining, single-molecule pretraining, and directed-pair pretraining, vary the amount and diversity of paired data, and track validation and downstream performance over the course of training. We find that directed-pair pretraining outperforms pretraining on individual molecules, but that its benefits emerge surprisingly early, with the best validation checkpoint reached before one full epoch over the pair dataset. We study this behavior with training-dynamics and data-scale ablations, finding that repeated molecular coverage across many pairs helps explain the early saturation and that additional pair data yield diminishing returns. 

PairIT deliberately combines existing components in a way tailored to low-data seeded molecular optimization, rather than introducing a new
architecture or dataset. Rather than relying on a large general-purpose language model, we use a compact 68M-parameter transformer with molecule-specific tokenization, matched-pair construction, and the training and inference strategies described above. We evaluate this setting on MoleculeNet, where, to our knowledge, seeded generation has not previously been studied systematically. We formulate the task, define metrics that capture property improvement, novelty, and fidelity to the seed, and compare to diverse baseline methods under the same limited-data setting.

Our main contributions are (i) a compact 68M-parameter matched-pair autoregressive model for low-data seeded molecular optimization that is
competitive with substantially larger or more elaborate baselines; (ii) iterative re-seeding and optional RL post-training that can further improve property optimization when additional inference compute or a reliable property predictor is available; (iii) structured property-control tokens that provide control over the direction and magnitude of property change and support multi-property optimization without retraining for each objective combination; and (iv) an analysis of paired-pretraining dynamics showing that pair structure provides benefits beyond exposure to the same molecules individually, and characterizing why useful pretraining can emerge before a full pass over the pair dataset.

\section{Related Work}
\label{sec:related_work}

The methods most closely related to our approach are matched-pair methods for seeded molecular optimization. PropEn~\citep{tagasovska2024implicitly} is a non-autoregressive encoder\footnote{While \citet{tagasovska2024implicitly} describe their architecture as encoder-decoder, they do not use an autoregressive decoder, which is the more common usage of the term today.} trained to predict the output sequence in a pair; the authors show that this objective approximates the property-gradient direction. Follow-up work extends the framework to conditional density estimation~\citep{pinheiroimplicit} and multi-objective optimization via a non-domination ranker applied during pair construction~\citep{tagasovska2025property,park2024botied}. Earlier
small-molecule work frames optimization as graph-to-graph or sequence-to-sequence translation~\citep{jinlearning,jin2020hierarchical, kimdiscrete}; \citet{damani2019black} introduce recursive inference, in which generated molecules are repeatedly fed back as inputs.

GeLLM4O-C~\citep{dey2025large} is the closest language-model baseline. It instruction-tunes 7--8B-parameter LLMs on approximately 256K molecule pairs with natural-language instructions specifying which properties to improve or maintain. PairIT similarly makes the requested property change explicit at inference time, but uses structured control tokens in a compact 68M-parameter autoregressive SMILES transformer rather than a general-purpose natural-language LLM, with only 88--662 downstream fine-tuning pairs per MoleculeNet task. Other molecular language-model work includes DrugAssist~\citep{ye2023drugassist} and SmileyLlama~\citep{cavanagh2024smileylama}. 

RL provides another route for molecular optimization. REINVENT~\citep{olivecrona2017molecular}, LLOME~\citep{chen2024generalists},
and MOLRL~\citep{haddad2025targeted} apply reinforcement learning to molecular or biological sequence generation under property-based rewards. Preference optimization, including DPO~\citep{rafailov2024direct}, provides a related pairwise-learning approach, with multi-objective and molecular extensions~\citep{xiong2025projection, zhou2024modpo, yang2024rewards, gupta2025robust, park2023preference, calanzone2025molmoe}. Appendix~\ref{app:theory} compares the training signal of PairIT's conditional likelihood objective with DPO as pair differences increase.

Property-model-guided search provides a complementary approach, including latent-space Bayesian optimization~\citep{gomez2018automatic,lee2025latent} and gradient-based steering~\citep{lee2023exploring}. These methods could in
principle be combined with PairIT, while vanilla PairIT learns the optimization objective from paired data and does not require a property oracle in the generation loop.

\section{Proposed Method: PairIT}
\label{sec:method}

Let $\mathcal{D} = \{(x_i, y_i)\}_{i=1}^n$ be a dataset of molecules, where each $x_i$ is a molecular representation (e.g., SMILES string) and $y_i \in \mathcal{Y}$ is its property, which may take discrete, continuous, or mixed values. We refer to the unknown property function as $g: \mathcal{X} \to \mathcal{Y}$ such that $y_i = g(x_i)$.
Our goal is \emph{seeded molecular generation}: given an \emph{initial} molecule $x$, we seek to generate one or more \emph{candidate} molecules $x'$ such that $s(x', x) \ge \tau$ and $g(x') > g(x)$,
where $s(\cdot,\cdot)$ is a molecular similarity measure (e.g., Tanimoto similarity) and $\tau$ is the chosen similarity threshold. This formulation reflects a realistic task in drug discovery where the aim is to propose local, structure-preserving modifications that improve a property of interest while remaining in the chemical neighborhood of the input molecule. During training, we instantiate this objective with directed matched seed-target pairs; the corresponding dataset-specific pair-construction procedures are described in Section~\ref{sec:experiments}. 

\subsection{PairIT objective and interpretation}
\label{subsec:matched_pair_training}

Given a seed molecule $x$ and optional conditioning variables $c$, PairIT models a conditional distribution $q_\theta(x' \mid x, c)$ over target molecules $x'$, represented as SMILES strings. Let $p_{\mathrm{pair}}(x,x',c)$ denote the empirical distribution over the directed training pairs introduced above. PairIT parameterizes an autoregressive distribution over the target SMILES sequence
$x' = (x'_1, \dots, x'_T)$, where $T$ is the target length in tokens: $q_\theta(x' \mid x, c)
=
\prod_{t=1}^{T} q_\theta\!\left(x'_t \mid x'_{<t}, x, c\right)$.

Training minimizes the conditional next-token negative log-likelihood,
$\mathcal{L}(\theta)
=
- \mathbb{E}_{(x,x',c)\sim p_{\mathrm{pair}}}
[\log q_\theta(x' \mid x,c)]$. Because $p_{\mathrm{pair}}$ is supported on directed pairs that are both structurally similar and property-improved, minimizing this objective fits $q_\theta(x' \mid x,c)$ to the empirical distribution of local property-improved neighbors of $x$. Thus, pair construction encodes both structural locality and the desired direction of property change, while the conditional model allows multiple plausible improved targets per seed. 

PairIT provides a base matched-pair generator that can be extended in two complementary ways: iterative re-seeding or RL post-training can strengthen property optimization, while structured prompts provide explicit control over which properties should change and by how much.

\subsection{PairIT implementation and variants}

\textbf{Model architecture.}
PairIT uses a 68M-parameter decoder-only autoregressive transformer over SMILES strings, instantiated with a Llama-style architecture \citep{touvron2023llama} and adapted for molecular sequence generation following prior 
implementations \citep{frey2024cramming, ismailconcept}. 
We deliberately use a small model: SMILES generation has a substantially
narrower vocabulary than natural language, and our
model-size ablation shows little benefit from scaling across the 11M--200M
range tested (Appendix~\ref{app:model_size_ablation}). For paired training, we concatenate seed and target SMILES separated by a \texttt{<sep>} token, with generation conditioned on the seed. Full architecture and training details are in Appendix~\ref{sec:appendix_arch}.

\textbf{PairIT and iterative PairIT-Beam.}
Vanilla PairIT samples property-improved candidates directly from the matched-pair model. PairIT-Beam additionally applies iterative refinement: at each of three rounds, it generates 8 candidates per seed and retains the top 4 as seeds for the next round, yielding up to 72 candidates per original seed. This follows the recursive-inference idea of~\citet{damani2019black}.

\textbf{PairIT-RL.}
When pair supervision alone provides insufficient signal for property improvement, we optionally post-train PairIT using a property model as a reward. We use Group Relative Policy Optimization (GRPO)~\citep{shao2024deepseekmath} with a property-model-based reward and seed-similarity regularization to limit exploitation of the surrogate model. Full reward and training details are provided in Appendix~\ref{app:rl}.

\subsection{Prompt-conditioning}
\label{subsec:prompt_control}

\textbf{Single-property conditioning.}
Without prompting, the desired property change is implicit in each training pair. To make this control explicit, we prepend structured property tokens to the seed molecule~\citep{kotsias2020direct}. For a single conditioned property, the prompt specifies the property identity and a signed delta bin: the sign gives the desired direction and the bin magnitude gives the approximate requested change. Prompting changes only the conditioning sequence and training uses the same autoregressive objective as above.

\textbf{Multi-property conditioning.}
The same mechanism extends directly to multiple properties by concatenating one prompt block per requested property. During pretraining, a generic target slot can be trained using available properties and reassigned to a new downstream property during fine-tuning. For example, we pretrain on ZINC-250K using only RDKit properties and assign the generic target slot to the downstream MoleculeNet property during fine-tuning. We also apply property dropout during training so that subsets of the trained properties can be specified at inference time.

\section{Experimental Results}
\label{sec:experiments}
   \begin{figure}[t]
  \centering
  \includegraphics[width=1\textwidth]{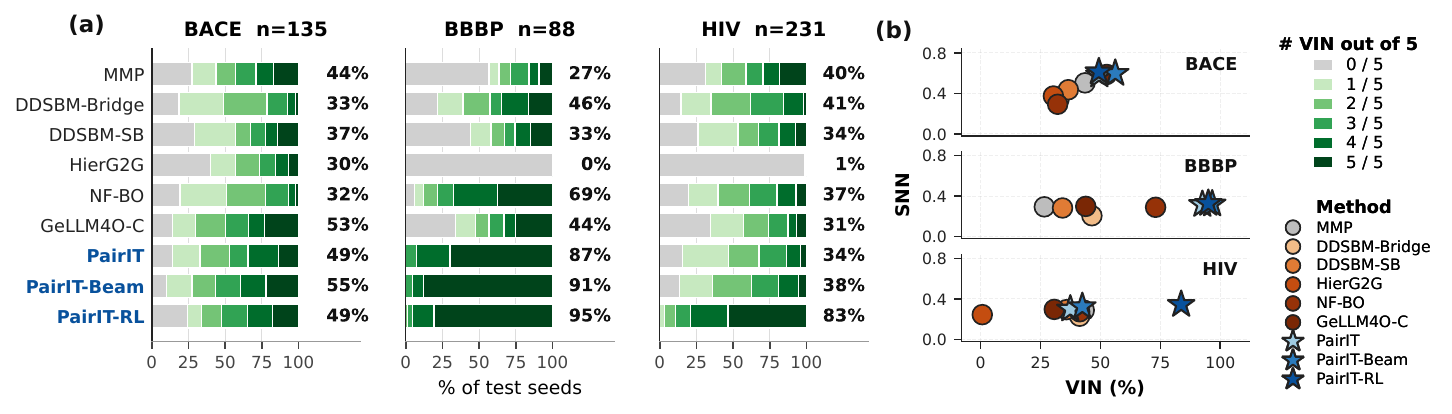}
    \caption{\small MoleculeNet seeded bioactivity optimization.
(a) Per-seed VIN (valid, novel, improved) distribution: each stacked bar gives the percentage of seeds for which $k$ of 5 retained generations are valid, novel, and improved.
(b) Aggregate VIN--SNN (similarity to nearest neighbor in the target class) tradeoff across methods.}
  \label{fig:main_results}
\end{figure}

\paragraph{Datasets and pair construction.} 
  We evaluate on three low-data MoleculeNet benchmarks (BACE, BBBP, HIV) using the class-balanced splits of GraphDiT~\citep{liu2024graph} which are binary tasks corresponding to $\beta$-secretase 1 inhibition, blood--brain barrier permeability, and HIV antiviral activity, respectively. We additionally evaluate on hERG~\citep{wang2016admet}, a low-data toxicity task from Therapeutics Data Commons (TDC)~\citep{huang2021therapeutics} (Appendix~\ref{app:herg}). For auxiliary pretraining, we construct ZINC-250K pairs by 2D Tanimoto similarity on Morgan fingerprints (ECFP4), directed toward a minimum logP improvement of $\Delta \mathrm{logP} > 1$, using 100 target neighbors per seed. For MoleculeNet fine-tuning, we construct matched molecular pairs (MMPs) using the \texttt{mmpdb} toolkit~\citep{dalke2018mmpdb}, directed from inactive (class~0) to active (class~1) molecules. The resulting MoleculeNet training sets contain only 88--662 pairs
from 517--1{,}420 molecules per dataset. Full pair construction details, including the rationale for the dataset-specific
pairing strategies, and dataset
statistics are in Appendix~\ref{app:pair_construction}.

   \begin{figure}[t]
   \vspace{-2mm}
    \centering
    \includegraphics[width=0.85\textwidth]{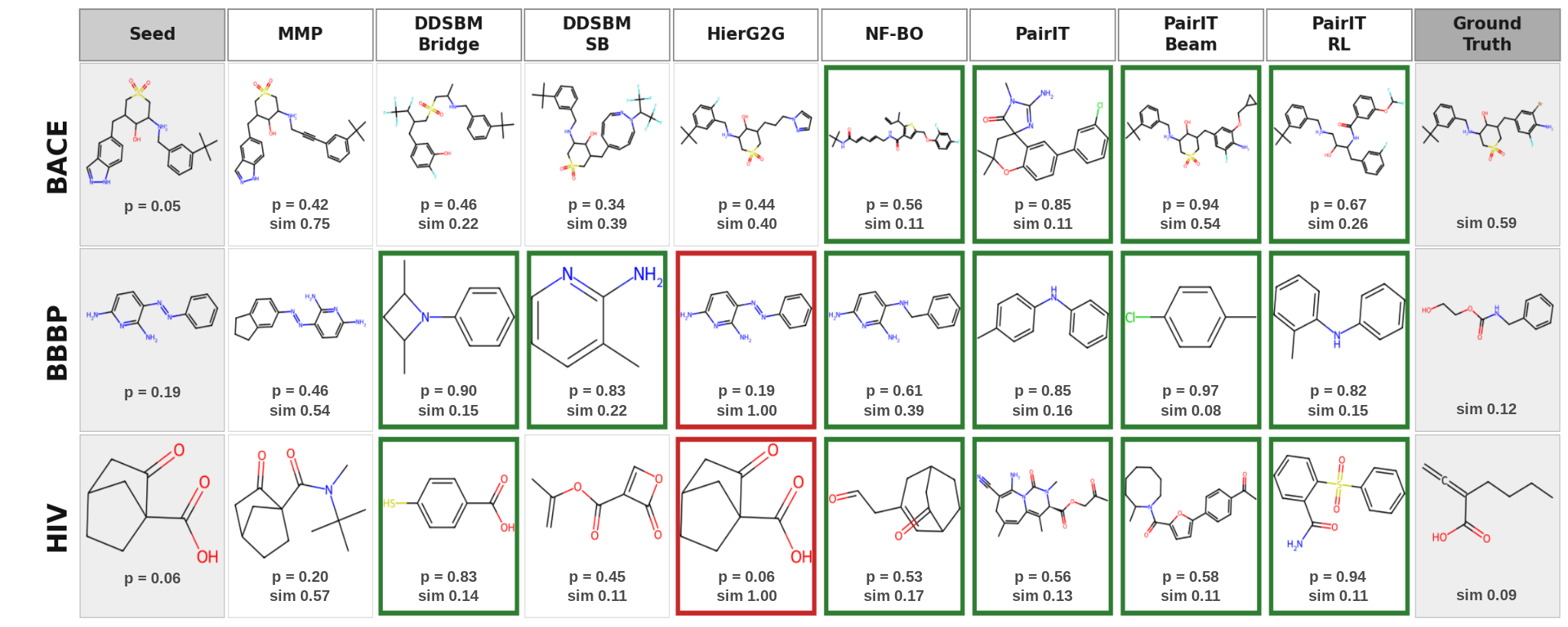}
    \caption{\small Example generations from PairIT and baselines. Leftmost column: the class-0 seed, rightmost column:  the corresponding \emph{ground-truth} class-1 MMP partner from the test set. Green = predicted class probability $>0.5$ (success); red = seed duplicate.}
    \label{fig:seeded_baselines}
\end{figure}

\textbf{Inference and evaluation.}
Each method produces 64 candidates per seed (PairIT-Beam: up to 72). We canonicalize and deduplicate the candidates, then rank them using validity and novelty together with property predictions from the random forest used in GraphDiT~\citep{liu2024graph} and Tanimoto similarity to the seed, retaining the top five candidates per seed for each method. These top-five selections are then independently scored with MiniMol~\citep{minimol} for the final reported results. Both property models are trained only on the corresponding MoleculeNet training splits. The same random forest is also used for NF-BO search, PairIT-Beam refinement, and PairIT-RL rewards. Agreement between the property predictions of these two oracles on held-out MoleculeNet test molecules and on generated molecules is analyzed in Appendix~\ref{app:oracle_agreement}.  Inference uses temperature 0.5; see Appendix~\ref{sec:appendix_temp} for a hyperparameter sweep. 

\textbf{Baselines.}
We compare against seeded generation baselines spanning hierarchical graph-to-graph translation (HierG2G)~\citep{jin2020hierarchical}, discrete diffusion Schrödinger bridge matching (DDSBM)~\citep{kimdiscrete}, latent Bayesian optimization with normalizing flows (NF-BO)~\citep{lee2025latent}, and the 7B-parameter instruction-tuned LLM GeLLM4O-C~\citep{dey2025large}. We also include MMP enumeration \citep{kramer2014matched} as a non-learning local-edit reference and GraphDiT \citep{liu2024graph} as a \textit{de novo} reference baseline. HierG2G and DDSBM are pretrained on the same ZINC-250K pair sets as PairIT; NF-BO uses the released GuacaMol checkpoint~\citep{brown2019guacamol}. \footnote{On BACE, the train-only Random Forest predicts zero class-1
probability for every class-0 test seed, causing NF-BO's GP surrogate to fail. Observation jitter did not resolve the issue, so we rerun NF-BO using a Random Forest trained on the full dataset, including the test set, giving NF-BO a slight advantage on BACE. See Appendix~\ref{appendix:baselines}.} All baselines are finetuned on the same MoleculeNet pairs as PairIT after pretraining. GeLLM4O-C uses the released 7B Mistral backbone with task-specific LoRA fine-tuning. Full baseline implementation details are in Appendix~\ref{appendix:baselines}.

\textbf{Metrics.}
We report \emph{Valid} (RDKit-parseable), \emph{Novel} (absent from the fine-tune training set), \emph{Improved} (class-1 generation from a class-0 seed on MoleculeNet), \emph{QED}~\citep{bickerton2012quantifying}, \emph{Unique} (distinct canonical SMILES across retained generations), and \emph{Diverse} (within-seed scaffold diversity). Our aggregate metric \emph{VIN} is the fraction of the top-5 retained generations per seed that are simultaneously Valid, Novel, and Improved. \emph{SNN} is the mean Tanimoto
similarity to the nearest target-class (class-1 for MoleculeNet corresponding to active molecules) training molecule over valid, non-memorized generations, measuring proximity to known target-class chemical space. Appendix~\ref{appendix:metrics} provides a broader distributional evaluation of the generated molecules relative to the target-class chemical space using additional metrics including average Tanimoto similarity to the seed, Fréchet ChemNet Distance, fragment and
scaffold similarity, and atom type coverage, allowing generation quality to be interpreted beyond the main VIN--SNN results.

\subsection{Single-property optimization from pair supervision.}

We first evaluate local single-property bioactivity optimization from class-0 seeds, with the goal of generating class-1 molecules, without prompting. Figure~\ref{fig:main_results} summarizes the per-seed VIN distribution and aggregate VIN--SNN tradeoff. Full VIN--seed-similarity distributions are shown in Appendix~\ref{app:vin_simseed_scatter} and additional failure-mode and pair-filtering analyses are in Appendices~\ref{app:failure_modes} and~\ref{app:property_filtering}, respectively.

Across MoleculeNet, PairIT and PairIT-Beam are competitive with the strongest baselines on BACE and substantially outperform them on BBBP, despite PairIT using 64 samples per seed versus 500 property-model evaluations for NF-BO and only 68M parameters versus 7B for GeLLM4O-C~\citep{dey2025large}. HIV is more challenging for pair supervision in general: PairIT and most baselines are around or below the non-learning MMP baseline (40.4\% VIN), which may reflect the greater heterogeneity of the active chemical space and
the weaker structural context shared by available pairs. Here PairIT-RL with post-training substantially improves performance to 82.6\% VIN. Because RL directly optimizes a learned property model and is therefore more sensitive to oracle quality, we treat PairIT-RL as an optional extension. Full RL training details and analyses are provided in Appendix~\ref{app:rl}. Additional results on hERG, a distinct toxicity dataset from TDC, are reported in Appendix~\ref{app:herg}.

\subsection{Pretraining ablations and training dynamics}

We first analyze the effects of pretraining and paired supervision separately. Without any ZINC-250K pretraining, training only on the small MoleculeNet pair sets leads to a large drop in VIN across all three tasks. Failure-mode analysis shows that these models generate many invalid SMILES (Appendix~\ref{app:pretrain_ablation}, Figure~\ref{fig:pretrain_failure_modes}), suggesting that a major role of pretraining in this low-data regime is to learn the SMILES grammar and generate valid molecules before downstream fine-tuning. Pretraining on molecules alone is not sufficient, however. Single-molecule pretraining on the same ZINC-250K distribution underperforms pretraining on directed pairs (Appendix~\ref{app:pretrain_ablation}, Table~\ref{tab:pretrain_pairs_vs_single}). Thus, the benefit is
not simply exposure to more molecules; pair supervision additionally teaches the seed-conditioned local transformations beneficial for molecular optimization. Within paired pretraining, performance tends to improve with the number of pairs per seed, although gains are not monotonic and eventually diminish. More diverse pair sets also increase seed similarity and expose the model to more unique molecules.

Surprisingly, the useful signal from paired ZINC-250K pretraining saturates very quickly. The best checkpoint by validation loss occurs before one full epoch over the pair dataset. Importantly, a pair epoch is not equivalent to a full pass over the unique molecules: because each molecule can be included in many matched pairs, the model repeatedly encounters the underlying molecules while seeing only a fraction of the constructed pair examples. Appendix~\ref{app:training_dynamics} explicitly quantifies the number of unique molecules encountered at the point of early stopping (Figure~\ref{fig:overfitting-vs-unique}), separating molecular coverage from pair-dataset coverage. Thus, early saturation does not imply that the model has seen little chemical diversity; rather, it suggests that additional pair examples become redundant relatively quickly.

\begin{wrapfigure}[16]{r}{0.42\textwidth}
      \vspace{-1.2em}
      \centering
      \includegraphics[width=\linewidth]{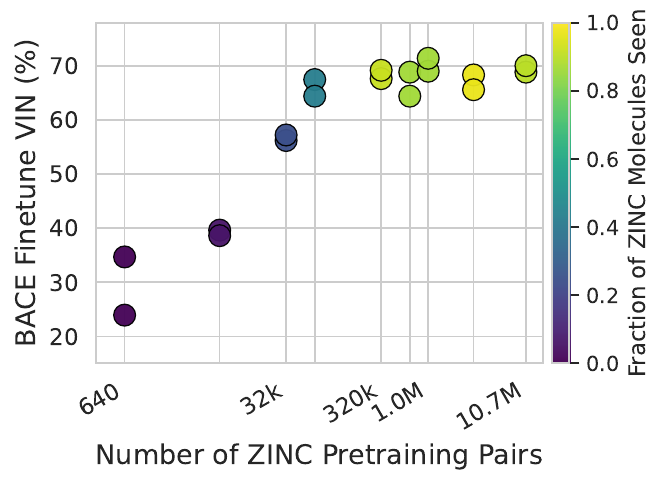}
      \caption{\small Scale of ZINC-250K pretraining dataset vs.\ downstream fine-tuned BACE performance. Saturation before one full epoch of pretraining is visible.}
      \label{fig:zinc-pretrain-size-vs-bace-vin}
  \end{wrapfigure}

Consistent with this, increasing the size of the paired pretraining set yields rapid initial gains followed by diminishing downstream improvement (\Cref{fig:zinc-pretrain-size-vs-bace-vin}). Once the model has learned to produce valid, similarity-preserving molecular edits, additional exposure to ZINC-250K pair examples appears to provide limited benefit, making aggressive early stopping important.

Importantly, ZINC-250K pretraining validation loss correlates with generation quality on the ZINC-250K validation set (Figure~\ref{fig:zinc-val-loss-and-zinc-vin}) and with downstream BACE VIN after fine-tuning (Figure~\ref{fig:zinc-val-loss-vs-bace-vin}). A similar relationship holds between pretraining validation loss and zero-shot logP optimization on MoleculeNet (Appendix~\ref{app:zero_shot_logp}, Figure~\ref{fig:zero-shot-logp-transfer}). We therefore select pretraining checkpoints by validation loss rather than training for a fixed number of epochs.

\subsection{Prompt-conditioned property control}
\label{sec:prompting_results}

\begin{figure}[t]
\centering
\includegraphics[width=0.85\textwidth]{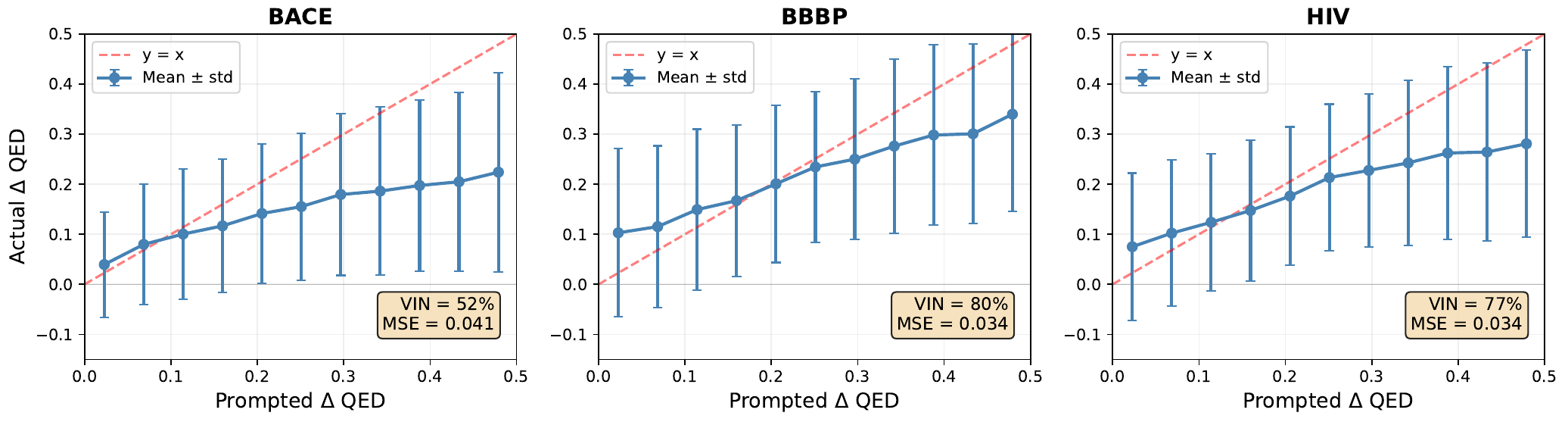}
\caption{\small QED controllability on MoleculeNet, with upward prompts over delta bins in $[0,0.5]$. Error bars show standard deviation across 5 generations per seed per bin. Realized changes scale with prompted magnitude, indicating coarse magnitude control. The corresponding logP results are in Appendix~\ref{app:single} (Figure~\ref{fig:single_prop_control_full}).}
\label{fig:single_prop_control}
\end{figure}

\begin{table}[t]
\centering
\scriptsize
\setlength{\tabcolsep}{3pt}
\caption{\small Two-property controllability on MoleculeNet, using 64 raw generations per held-out matched pair without top-five selection. \textbf{Per-property VIN}: is the
fraction of all attempts with an applicable nonzero request that are valid,
novel, and satisfy that property's criterion.
\textbf{Combined VIN}: fraction of generations that are valid, novel, and move
in the requested direction on \emph{both} prompted properties; zero-delta properties impose no condition, only pairs with at least one non-zero change requested are counted. For bioactivity, success
means an inactive seed yields a generation with train-only random-forest
activity probability greater than 0.5. \textbf{Random}: independently resample each
property's signed-bin index uniformly from 00--15 while retaining property
identities and block order. Both arms are scored against the original requests to serve as a baseline.
$\Delta$ is prompted minus randomized Combined VIN in percentage points, positive $\Delta$ in bold.}
\label{tab:normal_vs_random_property_ordering}

\begin{tabular}{llcccc}
\toprule
Setting & Dataset &
\makecell{\textbf{Combined}\\\textbf{VIN (\%)}} &
\makecell{\textbf{Per-property}\\\textbf{VIN (\%)}} &
\makecell{\textbf{Random}\\\textbf{(\%)}} &
\textbf{$\Delta$} \\
\midrule
Low-correlation pair & BACE & 12 & 13 / 16 & 17 & -4 \\
                     & BBBP & 32 & 38 / 35 & 21 & \textbf{+11} \\
                     & HIV  & 64 & 70 / 76 & 32 & \textbf{+32} \\
\midrule
QED + logP   & BACE & 19 & 22 / 21 & 15 & \textbf{+3} \\
             & BBBP & 30 & 33 / 43 & 17 & \textbf{+13} \\
             & HIV  & 42 & 55 / 63 & 30 & \textbf{+12} \\
\midrule
QED + bioactivity  & BACE &  8 & 12 / 10 &  7 & \textbf{+1} \\
                   & BBBP & 26 & 31 / 40 & 27 & -1 \\
                   & HIV  &  8 & 63 / 13 &  6 & \textbf{+2} \\
\midrule
TPSA + bioactivity & BACE & 11 & 17 / 13 & 12 & -1 \\
                   & BBBP & 48 & 53 / 55 & 37 & \textbf{+11} \\
                   & HIV  &  8 & 76 / 10 &  9 & 0 \\
\bottomrule
\end{tabular}
\end{table}

\begin{table}[t]
\centering
\scriptsize
\setlength{\tabcolsep}{4pt}
\caption{\small Direction adherence under multi-property prompting:
fraction of generations whose property delta matches the prompted direction,
averaged across the 9 non-activity properties. Higher is better.}
\label{tab:multiprop_adherence}

\begin{tabular}{lccc}
\toprule
Method & BACE & BBBP & HIV \\
\midrule
PairIT (10-property prompt) & \textbf{0.773} & \textbf{0.739} & \textbf{0.739} \\
Randomized prompt (control) & 0.433 & 0.499 & 0.559 \\
\bottomrule
\end{tabular}
\end{table}

Pair supervision implicitly determines the property changes represented by the training pairs. Prompting instead makes the desired property change explicit by encoding its direction and magnitude in conditioning tokens, allowing finer control over which properties are changed and by how much. We evaluate prompt conditioning in three settings, including single-property magnitude control, joint two-property control, and conditioning on many properties simultaneously. 

\paragraph{Single-property magnitude control.}
We first test whether the signed delta token carries information beyond a binary increase-or-decrease signal already available from the training pairs. On the small MoleculeNet datasets, we fine-tune with both orderings of each pair and sweep 16 signed delta bins at inference. Figure~\ref{fig:single_prop_control} shows the QED results: across all three datasets, larger requested increases generally produce larger realized changes. Although the realized changes do not match the requested bins exactly, this suggests that the prompt can support graded control even in the low-data MoleculeNet setting, despite each delta bin being trained on only a small subset of the available examples. Appendix~\ref{app:single} reports additional single-property results using logP, together with aggregate VIN and MSE values for both QED and logP (Figure~\ref{fig:single_prop_control_full}, Table~\ref{tab:single_prop_control}).

\paragraph{Two-property control.}
We next evaluate how well two requested property changes can be composed in the same prompt. On ZINC-250K, larger requested changes produce larger realized changes for both prompted properties, showing that graded control is retained under simultaneous conditioning (Appendix~\ref{app:multi}, Figure~\ref{fig:multi_property_delta}). On MoleculeNet, both properties are prompted simultaneously using signed delta bins derived from held-out matched test pairs, which provides chemically realistic requests. Combined VIN is computed by considering every prompt with at least one valid non-zero change and including generations that are valid, novel, and move in the prompted direction for the non-zero change properties. To test whether the model is actually using the conditioning tokens rather than merely reproducing correlations in the paired data, we compare against randomized prompt assignments (Table~\ref{tab:normal_vs_random_property_ordering}).

Prompt-conditioned generation exceeds the randomized-prompt baseline most consistently for descriptor pairs selected to have low empirical correlation and for QED+logP, indicating that the model can respond jointly to two property instructions. However, joint control involving bioactivity is less reliable. One likely reason is that, unlike continuously valued descriptors such as QED, logP, or TPSA, MoleculeNet bioactivity is binary and therefore provides a much coarser control signal. The small matched-pair datasets also contain relatively few examples of each joint property configuration. We therefore view the bioactivity results as evidence that joint control becomes harder when one
objective provides only coarse supervision, rather than as a failure of the prompting mechanism itself. Detailed two-property scatter plots and per-property results are provided in Appendix~\ref{app:two_property_control}.

\paragraph{Many-property control.}
Finally, we evaluate whether prompt dependence is retained when the conditioning space becomes substantially larger. We train a single model on nine computable RDKit properties plus a tenth target slot that is assigned to downstream
bioactivity during MoleculeNet fine-tuning. With all 10 properties prompted simultaneously, mean direction adherence across the nine non-activity properties reaches 0.739--0.773, compared with 0.433--0.559 under randomized prompts for matching the prompted increase/decrease direction (Table~\ref{tab:multiprop_adherence}). Thus, even in the 10-property setting, the generated edits remain substantially sensitive to the requested property directions rather than ignoring the prompt.

Many-property control is nevertheless sensitive to which objectives are included. Appendix~\ref{app:multi_prop_full} compares the full 10-property prompt with smaller property subsets and evaluates all $\binom{9}{3}=84$ choices of three auxiliary RDKit properties alongside bioactivity. These results show that conditioning on more properties is not necessarily better: performance can improve when the prompt is restricted to a smaller subset, and varies substantially depending on which auxiliary properties are included.

\section{Discussion}
\label{conclusions}

PairIT supports a progression from simple pair-supervised generation to more strongly guided optimization. Vanilla PairIT learns property-improving local edits directly from matched seed--target pairs. PairIT-Beam extends this by iteratively re-seeding high-scoring generations, allowing further property improvement at inference time at the cost of additional compute. When a reliable property oracle is available, PairIT-RL can push optimization further, but it is more sensitive to reward-model errors and can exhibit reward hacking under weak similarity constraints. Prompt conditioning adds a complementary form of control by making the desired direction and magnitude of property change explicit and extending the same framework to multi-property optimization.

PairIT also provides several ways to trade off property improvement against similarity to the seed. Sampling temperature changes this tradeoff without retraining (Fig.~\ref{fig:temp_metrics}), while the similarity-reward weight in PairIT-RL directly controls the balance between property optimization and seed similarity.

Pretraining is particularly important in the low-data setting. ZINC-250K pretraining substantially improves reliable molecular generation, while directed-pair pretraining is more effective than pretraining on the same molecules individually. More diverse pretraining pairs also change the locality of the learned edits (Appendix~\ref{app:pretrain_ablation}). At the same time, useful paired pretraining saturates quickly. In practice, this makes aggressive early stopping an important part of the training recipe.

On the surface, it may appear that the role of specialized domain knowledge has been diminished, yet it reappears in how the optimization problem is defined. The distance function determines what constitutes a ``local'' molecular edit, the construction of the paired data determines which transformations are learned, the prompted properties determine which objectives can be controlled, and the reward function determines what is optimized during post-training. As machine learning methods become increasingly effective at solving the problems we set before them, it becomes increasingly important that we formulate those problems well.

\textbf{Limitations and broader impact.}
\label{sec:limitations}
PairIT addresses the ``inner loop'' of generating high-scoring candidates under a scoring model, but lacks ``outer loop'' experimental validation. Any property oracle outside the wet lab is imperfect and susceptible to distribution shift, so oracle-dependent components inherit this uncertainty. Performance also depends on training-pair diversity, and multi-property control is less reliable when requested changes conflict with empirical property correlations or are weakly represented in the paired data (\Cref{tab:normal_vs_random_property_ordering}). We do not address synthesizability of generated candidate molecules, which we leave to future work. Generated candidates should be subject to expert review and safety screening before synthesis.

\clearpage
\bibliography{example_paper}

@article{pinheiro20233d,
      title={{3D} molecule generation by denoising voxel grids}, 
      author={Pedro O. Pinheiro and Joshua Rackers and Joseph Kleinhenz and Michael Maser and Omar Mahmood and Andrew Martin Watkins and Stephen Ra and Vishnu Sresht and Saeed Saremi},
      year={2023},
      journal={Advances in Neural Information Processing Systems}
}

@inproceedings{nowara2024nebula,
  title={{NEBULA}: Neural Empirical Bayes Under Latent Representations for Efficient and Controllable Design of Molecular Libraries},
  author={Nowara, Ewa and Pinheiro, Pedro O and Mahajan, Sai Pooja and Mahmood, Omar and Watkins, Andrew Martin and Saremi, Saeed and Maser, Michael},
  year={2024},
  booktitle={ICML 2024 AI for Science Workshop}
}

@article{vignac2023midi,
  title={{MiDi}: Mixed Graph and {3D} Denoising Diffusion for Molecule Generation},
  author={Vignac, Clement and Osman, Nagham and Toni, Laura and Frossard, Pascal},
  journal={arXiv preprint arXiv:2302.09048},
  year={2023}
}

@inproceedings{hoogeboom2022equivariant,
  title={Equivariant diffusion for molecule generation in {3D}},
  author={Hoogeboom, Emiel and Satorras, V{\i}ctor Garcia and Vignac, Cl{\'e}ment and Welling, Max},
  booktitle={International conference on machine learning},
  pages={8867--8887},
  year={2022}
}

@inproceedings{xu2023geometric,
  title={Geometric latent diffusion models for {3D} molecule generation},
  author={Xu, Minkai and Powers, Alexander S and Dror, Ron O and Ermon, Stefano and Leskovec, Jure},
  booktitle={International Conference on Machine Learning},
  pages={38592--38610},
  year={2023},
  organization={PMLR}
}

@inproceedings{joshi2025all,
  title={All-atom Diffusion Transformers: Unified Generative Modelling of Molecules and Materials},
  author={Joshi, Chaitanya K. and Fu, Xiang and Liao, Yi-Lun and Gharakhanyan, Vahe and Miller, Benjamin Kurt and Sriram, Anuroop and Ulissi, Zachary W.},
  booktitle={Proceedings of the 42nd International Conference on Machine Learning},
  series={Proceedings of Machine Learning Research},
  volume={267},
  pages={28393--28417},
  year={2025},
  publisher={PMLR}
}

@article{damani2019black,
  title={Black box recursive translations for molecular optimization},
  author={Damani, Farhan and Sresht, Vishnu and Ra, Stephen},
  journal={arXiv preprint arXiv:1912.10156},
  year={2019}
}

@inproceedings{kimdiscrete,
  title={Discrete Diffusion Schr{\"o}dinger Bridge Matching for Graph Transformation},
  author={Kim, Jun Hyeong and Kim, Seonghwan and Moon, Seokhyun and Kim, Hyeongwoo and Woo, Jeheon and Kim, Woo Youn},
  booktitle={The Thirteenth International Conference on Learning Representations},
  year={2025}
}

@article{liu2024graph,
  title={Graph diffusion transformers for multi-conditional molecular generation},
  author={Liu, Gang and Xu, Jiaxin and Luo, Tengfei and Jiang, Meng},
  journal={Advances in Neural Information Processing Systems},
  volume={37},
  pages={8065--8092},
  year={2024}
}

@inproceedings{jin2020hierarchical,
  title={Hierarchical generation of molecular graphs using structural motifs},
  author={Jin, Wengong and Barzilay, Regina and Jaakkola, Tommi},
  booktitle={International Conference on Machine Learning},
  pages={4839--4848},
  year={2020}
}

@inproceedings{lee2023exploring,
  title={Exploring chemical space with score-based out-of-distribution generation},
  author={Lee, Seul and Jo, Jaehyeong and Hwang, Sung Ju},
  booktitle={International Conference on Machine Learning},
  pages={18872--18892},
  year={2023}
}

@inproceedings{chitsaz2025novomolgen,
  title={{NovoMolGen}: Rethinking Molecular Language Model Pretraining},
  author={Chitsaz, Kamran and Balaji, Roshan and Fournier, Quentin and Bhatt, Nirav Pravinbhai and Chandar, Sarath},
  booktitle={International Conference on Learning Representations Generative AI and Biology (GenBio) Workshop},
year={2025}
}

@article{tagasovska2024implicitly,
  title={Implicitly guided design with {PropEn}: Match your data to follow the gradient},
  author={Tagasovska, Nata{\v{s}}a and Gligorijevic, Vladimir and Cho, Kyunghyun and Loukas, Andreas},
  journal={Advances in Neural Information Processing Systems},
  volume={37},
  pages={35973--36001},
  year={2024}
}

@inproceedings{pinheiroimplicit,
title={Generative Property Enhancer: Implicit Guided Generation Through Conditional Density Estimation},
author={Pinheiro, Pedro O and Kessel, Pan and Ismail, Aya Abdelsalam and Mahajan, Sai Pooja and Cho, Kyunghyun and Saremi, Saeed and Tagasovska, Natasa},
  booktitle={Advances in Neural Information Processing Systems},
  year={2025}
}

@article{weininger1988smiles,
  title={{SMILES}, a chemical language and information system. 1. Introduction to methodology and encoding rules},
  author={Weininger, David},
  journal={Journal of chemical information and computer sciences},
  volume={28},
  number={1},
  pages={31--36},
  year={1988},
  publisher={ACS Publications}
}

@inproceedings{huang2017arbitrary,
  title={Arbitrary style transfer in real-time with adaptive instance normalization},
  author={Huang, Xun and Belongie, Serge},
  booktitle={Proceedings of the IEEE international conference on computer vision},
  pages={1501--1510},
  year={2017}
}

@inproceedings{peebles2023scalable,
  title={Scalable diffusion models with transformers},
  author={Peebles, William and Xie, Saining},
  booktitle={Proceedings of the IEEE/CVF international conference on computer vision},
  pages={4195--4205},
  year={2023}
}

@article{irwin2012zinc,
  title={{ZINC}: a free tool to discover chemistry for biology},
  author={Irwin, John J and Sterling, Teague and Mysinger, Michael M and Bolstad, Erin S and Coleman, Ryan G},
  journal={Journal of chemical information and modeling},
  volume={52},
  number={7},
  pages={1757--1768},
  year={2012},
  publisher={ACS Publications}
}

@article{wu2018moleculenet,
  title={{MoleculeNet}: a benchmark for molecular machine learning},
  author={Wu, Zhenqin and Ramsundar, Bharath and Feinberg, Evan N and Gomes, Joseph and Geniesse, Caleb and Pappu, Aneesh S and Leswing, Karl and Pande, Vijay},
  journal={Chemical science},
  volume={9},
  number={2},
  pages={513--530},
  year={2018},
  publisher={Royal Society of Chemistry}
}

@inproceedings{frey2024cramming,
  title={Cramming Protein Language Model Training in 24 {GPU} Hours},
  author={Frey, Nathan C and Joren, Taylor and Ismail, Aya Abdelsalam and Goodman, Allen and Bonneau, Richard and Cho, Kyunghyun and Gligorijevic, Vladimir},
  booktitle={ICML 2024 Workshop on Efficient and Accessible Foundation Models for Biological Discovery},
  year={2024}
}

@inproceedings{ismailconcept,
  title={Concept Bottleneck Language Models for Protein Design},
  author={Ismail, Aya Abdelsalam and Oikarinen, Tuomas and Wang, Amy and Adebayo, Julius and Stanton, Samuel Don and Bravo, Hector Corrada and Cho, Kyunghyun and Frey, Nathan C},
  booktitle={The Thirteenth International Conference on Learning Representations},
  year={2025}
}

@article{touvron2023llama,
  title={{LLaMA}: Open and efficient foundation language models},
  author={Touvron, Hugo and Lavril, Thibaut and Izacard, Gautier and Martinet, Xavier and Lachaux, Marie-Anne and Lacroix, Timoth{\'e}e and Rozi{\`e}re, Baptiste and Goyal, Naman and Hambro, Eric and Azhar, Faisal and others},
  journal={arXiv preprint arXiv:2302.13971},
  year={2023}
}

@article{Bender2004MolecularSA,
  title={Molecular Similarity: A Key Technique in Molecular Informatics},
  author={Bender, Andreas and Glen, Robert C.},
  journal={Organic \& Biomolecular Chemistry},
  volume={2},
  number={22},
  pages={3204--3218},
  year={2004}
}

@article{rogers2010extended,
  title   = {Extended-Connectivity Fingerprints},
  author  = {Rogers, David and Hahn, Mathew},
  journal = {Journal of Chemical Information and Modeling},
  volume  = {50},
  number  = {5},
  pages   = {742--754},
  year    = {2010},
  doi     = {10.1021/ci100050t}
}

@article{kramer2014matched,
  title={Matched molecular pair analysis: significance and the impact of experimental uncertainty},
  author={Kramer, Christian and Fuchs, Julian E and Whitebread, Steven and Gedeck, Peter and Liedl, Klaus R},
  journal={Journal of medicinal chemistry},
  volume={57},
  number={9},
  pages={3786--3802},
  year={2014},
  publisher={ACS Publications}
}

@article{bohacek1996art,
  title={The art and practice of structure-based drug design: a molecular modeling perspective},
  author={Bohacek, Regine S and McMartin, Colin and Guida, Wayne C},
  journal={Medicinal research reviews},
  volume={16},
  number={1},
  pages={3--50},
  year={1996},
  publisher={Wiley Subscription Services, Inc., A Wiley Company New York}
}

@inproceedings{nowara2025we,
  title={Do we need equivariant models for molecule generation?},
  author={Nowara, Ewa and Rackers, Joshua A and Suriana, Patricia Adriana and Kessel, Pan and Shen, Max W and Watkins, Andrew Martin and Maser, Michael},
  booktitle={International Conference on Learning Representations Generative AI and Biology (GenBio) Workshop},
  year=2025
}

@article{shen2025training,
  title={Training Dynamics of Learning {3D}-Rotational Equivariance},
  author={Shen, Max W and Nowara, Ewa and Maser, Michael and Cho, Kyunghyun},
  journal={arXiv preprint arXiv:2512.02303},
  year={2025}
}

@article{kumar2024conditional,
  title={Conditional Diffusion on Web-Scale Image Pairs leads to Diverse Image Variations},
  author={Kumar, Manoj and Houlsby, Neil and Hoogeboom, Emiel},
  journal={arXiv preprint arXiv:2405.14857},
  year={2024}
}

@inproceedings{chen2024generalists,
  title={Generalists vs. Specialists: Evaluating {LLM}s on Highly-Constrained Biophysical Sequence Optimization Tasks},
  author={Chen, Angelica and Stanton, Samuel Don and Ding, Frances and Alberstein, Robert G and Watkins, Andrew Martin and Bonneau, Richard and Gligorijevic, Vladimir and Cho, Kyunghyun and Frey, Nathan C},
  booktitle={International Conference on Machine Learning},
  pages={9029--9072},
  year={2025},
  organization={PMLR}
}

@article{dey2025large,
  title={Large Language Models for Controllable Multi-property Multi-objective Molecule Optimization},
  author={Dey, Vishal and Hu, Xiao and Ning, Xia},
  journal={arXiv preprint arXiv:2505.23987},
  year={2025}
}

@article{lee2025latent,
  title={Latent bayesian optimization via autoregressive normalizing flows},
  author={Lee, Seunghun and Park, Jinyoung and Chu, Jaewon and Yoon, Minseo and Kim, Hyunwoo J},
  journal={arXiv preprint arXiv:2504.14889},
  year={2025}
}

@article{wang2007apparent,
  title={Apparent asymmetry in fingerprint similarity searching is a direct consequence of differences in bit densities and molecular size},
  author={Wang, Yuan and Eckert, Hanna and Bajorath, J{\"u}rgen},
  journal={ChemMedChem},
  volume={2},
  number={7},
  pages={1037--1042},
  year={2007},
  doi={10.1002/cmdc.200700050}
}

@article{preuer2018frechet,
  title={Fr{\'e}chet {ChemNet} distance: a metric for generative models for molecules in drug discovery},
  author={Preuer, Kristina and Renz, Philipp and Unterthiner, Thomas and Hochreiter, Sepp and Klambauer, Gunter},
  journal={Journal of chemical information and modeling},
  volume={58},
  number={9},
  pages={1736--1741},
  year={2018},
  publisher={ACS Publications}
}

@inproceedings{rafailov2024direct,
  title     = {Direct Preference Optimization: Your Language Model is Secretly a Reward Model},
  author    = {Rafailov, Rafael and Sharma, Archit and Mitchell, Eric and Manning, Christopher D. and Ermon, Stefano and Finn, Chelsea},
  booktitle = {Advances in Neural Information Processing Systems},
  volume    = {36},
  year      = {2023},
  url       = {https://papers.nips.cc/paper_files/paper/2023/hash/a85b405ed65c6477a4fe8302b5e06ce7-Abstract-Conference.html}
}

@inproceedings{xiong2025projection,
  title     = {Projection Optimization: A General Framework for Multi-Objective and Multi-Group {RLHF}},
  author    = {Xiong, Nuoya and Singh, Aarti},
  booktitle = {Proceedings of the 42nd International Conference on Machine Learning},
  series    = {Proceedings of Machine Learning Research},
  volume    = {267},
  pages     = {69017--69058},
  year      = {2025},
  publisher = {PMLR},
  url       = {https://proceedings.mlr.press/v267/xiong25c.html}
}

@inproceedings{zhou2024modpo,
  title     = {Beyond One-Preference-Fits-All Alignment: Multi-Objective Direct Preference Optimization},
  author    = {Zhou, Zhanhui and Liu, Jie and Shao, Jing and Yue, Xiangyu and Yang, Chao and Ouyang, Wanli and Qiao, Yu},
  booktitle = {Findings of the Association for Computational Linguistics: ACL 2024},
  pages     = {10586--10613},
  address   = {Bangkok, Thailand},
  year      = {2024},
  publisher = {Association for Computational Linguistics},
  doi       = {10.18653/v1/2024.findings-acl.630},
  url       = {https://aclanthology.org/2024.findings-acl.630/}
}

@inproceedings{yang2024rewards,
  title     = {{Rewards-in-Context}: Multi-objective Alignment of Foundation Models with Dynamic Preference Adjustment},
  author    = {Yang, Rui and Pan, Xiaoman and Luo, Feng and Qiu, Shuang and Zhong, Han and Yu, Dong and Chen, Jianshu},
  booktitle = {Proceedings of the 41st International Conference on Machine Learning},
  series    = {Proceedings of Machine Learning Research},
  volume    = {235},
  pages     = {56276--56297},
  year      = {2024},
  publisher = {PMLR},
  url       = {https://proceedings.mlr.press/v235/yang24q.html}
}

@article{gupta2025robust,
  title   = {Robust Multi-Objective Preference Alignment with Online {DPO}},
  author  = {Gupta, Raghav and Sullivan, Ryan and Li, Yunxuan and Phatale, Samrat and Rastogi, Abhinav},
  journal = {Proceedings of the AAAI Conference on Artificial Intelligence},
  volume  = {39},
  number  = {26},
  pages   = {27321--27329},
  year    = {2025},
  doi     = {10.1609/aaai.v39i26.34942},
  url     = {https://ojs.aaai.org/index.php/AAAI/article/view/34942}
}

@article{park2023preference,
  title   = {Preference Optimization for Molecular Language Models},
  author  = {Park, Ryan and Theisen, Ryan and Sahni, Navriti and Patek, Marcel and Cicho{\'n}ska, Anna and Rahman, Rayees},
  journal = {arXiv preprint arXiv:2310.12304},
  year    = {2023},
  doi     = {10.48550/arXiv.2310.12304},
  url     = {https://arxiv.org/abs/2310.12304}
}

@article{calanzone2025molmoe,
  title   = {{Mol-MoE}: Training Preference-Guided Routers for Molecule Generation},
  author  = {Calanzone, Diego and {D'Oro}, Pierluca and Bacon, Pierre-Luc},
  journal = {arXiv preprint arXiv:2502.05633},
  year    = {2025},
  doi     = {10.48550/arXiv.2502.05633},
  url     = {https://arxiv.org/abs/2502.05633}
}

@article{bickerton2012quantifying,
  title={Quantifying the Chemical Beauty of Drugs},
  author={Bickerton, G. Richard and Paolini, Gaia V. and Besnard, J{\'e}r{\'e}my and Muresan, Sorel and Hopkins, Andrew L.},
  journal={Nature Chemistry},
  volume={4},
  number={2},
  pages={90--98},
  year={2012},
  doi={10.1038/nchem.1243}
}

@article{shao2024deepseekmath,
  title={{DeepSeekMath}: Pushing the Limits of Mathematical Reasoning in Open Language Models},
  author={Shao, Zhihong and Wang, Peiyi and Zhu, Qihao and Xu, Runxin and Song, Junxiao and Bi, Xiao and Zhang, Haowei and Zhang, Mingchuan and Li, Y. K. and Wu, Y. and Guo, Daya},
  journal={arXiv preprint arXiv:2402.03300},
  year={2024},
  doi={10.48550/arXiv.2402.03300},
  url={https://arxiv.org/abs/2402.03300}
}

@article{kotsias2020direct,
  title = {Direct steering of de novo molecular generation with descriptor conditional recurrent neural networks},
  author = {Kotsias, Panagiotis-Christos and Ar{\'u}s-Pous, Josep and Chen, Hongming and Engkvist, Ola and Tyrchan, Christian and Bjerrum, Esben Jannik},
  journal = {Nature Machine Intelligence},
  volume = {2},
  number = {5},
  pages = {254--265},
  year = {2020},
  publisher = {Springer Nature},
  doi = {10.1038/s42256-020-0174-5},
  url = {https://doi.org/10.1038/s42256-020-0174-5}
}

@article{aldeghi2022roughness,
  title={Roughness of Molecular Property Landscapes and Its Impact on Modellability},
  author={Aldeghi, Matteo and Graff, David E. and Frey, Nathan and Morrone, Joseph A. and Pyzer-Knapp, Edward O. and Jordan, Kirk E. and Coley, Connor W.},
  journal={Journal of Chemical Information and Modeling},
  volume={62},
  number={19},
  pages={4660-4671},
  year={2022},
  doi={10.1021/acs.jcim.2c00903},
  url = {https://doi.org/10.1021/acs.jcim.2c00903}
}

@article{tagasovska2025property,
  title={Property Enhancer--a data efficient multi-objective approach for functional antibody optimization},
  author={Tagasovska, Nata{\v{s}}a and Ludwiczak, Jan and Loukas, Andreas and Makowski, Emily K and Mohammadi-Peyhani, Homa and Mahajan, Sai Pooja and Zadorozhny, Karina and Lee, Donald and Goodman, Allen and Lin, Joshua Yao-Yu and others},
  journal={bioRxiv},
  year={2025},
  doi={10.1101/2025.10.08.681101},
  url={https://www.biorxiv.org/content/10.1101/2025.10.08.681101v1}
}

@inproceedings{park2024botied,
  title={{BOtied}: Multi-objective Bayesian optimization with tied multivariate ranks},
  author={Park, Ji Won and Tagasovska, Natasa and Maser, Michael and Ra, Stephen and Cho, Kyunghyun},
  booktitle={International Conference on Machine Learning},
  pages={39813--39833},
  year={2024},
  organization={PMLR}
}

@article{gomez2018automatic,
  title={Automatic chemical design using a data-driven continuous representation of molecules},
  author={G{\'o}mez-Bombarelli, Rafael and Wei, Jennifer N and Duvenaud, David and Hern{\'a}ndez-Lobato, Jos{\'e} Miguel and S{\'a}nchez-Lengeling, Benjam{\'\i}n and Sheberla, Dennis and Aguilera-Iparraguirre, Jorge and Hirzel, Timothy D and Adams, Ryan P and Aspuru-Guzik, Al{\'a}n},
  journal={ACS central science},
  volume={4},
  number={2},
  pages={268--276},
  year={2018},
  publisher={ACS Publications}
}

@inproceedings{jinlearning,
  title={Learning Multimodal Graph-to-Graph Translation for Molecule Optimization},
  author={Jin, Wengong and Yang, Kevin and Barzilay, Regina and Jaakkola, Tommi},
  booktitle={International Conference on Learning Representations},
  year={2019}
}

@inproceedings{razin2025unintentional,
  title={Unintentional Unalignment: Likelihood Displacement in Direct Preference Optimization},
  author={Razin, Noam and Malladi, Sadhika and Bhaskar, Adithya and Chen, Danqi and Arora, Sanjeev and Hanin, Boris},
  booktitle={The Thirteenth International Conference on Learning Representations},
  year={2025}
}

@article{olivecrona2017molecular,
  title = {Molecular de-novo design through deep reinforcement learning},
  author = {Olivecrona, Marcus and Blaschke, Thomas and Engkvist, Ola and Chen, Hongming},
  journal = {Journal of Cheminformatics},
  volume = {9},
  number = {1},
  pages = {48},
  year = {2017},
  doi = {10.1186/s13321-017-0235-x}
}

@article{haddad2025targeted,
  title = {Targeted molecular generation with latent reinforcement learning},
  author = {Haddad, Ragy and Litsa, Eleni E. and Liu, Zhen and Yu, Xin and Burkhardt, Daniel and Bhisetti, Govinda},
  journal = {Scientific Reports},
  volume = {15},
  number = {1},
  pages = {15202},
  year = {2025},
  doi = {10.1038/s41598-025-99785-0}
}

@article{dalke2018mmpdb,
    title   = {{mmpdb}: An Open-Source Matched Molecular Pair Platform for Large Multiproperty Data Sets},
    author  = {Dalke, Andrew and Hert, Jerome and Kramer, Christian},
    journal = {Journal of Chemical Information and Modeling},
    volume  = {58},
    number  = {5},
    pages   = {902--910},
    year    = {2018},
    doi     = {10.1021/acs.jcim.8b00173}
  }

@article{minimol,
  title={{MiniMol}: A Parameter-Efficient Foundation Model for Molecular Learning},
  author={Kl{\"a}ser, Kerstin and Banaszewski, B{\l}a{\.z}ej and Maddrell-Mander, Samuel and McLean, Callum and M{\"u}ller, Luis and Parviz, Ali and Huang, Shenyang and Fitzgibbon, Andrew},
  journal={arXiv preprint arXiv:2404.14986},
  year={2024}
}

@article{cavanagh2024smileylama,
    title={{SmileyLlama}: Modifying Large Language Models for Directed Chemical Space Exploration},
    author={Cavanagh, Joseph M. and Sun, Kunyang and Gritsevskiy, Andrew and Bagni, Dorian and Wang, Yingze and Bannister, Thomas D. and Head-Gordon,
  Teresa},
    journal={arXiv preprint arXiv:2409.02231},
    year={2024}
  }

@article{ye2023drugassist,
    title={{DrugAssist}: A Large Language Model for Molecule Optimization},
    author={Ye, Geyan and Cai, Xibao and Lai, Houtim and Wang, Xing and Huang, Junhong and Wang, Longyue and Liu, Wei and Zeng, Xiangxiang},
    journal={arXiv preprint arXiv:2401.10334},
    year={2023}
  }

@article{brown2019guacamol,
    author  = {Brown, Nathan and Fiscato, Marco and Segler,
  Marwin H. S. and Vaucher, Alain C.},
    title   = {{GuacaMol}: Benchmarking Models for de Novo
  Molecular Design},
    journal = {Journal of Chemical Information and Modeling},
    volume  = {59},
    number  = {3},
    pages   = {1096--1108},
    year    = {2019},
    doi     = {10.1021/acs.jcim.8b00839},
    publisher = {American Chemical Society},
  }

@article{huang2021therapeutics,
  title={Therapeutics Data Commons: Machine Learning Datasets and Tasks for Drug Discovery and Development},
  author={Huang, Kexin and Fu, Tianfan and Gao, Wenhao and Zhao, Yue and Roohani, Yusuf and Leskovec, Jure and Coley, Connor W. and Xiao, Cao and Sun, Jimeng and Zitnik, Marinka},
  journal={Proceedings of Neural Information Processing Systems, NeurIPS Datasets and Benchmarks},
  year={2021}
}

@article{wang2016admet,
  title={{ADMET} evaluation in drug discovery. 16. Predicting {hERG} blockers by combining multiple pharmacophores and machine learning approaches},
  author={Wang, Shuangquan and Sun, Huiyong and Liu, Hui and Li, Dan and Li, Youyong and Hou, Tingjun},
  journal={Molecular pharmaceutics},
  volume={13},
  number={8},
  pages={2855--2866},
  year={2016},
  publisher={ACS Publications}
}
\bibliographystyle{plainnat}

\newpage
\appendix

\onecolumn

\clearpage

\startcontents[appendices]
\printcontents[appendices]{l}{1}{\section*{Appendix Contents}\setcounter{tocdepth}{2}}

\section{Related work comparison}
\label{app:related_work}

Table~\ref{tab:method_comparison} summarizes how PairIT compares with related molecular generation and optimization methods along the dimensions most relevant to our setting. \textit{property model-free} denotes methods that do not use iterative property model-guided search during generation.

\begin{table}[!htbp]
\centering
\small
\setlength{\tabcolsep}{3pt}
\renewcommand{\arraystretch}{1.15}
\begin{tabularx}{\textwidth}{lccccc>{\raggedright\arraybackslash}X}
\toprule
\textbf{Method} 
& \makecell{\textbf{Seeded}} 
& \makecell{\textbf{Single}\\\textbf{Step}} 
& \makecell{\textbf{Multi-}\\\textbf{Property}} 
& \makecell{\textbf{Small}\\\textbf{Datasets}}
& \makecell{\textbf{Property}\\\textbf{Model-Free}}   & \makecell[l]{\textbf{Optimization}\\\textbf{Mechanism}} \\
\midrule

\makecell[l]{GraphDiT\\\hspace{0.75em}{\footnotesize \citep{liu2024graph}}}
& $\times$ & $\checkmark$ & $\checkmark$ & $\checkmark$ & $\checkmark$ & AdaLN \\

\makecell[l]{PropEn\\\hspace{0.75em}{\footnotesize \citep{tagasovska2024implicitly}}}
& $\checkmark$ & $\times$ & $\times$ & $\checkmark$ & $\checkmark$ & Pairs \\

\makecell[l]{LLOME\\\hspace{0.75em}{\footnotesize \citep{chen2024generalists}}}
& $\checkmark$ & $\times$ & $\times$ & $\checkmark$ & $\checkmark$ & LLM + online black-box optimization \\

\makecell[l]{NF-BO\\\hspace{0.75em}{\footnotesize \citep{lee2025latent}}}
& $\checkmark$ & $\times$ & $\checkmark$ & $\times$ & $\times$ & Bayesian optimization \\

\makecell[l]{HierG2G\\\hspace{0.75em}{\footnotesize \citep{jin2020hierarchical}}}
& $\checkmark$ & $\checkmark$ & $\times$ & $\times$ & $\checkmark$ & Graph translation \\

\makecell[l]{DDSBM\\\hspace{0.75em}{\footnotesize \citep{kimdiscrete}}}
& $\checkmark$ & $\checkmark$ & $\times$ & $\times$ & $\checkmark$ & Schr\"odinger bridge \\

\makecell[l]{GeLLM4O-C\\\hspace{0.75em}{\footnotesize \citep{dey2025large}}}
& $\checkmark$ & $\checkmark$ & $\checkmark$ & $\checkmark$ & $\checkmark$ & LLM prompt \\

\midrule
\textbf{PairIT} 
& $\checkmark$ & $\checkmark$ & $\checkmark$ & $\checkmark$ & Optional & Pairs \\
\bottomrule
\end{tabularx}
\caption{Comparison of molecular generation and optimization methods.}
\label{tab:method_comparison}
\end{table}

\subsection{Selected \textit{de novo} generation related work}

A large body of work targets \emph{de novo} property-conditioned generation, including equivariant diffusion~\citep{hoogeboom2022equivariant, vignac2023midi, liu2024graph}, flow-matching~\citep{joshi2025all}, walk-jump-style methods~\citep{pinheiro20233d, nowara2024nebula}, non-equivariant 
alternatives with augmentation~\citep{nowara2025we, shen2025training}, and SMILES language models~\citep{chitsaz2025novomolgen}; property control in these models typically enters through activations (e.g., AdaLN-style modulation~\citep{huang2017arbitrary, peebles2023scalable}) or inputs.

\FloatBarrier

\section{Implementation and evaluation details}
\label{app:method_experimental_details}

\subsection{Model architecture and training hyperparameters}
\label{sec:appendix_arch}


SMILES are tokenized using a regex-based pre-tokenizer that separates
chemically meaningful SMILES units, followed by a fixed 1,225-token WordLevel
vocabulary. No BPE or learned merges are used. The vocabulary includes the special tokens \texttt{<cls>}, \texttt{<pad>},
\texttt{<eos>}, \texttt{<unk>}, \texttt{<mask>}, and \texttt{<sep>}. All PairIT experiments were run on multi-GPU cluster nodes with NVIDIA~B200
GPUs (180~GB per GPU). Pretraining runs took up to 1.5 hours per run.

\begin{table}[!htbp]
\centering
\small
\caption{PairIT architecture and training hyperparameters for the 68M model used in the main experiments.}
\label{tab:pairit_hparams}
\begin{tabular}{ll}
\toprule
Setting & Value \\
\midrule
Layers & 12 \\
Hidden size & 648 \\
Attention heads & 12 \\
KV heads & 12 (MHA) \\
FFN size & 2,048 \\
Vocabulary size & 1,225 \\
Maximum sequence length & 1,024 \\
Optimizer & AdamW ($\beta_1{=}0.9$, $\beta_2{=}0.98$, $\varepsilon{=}10^{-12}$) \\
Learning rate & $10^{-4}$ \\
Weight decay & 0.01 \\
Warmup steps / fraction & 1,000 steps \\
Scheduler & Constant with warmup \\
Batch size & 64 \\
Gradient accumulation & 1 \\
Precision & FP32 \\
Pretraining duration & up to 1.5 hours \\
Fine-tuning epochs / steps & Best checkpoint by validation loss \\
Inference temperature & 0.5 \\
\bottomrule
\end{tabular}
\end{table}

\FloatBarrier

\subsection{Model-size ablation}
\label{app:model_size_ablation}

\begin{figure}[t]
    \centering
    \includegraphics[width=\textwidth]{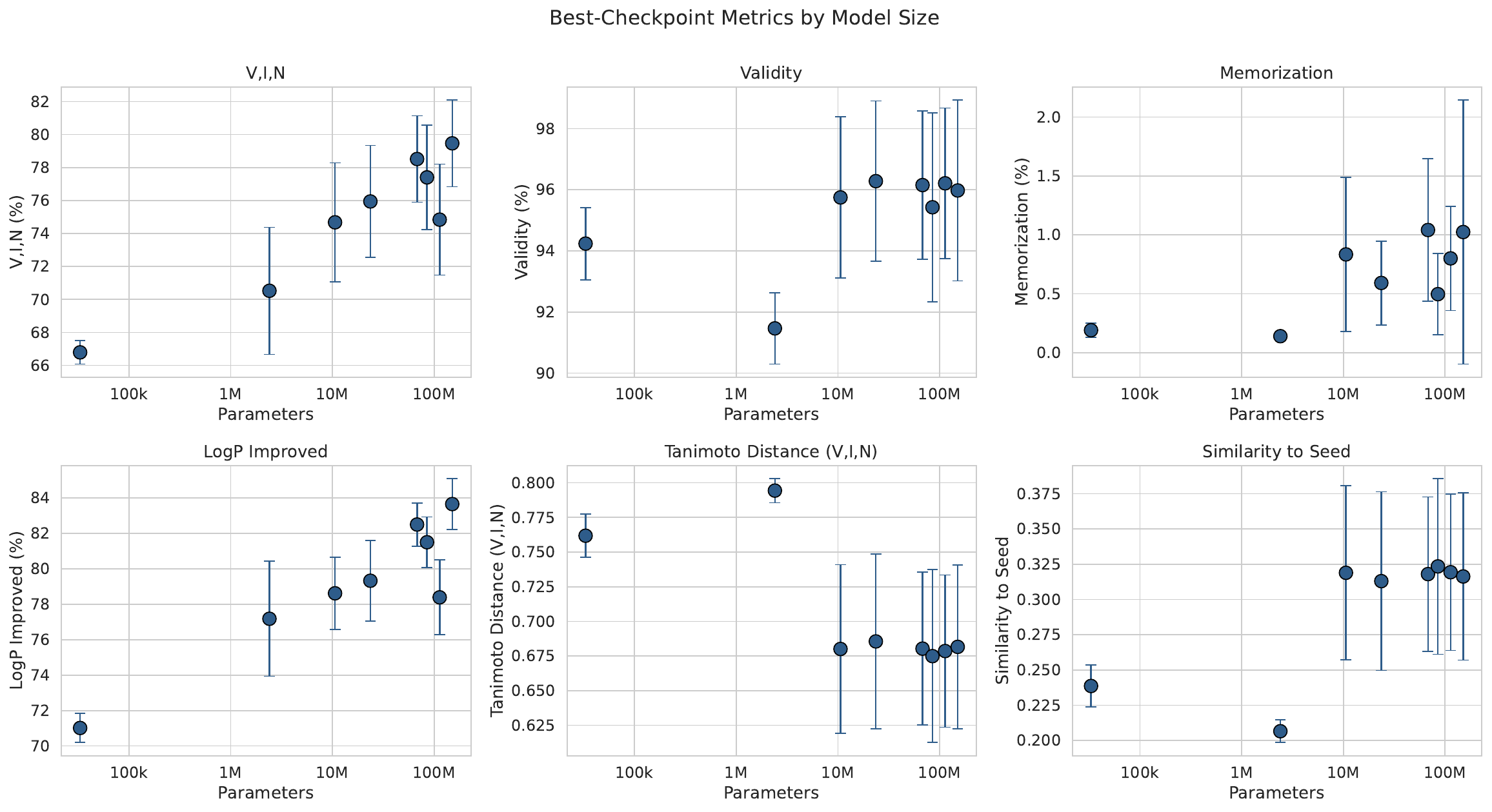} 
    \caption{We observe negligible gains from scaling the model size; larger models show increased memorization, indicating overfitting without performance benefits.}
    \label{fig:model_size}
\end{figure}

We analyze the effect of model scaling on generation quality and find that increasing model size yields little to no improvement across key metrics, including validity, VIN (valid, property improved and novel) fraction, target property optimization, and diversity. Performance plateaus already at medium scales, with the 68M model achieving comparable results to larger models. In contrast, the largest models show an increase in memorization, suggesting overfitting to the training data without meaningful gains in generation quality. Based on this trade-off, we use the 68M model as a practical default.

\FloatBarrier

\subsection{Inference-temperature sweep}
\label{sec:appendix_temp}

\begin{figure}[!htbp]
    \centering
    \includegraphics[width=0.85\textwidth]{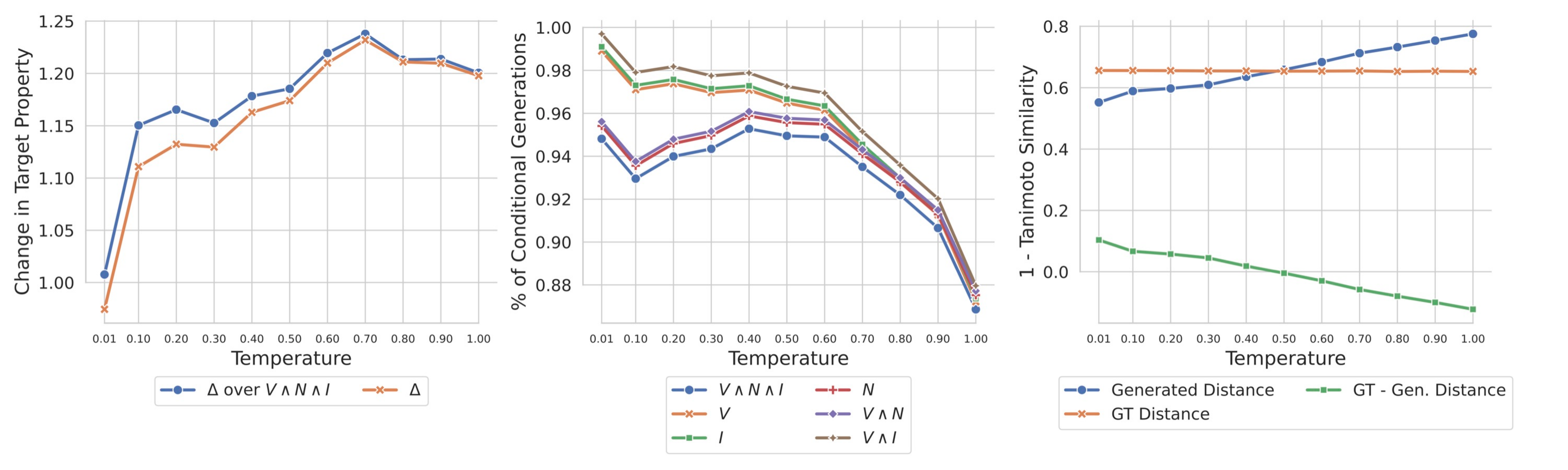}    
    \caption{Inference-time temperature sweep ($0.1$--$1.0$) showing the trade-off between diversity and validity: lower temperatures favor similarity to the seed, while higher temperatures increase novelty and property improvement at the cost of validity.}
    \label{fig:temp_metrics}
\end{figure}

The temperature is an important sampling parameter for PairIT that must be tuned. We vary the temperature at inference across a range of $0.1$--$1.0$. It modulates the randomness of softmax sampling during generation. As shown in Figure~\ref{fig:temp_metrics}, lower temperatures reduce diversity and lead to near-repetition of the seed, while higher temperatures increase novelty but degrade validity. Higher temperatures tend to yield larger improvements in the target property, at the cost of reduced fidelity. We use the temperature of 0.5 for all datasets.

\begin{figure}[t]
    \centering
    \includegraphics[width=0.85\textwidth]{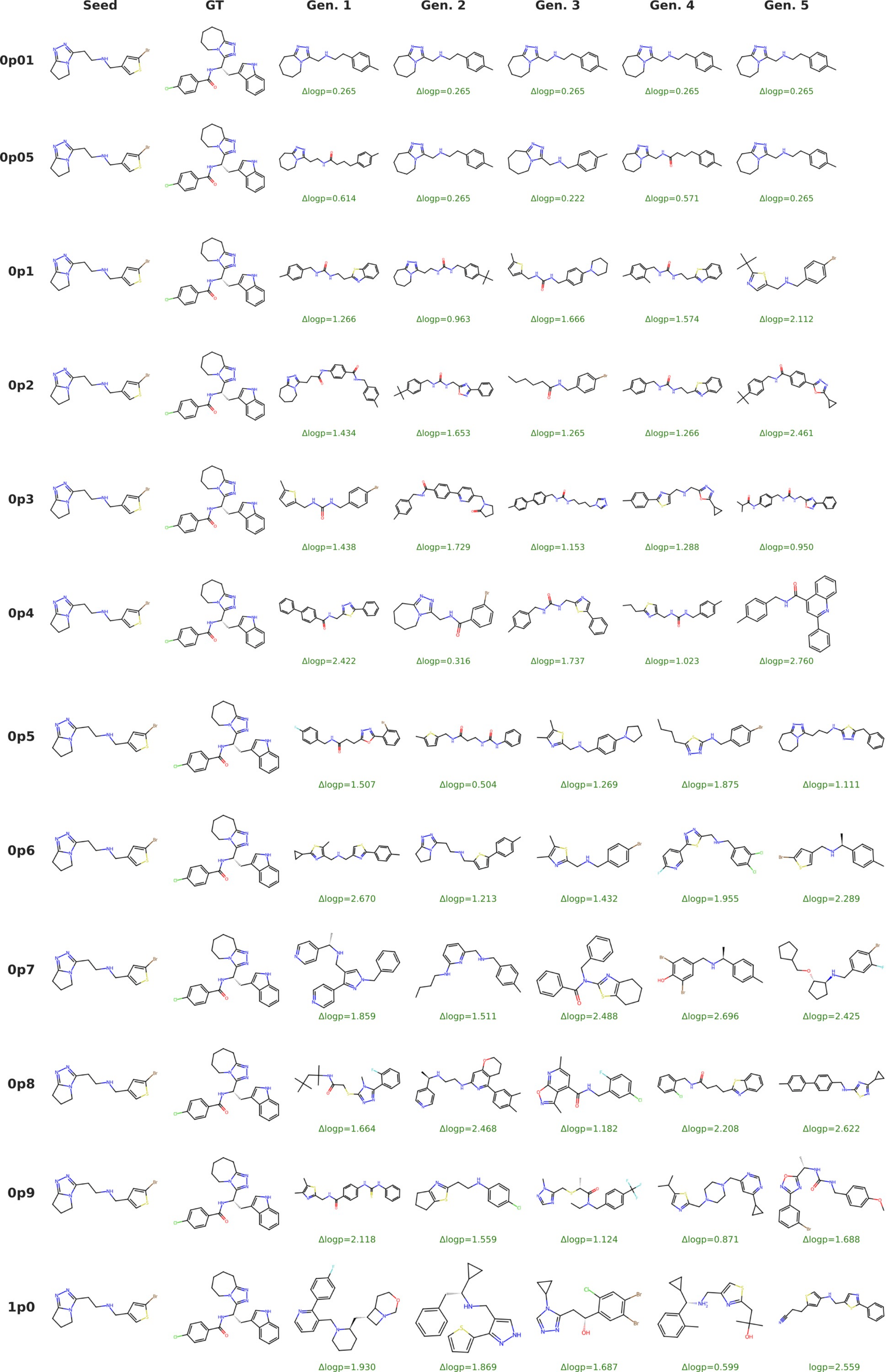}
    \caption{Examples of generations with different temperature settings.}
    \label{fig:temp_gens}
\end{figure}

\clearpage

\subsection{Pair construction details}
\label{app:pair_construction}

For ZINC-250K auxiliary pretraining, we construct directed logP-improvement pairs using 2D Tanimoto similarity \citep{Bender2004MolecularSA} on Morgan fingerprints (ECFP4,  radius~2, 2048-bit) \citep{rogers2010extended}. A pair $(x,x')$ is directed from seed $x$ to target $x'$ when the two molecules are structurally similar and $x'$ has higher logP than $x$. We use this fingerprint-based pairing on ZINC-250K because it provides dense local neighborhoods and many improving examples per seed, which is useful for learning SMILES syntax and local property-improving edits before downstream fine-tuning.

For MoleculeNet fine-tuning and evaluation, we instead construct directed matched molecular pairs (MMPs), which share a common core and differ by a localized fragment replacement \citep{kramer2014matched}. For the binary BACE, BBBP, and HIV tasks, pairs are directed from inactive to active molecules, so the seed is class~0 and the target is class~1. We use MMPs on MoleculeNet to emphasize localized medicinal-chemistry-style edits in the low-data downstream setting, whereas fingerprint similarity is a broader notion of locality and can be affected by fingerprint bit density and molecular size \citep{wang2007apparent}. We invoke \texttt{mmpdb fragment} and \texttt{mmpdb index}~\citep{dalke2018mmpdb} with default parameters (single-bond cuts), then post-filter to pairs whose seed and target have a heavy-atom ratio $\min(|x|,|x'|)/\max(|x|,|x'|) \geq 0.6$ to avoid pairing a small seed with a much larger target (or vice versa).

For the two- and ten-property signed-prompt models, MoleculeNet
fine-tuning instead includes both orderings of every retained training pair,
as in the augmented single-property training construction. Property changes
are recomputed after swapping seed and target: a nonzero change becomes its
negative, whereas a zero change remains zero. This supplies supervision for
both signs of observed changes, without requiring every bin to be populated.
For signed-prompt ZINC pretraining, we retain the original directed
nearest-neighbor pairs without a positive-change filter; the reverse of every
pair is not explicitly appended.

\begin{table}[!htbp]
    \centering
    \caption{Dataset statistics for fine-tuning and evaluation. Usable pairs are direction-improving transformations retained after filtering and used for
    fine-tuning. BACE, BBBP, and HIV use matched molecular pairs. For hERG, MMPs are too sparse (41 direction-improving training pairs), so we use Tanimoto-neighbor pairs, yielding 76 usable pairs after filtering. Test seeds are class-0 molecules for MoleculeNet and class-1 molecules (blockers) for hERG}
    \label{tab:num_pairs}
    \begin{tabular}{@{}lcccc@{}}
        \toprule
        & \textbf{BACE} & \textbf{BBBP} & \textbf{HIV} & \textbf{hERG} \\
        & \emph{Train/Test} & \emph{Train/Test} & \emph{Train/Test} & \emph{Train/Test} \\
        \midrule
        Molecules      & 798 / 267   & 517 / 175 & 1,420 / 475 & 414 / 111 \\
        Pairs          & 2,546 / 217 & 512 / 40  & 1,009 / 165 & 2,865 / 764 \\
        Usable pairs   & 662 / 51    & 88 / 14   & 256 / 46    & 76 / -- \\
        Test seeds     & -- / 135    & -- / 88   & -- / 231    & -- / 85 \\
        \bottomrule
    \end{tabular}
\end{table}

\clearpage

\subsection{Evaluation metrics}
\label{appendix:metrics}
Additional distributional metrics reported in
Tables~\ref{tab:bace_real-only_best64_appendix}--\ref{tab:hiv_real-only_best64_appendix} are defined below.

\textbf{Tanimoto similarity to seed (SimSeed, $\uparrow$).}
Mean Tanimoto similarity between generated molecules and their input seeds,
computed using Morgan fingerprints. Higher values indicate greater structural
fidelity to the seed.

\textbf{Fr\'echet ChemNet Distance (FCD, $\downarrow$).}
Distance between the distributions of generated molecules and reference molecules in the ChemNet feature space \citep{preuer2018frechet}. Lower values indicate that the generated set more closely matches the reference distribution.

\textbf{Similarity to nearest neighbor (SNN, $\uparrow$).}
Mean Tanimoto similarity between each novel valid generated molecule and its
nearest neighbor in the target-class training set, computed using Morgan
fingerprints. Higher values indicate that generated molecules lie closer to the reference distribution in fingerprint space.

\textbf{Fragment Cosine ($\uparrow$).}
Cosine similarity between fragment-frequency vectors computed on generated molecules and the reference set. Higher values indicate better agreement in fragment composition.

\textbf{Scaffold Cosine ($\uparrow$).}
Cosine similarity between Bemis--Murcko scaffold-frequency vectors of generated molecules and the reference set. Higher values indicate better agreement in scaffold usage.

\textbf{Coverage ($\uparrow$).}
Fraction of reference molecules covered by the generated set under the chosen descriptor-space neighborhood criterion. Higher values indicate broader support over the reference distribution.

\subsubsection{Property-oracle agreement}
\label{app:oracle_agreement}

To assess sensitivity to the choice of property oracle, we compare the
random-forest and MiniMol predictions on benchmark test molecules and on
generated molecules (Table~\ref{tab:oracle_agreement}). Agreement is
consistently higher on benchmark molecules than on generated molecules, with
the weakest generated-molecule agreement on HIV. For hERG, classification
agreement should be interpreted alongside the correlation metrics because the
benchmark test set is class-imbalanced.

\begin{table}[ht]
\centering
\footnotesize
\caption{Agreement between the random-forest and MiniMol property oracles.
Classification agreement is the fraction of molecules assigned the same
active/inactive label by both models.}
\label{tab:oracle_agreement}
\begin{tabular}{llcccc}
\toprule
& Metric & BACE & BBBP & HIV & hERG \\
\midrule
\multirow{4}{*}{Benchmark test}
& $n$ molecules & 261 & 172 & 475 & 111 \\
& Pearson $r$ & 0.791 & 0.819 & 0.576 & 0.783 \\
& Spearman $\rho$ & 0.787 & 0.825 & 0.616 & 0.685 \\
& Classification agreement (\%) & 83.5 & 84.9 & 69.7 & 87.6 \\
\midrule
\multirow{4}{*}{Generated}
& $n$ molecules & 1618 & 1389 & 3751 & 1353 \\
& Pearson $r$ & 0.438 & 0.631 & 0.290 & 0.595 \\
& Spearman $\rho$ & 0.424 & 0.669 & 0.290 & 0.635 \\
& Classification agreement (\%) & 63.0 & 64.7 & 51.9 & 71.8 \\
\bottomrule
\end{tabular}
\end{table}

\subsection{Baseline methods}
\label{appendix:baselines}

All baselines use the same task splits, task directions, metrics, and final selection procedure as PairIT; below we report method-specific implementation details.

\textbf{NF-BO.}
We compare against NF-BO, a normalizing-flow-based Bayesian optimization method that combines a SELFIES-based autoregressive generative model with a Gaussian-process surrogate and trust-region Bayesian optimization. The generative model is an LSTM normalizing flow trained on SELFIES sequences, pretrained on GuacaMol~\citep{brown2019guacamol}, and fine-tuned separately for each task. Models were fine-tuned for 50 epochs using Adam with learning rate $10^{-4}$ and batch size 32.

For BACE, the Random Forest classifier returns $p_{\text{class 1}} = 0$ for every input class-0 test seed; this drives the initial Gaussian-process fit to a degenerate 1$\times$1 kernel with NaN Cholesky, halting BO before any candidate is written. We therefore run NF-BO's inner BO loop with the Random Forest classifier trained on the full dataset, including training and test sets, for BACE only; this provides sufficient variation in the surrogate scores for BO to proceed. All 64 generations per BACE seed are otherwise produced by the same NF-BO procedure, and are re-scored downstream with the shared MiniMol oracle at evaluation time. BBBP and HIV use the correct train-only Random Forest classifier throughout.

At inference time, NF-BO performs iterative Bayesian optimization in the learned latent space with a property model budget of 500 evaluations per seed molecule. A deep-kernel Gaussian-process surrogate is trained for 5 epochs at each iteration on accumulated observations, and candidate molecules are proposed using TuRBO-based Thompson sampling with acquisition batch size 100; the top-10 current observations are used as trust-region centers at each step, and a running top-64 set of valid generations is maintained across the budget. Generated molecules are decoded via SELFIES and filtered for validity. Optimization objectives maximize either logP improvement or predicted class-1 probability, subject to a structural similarity constraint enforced via Tanimoto similarity (Morgan fingerprints, radius 2, 2048 bits) with a minimum threshold of 0.4 and a soft penalty below that threshold. Generation terminates upon exhausting the property model budget.

\textbf{DDSBM (Bridge and SB variants).}
We compare against DDSBM, a discrete diffusion--based Schr\"odinger Bridge model for graph-to-graph molecular transformation. DDSBM operates directly on molecular graphs and learns stochastic forward and reverse diffusion processes that transport molecules from a source distribution to a target distribution using entropy-regularized optimal transport. The model is implemented with a graph transformer backbone.

We evaluate two DDSBM variants that differ only in the number of Schr\"odinger Bridge refinement steps. \emph{DDSBM-Bridge} performs a single forward training pass from source to target (one outer loop), providing a fast but coarse approximation of the bridge. \emph{DDSBM-SB} applies 5 outer iterative-proportional-fitting loops, corresponding to 10 training runs alternating backward and forward updates, yielding a tighter approximation to the true Schr\"odinger bridge at increased computational cost. Both variants are fine-tuned from a ZINC-250K paired pretraining checkpoint (Bridge and SB pretrained checkpoints are kept separate). Each training run is performed for 300 epochs, resulting in 300 and 3,000 effective training epochs for Bridge and SB, respectively.

At inference time, generation is performed via 100-step discrete reverse diffusion from the input molecule without explicit similarity constraints or property model budgets. We generate 64 stochastic samples per seed and evaluate generations using the metrics described in Appendix~\ref{appendix:metrics}.

\textbf{HierG2G.}
We compare against Hierarchical Graph-to-Graph (HierG2G), a variational graph translation model that generates molecules by hierarchically assembling substructures into molecular graphs. HierG2G represents molecules using coupled tree and graph representations and performs conditional translation via a hierarchical variational graph neural network. We use the public HierG2G implementation and fine-tune models separately for each task, starting from a checkpoint pretrained on ZINC-250K.

For MoleculeNet tasks (BACE, BBBP, and HIV), we fine-tune task-specific translation models for 12 epochs using Adam (learning rate $10^{-3}$, batch size 32, latent dimension 4), following default hyperparameters from prior work. At inference time, conditional translation is performed by sampling multiple candidate molecules per input via stochastic decoding in the learned latent space, with an increased maximum decoding length to reduce fragment generation. We generate 64 candidates per seed molecule using the checkpoint at epoch 11 and evaluate generations using the metrics described in Appendix~\ref{appendix:metrics}.

\textbf{MMP baseline.}
We include a property-aware matched molecular pair baseline (\textsc{MMP-Prop}), a non-learning cheminformatics enumeration method. Starting from each seed molecule, we queried an \texttt{mmpdb} database of matched molecular transformations built from the training set. We augmented the database with training-set property scores from a task-specific classifier, so that each transformation carried an estimated property shift. For each test seed, we first enumerated candidate edits using \texttt{mmpdb transform --property}, which ranks transformations by their learned average property change; when needed, we supplemented the candidate pool with additional \texttt{mmpdb generate} proposals. This produced up to 64 candidate molecules per seed. As in the other 64-sample baselines, we then evaluated the candidate pool with the common benchmark metrics and reported the best-5 subset under the same selection procedure used for all methods.

\textbf{GraphDiT.}
We used the publicly released pretrained GraphDiT checkpoints for the MoleculeNet datasets (BACE, BBBP, and HIV) as released by the original authors, without any retraining or fine-tuning on our side. The pretrained models use dropout for classifier-free guidance regularization during training ($\texttt{drop\_condition}=0.01$). For generation, we sampled 10{,}000 molecules per dataset using classifier-free guidance with $\texttt{guide\_scale}=2$. These generations are de novo samples from random noise and are therefore not conditioned on an input molecule.

 \textbf{GeLLM4O-C.}                                      
  We compare against GeLLM4O-C~\citep{dey2025large}, an instruction-tuned
  7B-parameter Mistral LLM (Mistral-7B-Instruct-v0.3) trained with LoRA
  adapters on the C-MuMOInstruct multi-property optimization corpus. We use
  the authors' released BBBP, logP, and QED adapters untouched. For BACE and
  HIV, we fine-tune task-specific LoRA adapters (r$=16$, $\alpha=16$, dropout
  $0.05$, targeting q/k/v/o/gate/up/down projections and the LM head; 0.58\%
  trainable parameters, no quantization) using their trainer on directed
  \texttt{mmpdb} pairs converted to C-MuMOInstruct JSON format
  (\texttt{source} = inactive molecule, \texttt{target} = active molecule;
  direction encoded categorically as "increase" via the instruction template,
  with numeric activity from the RF oracle not surfaced in the prompt). Models were trained with AdamW at learning rate $10^{-4}$, cosine schedule,
warmup ratio 0.1, effective batch size 16 (per-device 1 with 16-step gradient
accumulation), and \texttt{cutoff\_len=4096}. Training uses \texttt{train\_on\_inputs=False} so loss is computed only on the response
  SMILES. At inference, we sample 64 candidates per seed via stochastic
  decoding and evaluate under the shared selection protocol described in
  Appendix~\ref{appendix:metrics}.

\clearpage

\section{Theoretical Comparison to DPO}
\label{app:theory}

Because PairIT is trained on paired examples, a preference objective such as
DPO~\citep{rafailov2024direct} is a natural alternative. We instead use
conditional likelihood: prior work has found DPO to underperform SFT and
SFT followed by RL on related biomolecular sequence-optimization tasks
~\citep{chen2024generalists}, and DPO's contrastive objective can shift
probability mass toward outputs outside the observed preference pair
~\citep{razin2025unintentional}. Below, we provide a complementary analytical
comparison of the training signals produced by the two objectives.

We compare DPO's pair-level parameter gradient to an aggregate tokenwise supervision proxy for PairIT as matched molecules become more different. The comparison is intentionally asymmetric: DPO naturally produces a single sequence-level gradient through one sigmoid margin term, whereas PairIT distributes supervision across multiple edited token positions. Because token-level parameter gradients in PairIT can partially cancel when summed in parameter space, we track an aggregate tokenwise proxy rather than the norm of the full PairIT parameter gradient.

Let $x$ denote the seed molecule, $x'$ the matched target molecule, and $c$ the conditioning information. PairIT is trained with
\begin{equation}
\mathcal{L}_{\mathrm{PairIT}}(\theta)
=
-\log q_\theta(x' \mid x,c)
=
\sum_{t=1}^{T}
\ell_t(\theta),
\qquad
\ell_t(\theta)
:=
-\log q_\theta(x'_t \mid x'_{<t},x,c).
\end{equation}

That is, PairIT uses autoregressive cross-entropy with the loss computed only on the target tokens of the second molecule $x'$, conditioned on the seed molecule $x$ and the control tokens $c$.

Note that
\begin{equation}
\nabla_\theta \mathcal{L}_{\mathrm{PairIT}}(\theta)
=
\sum_{t=1}^{T}\nabla_\theta \ell_t(\theta).
\end{equation}
These token-level gradients are vector-valued and live in the same shared parameter space, so they may partially cancel. For this reason, we will compare DPO's pair-level parameter gradient to a tokenwise supervision proxy for PairIT rather than to $\|\nabla_\theta \mathcal{L}_{\mathrm{PairIT}}\|$ directly.

Let $d_{\mathrm{tok}}(x,x')$ denote a token edit distance between $x$ and $x'$, and let
$E(x,x') \subseteq \{1,\dots,T\}$ denote the set of target positions affected by these edits.
We assume there exists a constant $c_{\mathrm{edit}} > 0$ such that $
|E(x,x')| \ge c_{\mathrm{edit}}\, d_{\mathrm{tok}}(x,x').$
That is, the number of target positions affected by the edits grows at least linearly with the token edit distance.


For DPO, we use the same conditioning input $(x,c)$ and a preference pair $(x^+,x^-)$, where $x^+$ is the preferred output and $x^-$ is the less preferred output. In the matched-pair setting, $x^+$ corresponds to the improved target molecule $x'$.

Let $z_t(\theta)\in\mathbb{R}^{V}$ denote the decoder logits at target position $t$, where $V$ is the vocabulary size, and let
\begin{equation}
p_t(\theta) := \mathrm{softmax}(z_t(\theta)).
\end{equation}
We measure the PairIT supervision strength by the following aggregate logit-space tokenwise supervision proxy:
\begin{equation}
S_{\mathrm{PairIT}}^{\mathrm{logit}}(\theta; x,x',c)
:=
\sum_{t \in E(x,x')}
\bigl\|\nabla_{z_t}\ell_t(\theta)\bigr\|_2.
\end{equation}
Assume:

\begin{enumerate}

    \item \textbf{Non-saturated edited tokens.}
There exists $p_{\max}<1$ such that for every edited position $t \in E(x,x')$,
\begin{equation}
p_t(\theta)_{x'_t} \le p_{\max}.
\end{equation}
That is, on edited positions, the model is not already assigning probability arbitrarily close to $1$ to the correct target token.

    \item \textbf{Bounded score gradients for DPO.}
    There exists $M>0$ such that
    \begin{equation}
    \bigl\|\nabla_\theta \log q_\theta(z \mid x,c)\bigr\| \le M
    \qquad
    \text{for all relevant } (x,c,z).
    \end{equation}

    \item \textbf{Margin growth with pair difference.}
    For a pair consisting of a more preferred output $x^+$ and a less preferred output $x^-$ under the same conditioning $(x,c)$, define
    \begin{equation}
    Z_\theta
    :=
    \log q_\theta(x^+ \mid x,c)
    -
    \log q_\theta(x^- \mid x,c)
    -
    \log q_0(x^+ \mid x,c)
    +
    \log q_0(x^- \mid x,c).
    \end{equation}
    Assume there exists $\alpha>0$ such that
    \begin{equation}
    Z_\theta \ge \alpha\, d_{\mathrm{tok}}(x^+,x^-).
    \end{equation}
\end{enumerate}



For cross-entropy, the gradient with respect to the logits at position $t$ is
\begin{equation}
\nabla_{z_t}\ell_t(\theta)
=
p_t(\theta) - e_{x'_t},
\end{equation}
where $e_{x'_t}\in\mathbb{R}^{V}$ is the one-hot vector of the target token $x'_t$.
Therefore,
\begin{align}
\bigl\|\nabla_{z_t}\ell_t(\theta)\bigr\|_2
&=
\|p_t(\theta)-e_{x'_t}\|_2 \\
&\ge
\bigl|(p_t(\theta)-e_{x'_t})_{x'_t}\bigr| \\
&=
1 - p_t(\theta)_{x'_t} \\
&\ge
1-p_{\max}
\qquad \text{for every } t\in E(x,x').
\end{align}
Summing over edited positions gives
\begin{equation}
S_{\mathrm{PairIT}}^{\mathrm{logit}}(\theta; x,x',c)
\ge
(1-p_{\max})\,|E(x,x')|.
\end{equation}
Using $|E(x,x')| \ge c_{\mathrm{edit}}\, d_{\mathrm{tok}}(x,x')$, we obtain
\begin{equation}
S_{\mathrm{PairIT}}^{\mathrm{logit}}(\theta; x,x',c)
\ge
(1-p_{\max})\,c_{\mathrm{edit}}\, d_{\mathrm{tok}}(x,x').
\end{equation}
Thus, the aggregate logit-space tokenwise supervision proxy grows at least linearly with the sequence difference between the matched molecules.

For scalar ordered properties, define the matched-pair property gap as
\begin{equation}
\Delta y_{\mathrm{pair}} := g(x') - g(x) > 0.
\end{equation}
If, in addition, the property gap is controlled by edit size,
\begin{equation}
\Delta y_{\mathrm{pair}} \le \ell\, d_{\mathrm{tok}}(x,x'),
\end{equation}
then the PairIT bound implies
\begin{equation}
S_{\mathrm{PairIT}}^{\mathrm{logit}}(\theta; x,x',c)
\ge
\frac{(1-p_{\max})\,c_{\mathrm{edit}}}{\ell}\,\Delta y_{\mathrm{pair}}.
\end{equation}

For the DPO preference pair $(x^+,x^-)$, the DPO loss is
\begin{equation}
\mathcal{L}_{\mathrm{DPO}}(\theta)
=
-\log \sigma(\beta Z_\theta),
\end{equation}
where $\sigma$ is the sigmoid function and $\beta>0$ is the DPO inverse-temperature parameter. Since $q_0$ is fixed,
\begin{equation}
\nabla_\theta \mathcal{L}_{\mathrm{DPO}}(\theta)
=
-\beta\,\sigma(-\beta Z_\theta)
\left[
\nabla_\theta \log q_\theta(x^+ \mid x,c)
-
\nabla_\theta \log q_\theta(x^- \mid x,c)
\right].
\end{equation}
Using the bounded score-gradient assumption,
\begin{equation}
\|\nabla_\theta \mathcal{L}_{\mathrm{DPO}}\|
\le
2M\beta\,e^{-\beta Z_\theta}.
\end{equation}

For the DPO preference pair, define
\begin{equation}
\Delta y_{\mathrm{DPO}} := g(x^+) - g(x^-) > 0.
\end{equation}
Likewise, if the DPO margin grows with property gap, $Z_\theta \ge \alpha \Delta y_{\mathrm{DPO}}$, then
\begin{equation}
\|\nabla_\theta \mathcal{L}_{\mathrm{DPO}}\|
\le
2M\beta\, e^{-\beta\alpha \Delta y_{\mathrm{DPO}}}.
\end{equation}

\textbf{Takeaway.}
PairIT uses autoregressive cross-entropy on matched pairs, with the loss computed on the target tokens of the second molecule. As matched molecules become more different, more target positions are affected, so PairIT receives supervision at more edited token positions. This yields an aggregate tokenwise supervision proxy that grows at least linearly with sequence difference. Under the margin-growth assumption above, DPO instead applies a single sigmoid margin term to the entire pair, and the corresponding pair-level parameter gradient becomes exponentially small as the learned preference margin grows. Thus, PairIT is less prone to this particular sequence-level saturation effect.

\clearpage

\section{Additional seeded-generation results and ablations}
\label{app:additional_seeded_results}

\subsection{Detailed MoleculeNet results}
\label{app:full_tables}
\begin{figure}[!htbp]
  \centering
  \includegraphics[width=\textwidth]{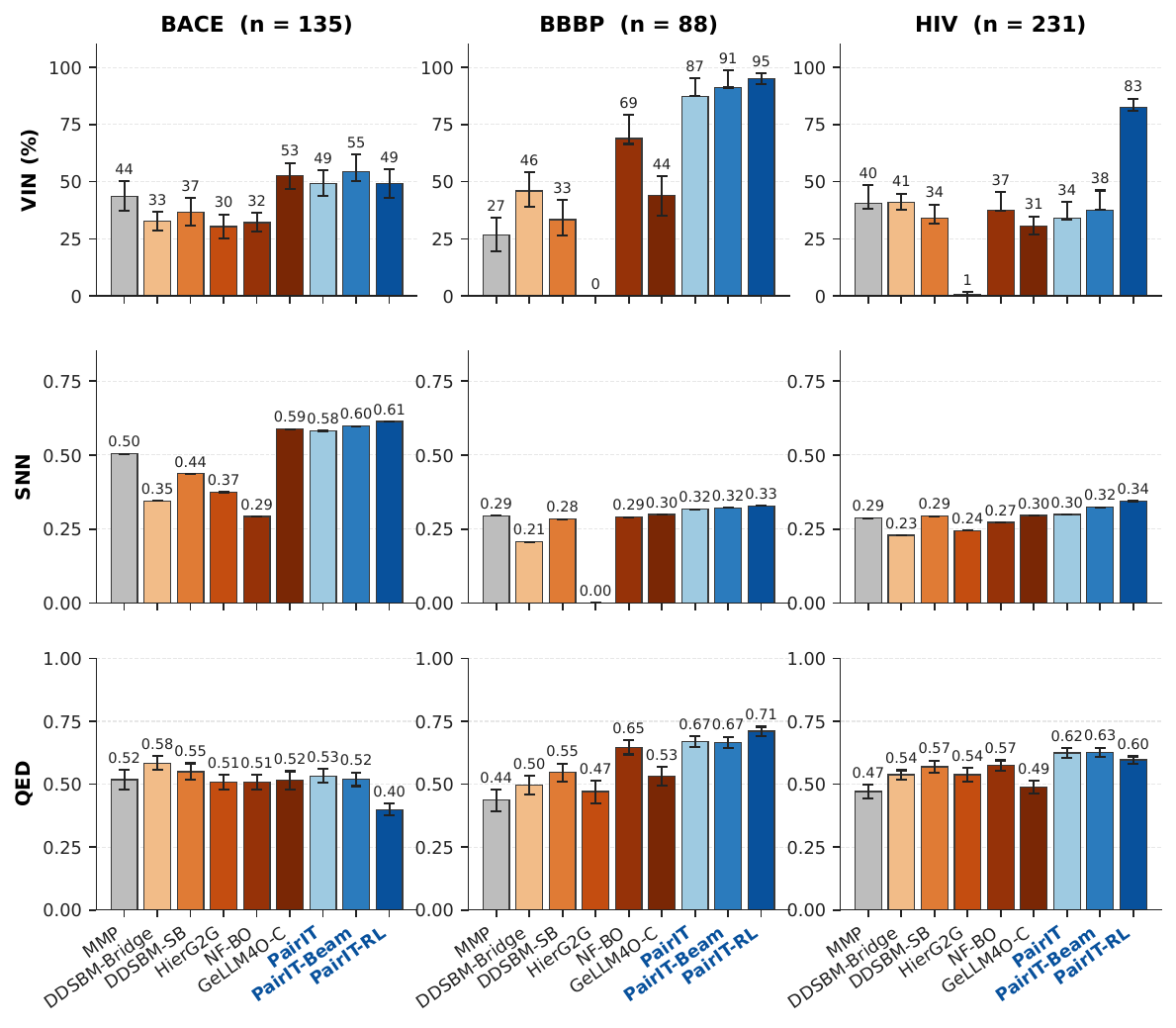}
  \caption{Aggregate MoleculeNet seeded-generation metrics corresponding to Figure~\ref{fig:main_results}. Bars show mean performance across test seeds after within-seed de-duplication and final top-5 selection; error bars are bootstrap 95\% confidence intervals over seeds. VIN is the fraction of retained generations that are valid, novel, and improved. Similarity to nearest neighbor (SNN) is computed over novel and valid generations only. QED is the quantitative estimate of drug-likeness.}
  \label{fig:main_results_vin_simseed_bars}
\end{figure}

Figure~\ref{fig:main_results_vin_simseed_bars} reports aggregate VIN, SimSeed, and QED with bootstrap confidence intervals. Tables~\ref{tab:bace_real-only_best64_main}--\ref{tab:hiv_real-only_best64_main} provide the corresponding numeric results. Entries are means across seeds after within-seed canonical-SMILES de-duplication and final top-5 selection; metrics are defined in Appendix~\ref{appendix:metrics}.

\begin{table*}[t]
\centering
\footnotesize
\compacttablesetup
\caption{MoleculeNet BACE bioactivity improvement ($0 \rightarrow 1$) under seeded generation. Best 5 of 64 per seed selected after within-seed canonical-SMILES de-duplication, so each retained molecule is distinct within its seed. Per-seed bootstrap 95\% CIs for VIN and SNN are shown in Fig.~\ref{fig:main_results_vin_simseed_bars}. SNN is the mean per-generation Tanimoto to the nearest class-1 training molecule, computed over novel valid generations only.}
\label{tab:bace_real-only_best64_main}
\begin{tabular}{lcccccccc}
\toprule
Model & VIN & QED & Valid & Novel & Unique & Diverse & Improved & SNN \\
\midrule
MMP
& \onescore{43.56}
& \onescore{0.52}
& \onescore{100.00}
& \onescore{99.11}
& \onescore{100.00}
& \onescore{38.00}
& \onescore{44.30}
& \onescore{0.50} \\
  
GraphDiT
& \onescore{--}
& \onescore{--}
& \onescore{87.32}
& \onescore{100.00}
& \onescore{11.28}
& \onescore{--}
& \onescore{86.60}
& \onescore{--} \\
\midrule
DDSBM-Bridge
& \onescore{32.74}
& \onescore{\textbf{0.58}}
& \onescore{\textbf{100.00}}
& \onescore{\underline{99.85}}
& \onescore{\textbf{100.00}}
& \onescore{\textbf{79.00}}
& \onescore{32.89}
& \onescore{0.35} \\
DDSBM-SB
& \onescore{36.59}
& \onescore{\underline{0.55}}
& \onescore{92.89}
& \onescore{97.77}
& \onescore{92.15}
& \onescore{46.00}
& \onescore{41.16}
& \onescore{0.44} \\
HierG2G
& \onescore{30.37}
& \onescore{0.50}
& \onescore{79.85}
& \onescore{93.69}
& \onescore{74.81}
& \onescore{62.00}
& \onescore{40.59}
& \onescore{0.37} \\
NF-BO
& \onescore{32.15}
& \onescore{0.51}
& \onescore{\textbf{100.00}}
& \onescore{\textbf{100.00}}
& \onescore{\textbf{100.00}}
& \onescore{\underline{74.00}}
& \onescore{32.15}
& \onescore{0.29} \\
GeLLM4O-C
& \onescore{\underline{52.59}}
& \onescore{0.51}
& \onescore{\underline{96.89}}
& \onescore{92.35}
& \onescore{\underline{96.74}}
& \onescore{42.00}
& \onescore{\underline{60.34}}
& \onescore{0.59} \\
\midrule
\textbf{PairIT}
& \onescore{49.19}
& \onescore{0.54}
& \onescore{96.74}
& \onescore{87.90}
& \onescore{96.00}
& \onescore{54.00}
& \onescore{58.80}
& \onescore{0.58} \\
\textbf{PairIT-Beam}
& \onescore{\textbf{54.52}}
& \onescore{0.53}
& \onescore{94.52}
& \onescore{88.71}
& \onescore{94.52}
& \onescore{51.00}
& \onescore{\textbf{65.52}}
& \onescore{0.60} \\
\textbf{PairIT-RL}
& \onescore{49.33}
& \onescore{0.39}
& \onescore{92.30}
& \onescore{\textbf{100.00}}
& \onescore{92.30}
& \onescore{27.00}
& \onescore{53.45}
& \onescore{0.61} \\
\bottomrule
\end{tabular}
\end{table*}

\begin{table*}[t]
\centering
\footnotesize
\compacttablesetup
\caption{MoleculeNet BBBP bioactivity improvement ($0 \rightarrow 1$) under seeded generation. Best 5 of 64 per seed selected after within-seed canonical-SMILES de-duplication, so each retained molecule is distinct within its seed. Per-seed bootstrap 95\% CIs for VIN and SNN are shown in Fig.~\ref{fig:main_results_vin_simseed_bars}. SNN is the mean per-generation Tanimoto to the nearest class-1 training molecule, computed over novel valid generations only.}
\label{tab:bbbp_real-only_best64_main}
\begin{tabular}{lcccccccc}
\toprule
Model & VIN & QED & Valid & Novel & Unique & Diverse & Improved & SNN \\
\midrule
MMP
& \onescore{26.59}
& \onescore{0.44}
& \onescore{99.55}
& \onescore{100.00}
& \onescore{99.55}
& \onescore{43.00}
& \onescore{26.71}
& \onescore{0.29} \\

GraphDiT
& \onescore{--}
& \onescore{--}
& \onescore{83.93}
& \onescore{100.00}
& \onescore{20.33}
& \onescore{--}
& \onescore{82.30}
& \onescore{--} \\
\midrule
DDSBM-Bridge
& \onescore{45.91}
& \onescore{0.51}
& \onescore{91.14}
& \onescore{\textbf{100.00}}
& \onescore{91.14}
& \onescore{\textbf{88.00}}
& \onescore{50.37}
& \onescore{0.21} \\
DDSBM-SB
& \onescore{33.41}
& \onescore{0.55}
& \onescore{88.86}
& \onescore{95.91}
& \onescore{85.91}
& \onescore{57.00}
& \onescore{39.68}
& \onescore{0.28} \\
HierG2G
& \onescore{0.00}
& \onescore{0.47}
& \onescore{20.00}
& \onescore{0.00}
& \onescore{0.00}
& \onescore{--}
& \onescore{0.00}
& \onescore{--} \\
NF-BO
& \onescore{68.86}
& \onescore{0.65}
& \onescore{90.45}
& \onescore{98.74}
& \onescore{89.32}
& \onescore{70.00}
& \onescore{77.10}
& \onescore{0.29} \\
GeLLM4O-C
& \onescore{43.86}
& \onescore{0.53}
& \onescore{\textbf{100.00}}
& \onescore{\textbf{100.00}}
& \onescore{\textbf{100.00}}
& \onescore{50.00}
& \onescore{43.86}
& \onescore{0.30} \\
\midrule
\textbf{PairIT}
& \onescore{87.27}
& \onescore{\underline{0.67}}
& \onescore{94.09}
& \onescore{\underline{99.76}}
& \onescore{93.86}
& \onescore{\underline{78.00}}
& \onescore{92.98}
& \onescore{0.32} \\
\textbf{PairIT-Beam}
& \onescore{\underline{90.91}}
& \onescore{0.67}
& \onescore{94.32}
& \onescore{\textbf{100.00}}
& \onescore{94.32}
& \onescore{73.00}
& \onescore{\textbf{96.39}}
& \onescore{0.32} \\
\textbf{PairIT-RL}
& \onescore{\textbf{95.00}}
& \onescore{\textbf{0.71}}
& \onescore{\underline{99.55}}
& \onescore{\textbf{100.00}}
& \onescore{\underline{99.55}}
& \onescore{66.00}
& \onescore{\underline{95.43}}
& \onescore{0.33} \\
\bottomrule
\end{tabular}
\end{table*}

\begin{table*}[t]
\centering
\footnotesize
\compacttablesetup
\caption{MoleculeNet HIV bioactivity improvement ($0 \rightarrow 1$) under seeded generation. Best 5 of 64 per seed selected after within-seed canonical-SMILES de-duplication, so each retained molecule is distinct within its seed. Per-seed bootstrap 95\% CIs for VIN and SNN are shown in Fig.~\ref{fig:main_results_vin_simseed_bars}. SNN is the mean per-generation Tanimoto to the nearest class-1 training molecule, computed over novel valid generations only.}
\label{tab:hiv_real-only_best64_main}
\begin{tabular}{lcccccccc}
\toprule
Model & VIN & QED & Valid & Novel & Unique & Diverse & Improved & SNN \\
\midrule
MMP
& \onescore{40.43}
& \onescore{0.47}
& \onescore{92.03}
& \onescore{99.91}
& \onescore{91.95}
& \onescore{41.00}
& \onescore{43.97}
& \onescore{0.29} \\

GraphDiT
& \onescore{--}
& \onescore{--}
& \onescore{80.16}
& \onescore{100.00}
& \onescore{16.55}
& \onescore{--}
& \onescore{95.55}
& \onescore{--} \\
\midrule
DDSBM-Bridge
& \onescore{\underline{40.95}}
& \onescore{0.54}
& \onescore{\textbf{97.66}}
& \onescore{\underline{99.91}}
& \onescore{\textbf{97.66}}
& \onescore{\textbf{85.00}}
& \onescore{42.02}
& \onescore{0.23} \\
DDSBM-SB
& \onescore{33.94}
& \onescore{0.56}
& \onescore{87.53}
& \onescore{97.03}
& \onescore{86.15}
& \onescore{61.00}
& \onescore{40.70}
& \onescore{0.29} \\
HierG2G
& \onescore{0.69}
& \onescore{0.53}
& \onescore{21.04}
& \onescore{7.00}
& \onescore{1.47}
& \onescore{71.00}
& \onescore{\underline{47.06}}
& \onescore{0.24} \\
NF-BO
& \onescore{37.40}
& \onescore{0.57}
& \onescore{87.01}
& \onescore{98.31}
& \onescore{85.54}
& \onescore{70.00}
& \onescore{43.72}
& \onescore{0.27} \\
GeLLM4O-C
& \onescore{30.74}
& \onescore{0.48}
& \onescore{95.76}
& \onescore{88.16}
& \onescore{92.73}
& \onescore{56.00}
& \onescore{41.55}
& \onescore{0.30} \\
\midrule
\textbf{PairIT}
& \onescore{33.94}
& \onescore{\textbf{0.63}}
& \onescore{87.79}
& \onescore{99.70}
& \onescore{87.53}
& \onescore{\underline{73.00}}
& \onescore{38.77}
& \onescore{0.30} \\
\textbf{PairIT-Beam}
& \onescore{37.75}
& \onescore{\underline{0.63}}
& \onescore{87.36}
& \onescore{\textbf{100.00}}
& \onescore{87.36}
& \onescore{64.00}
& \onescore{43.21}
& \onescore{0.32} \\
\textbf{PairIT-RL}
& \onescore{\textbf{82.60}}
& \onescore{0.60}
& \onescore{\underline{97.06}}
& \onescore{99.82}
& \onescore{\underline{96.88}}
& \onescore{52.00}
& \onescore{\textbf{85.25}}
& \onescore{0.34} \\
\bottomrule
\end{tabular}
\end{table*}

\FloatBarrier

\subsection{Distribution-level results}
\label{app:distribution_metrics}

Tables~\ref{tab:bace_real-only_best64_appendix}--\ref{tab:hiv_real-only_best64_appendix} report distribution-level metrics comparing generated molecules to the class-1 train reference set, as the test set is too small to provide a stable comparison and would result in high variance in the estimated metrics. Definitions of these metrics are given in Appendix~\ref{appendix:metrics}.

\begin{table*}[!htbp]
\centering
\caption{MoleculeNet BACE bioactivity improvement ($0 \rightarrow 1$): distribution metrics on best-5-of-64 real-only selection.}
\label{tab:bace_real-only_best64_appendix}
\begin{adjustbox}{max width=\textwidth}
\begin{tabular}{lccccc}
\toprule
Model & SimSeed$\uparrow$ & FCD$\downarrow$ & FragCos$\uparrow$ & ScafCos$\uparrow$ & Cov$\uparrow$ \\
\midrule
MMP & 0.71 & 10.80 & 0.92 & 0.54 & 1.00 \\
DDSBM-Bridge & 0.22 & 10.32 & 0.85 & 0.46 & 1.00 \\
DDSBM-SB & 0.44 & 10.16 & 0.90 & 0.41 & 1.00 \\
HierG2G & 0.33 & 15.56 & 0.89 & 0.38 & 1.00 \\
NF-BO & 0.30 & 21.19 & 0.88 & 0.17 & 1.00 \\
GeLLM4O-C & 0.54 & 7.16 & 0.96 & 0.71 & 1.00 \\
\textbf{PairIT} & 0.37 & 9.27 & 0.96 & 0.61 & 1.00 \\
\textbf{PairIT-Beam} & 0.31 & 10.29 & 0.93 & 0.65 & 1.00 \\
\textbf{PairIT-RL} & 0.24 & 27.47 & 0.88 & 0.19 & 1.00 \\
\bottomrule
\end{tabular}
\end{adjustbox}
\end{table*}

\begin{table*}[!htbp]
\centering
\caption{MoleculeNet BBBP bioactivity improvement ($0 \rightarrow 1$): distribution metrics on best-5-of-64 real-only selection.}
\label{tab:bbbp_real-only_best64_appendix}
\begin{adjustbox}{max width=\textwidth}
\begin{tabular}{lccccc}
\toprule
Model & SimSeed$\uparrow$ & FCD$\downarrow$ & FragCos$\uparrow$ & ScafCos$\uparrow$ & Cov$\uparrow$ \\
\midrule
MMP & 0.67 & 24.07 & 0.88 & 0.54 & 0.89 \\
DDSBM-Bridge & 0.12 & 17.61 & 0.78 & 0.51 & 0.89 \\
DDSBM-SB & 0.39 & 19.13 & 0.83 & 0.50 & 0.89 \\
HierG2G & -- & -- & -- & -- & -- \\
NF-BO & 0.27 & 12.90 & 0.91 & 0.74 & 0.89 \\
GeLLM4O-C & 0.56 & 21.06 & 0.89 & 0.64 & 0.89 \\
\textbf{PairIT} & 0.17 & 16.49 & 0.88 & 0.72 & 0.78 \\
\textbf{PairIT-Beam} & 0.12 & 19.90 & 0.87 & 0.71 & 0.78 \\
\textbf{PairIT-RL} & 0.23 & 17.75 & 0.88 & 0.72 & 0.78 \\
\bottomrule
\end{tabular}
\end{adjustbox}
\end{table*}

\begin{table*}[!htbp]
\centering
\caption{MoleculeNet HIV bioactivity improvement ($0 \rightarrow 1$): distribution metrics on best-5-of-64 real-only selection.}
\label{tab:hiv_real-only_best64_appendix}
\begin{adjustbox}{max width=\textwidth}
\begin{tabular}{lccccc}
\toprule
Model & SimSeed$\uparrow$ & FCD$\downarrow$ & FragCos$\uparrow$ & ScafCos$\uparrow$ & Cov$\uparrow$ \\
\midrule
MMP & 0.67 & 9.44 & 0.93 & 0.36 & 0.73 \\
DDSBM-Bridge & 0.17 & 13.09 & 0.87 & 0.35 & 0.50 \\
DDSBM-SB & 0.33 & 10.82 & 0.91 & 0.41 & 0.50 \\
HierG2G & 0.28 & 45.42 & 0.51 & 0.05 & 0.32 \\
NF-BO & 0.34 & 8.89 & 0.94 & 0.48 & 0.55 \\
GeLLM4O-C & 0.50 & 9.19 & 0.93 & 0.38 & 0.73 \\
\textbf{PairIT} & 0.23 & 14.91 & 0.89 & 0.30 & 0.45 \\
\textbf{PairIT-Beam} & 0.17 & 19.64 & 0.83 & 0.17 & 0.41 \\
\textbf{PairIT-RL} & 0.24 & 18.87 & 0.61 & 0.37 & 0.32 \\
\bottomrule
\end{tabular}
\end{adjustbox}
\end{table*}

\FloatBarrier

\subsection{VIN--Similarity trade-offs}
\label{app:vin_simseed_scatter}

\begin{figure}[!htbp]
    \centering
    \includegraphics[width=\textwidth]{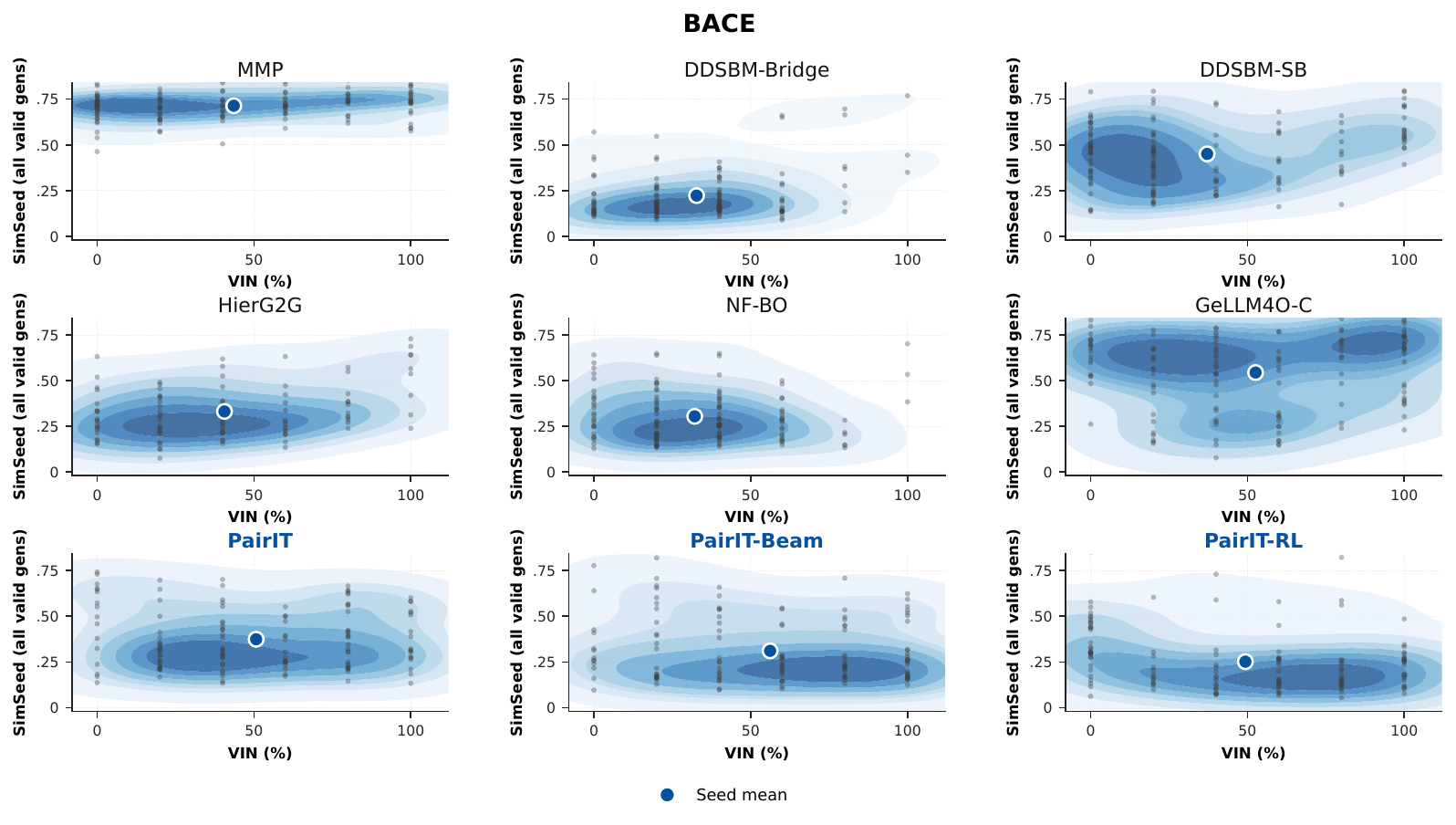}
    \caption{Scatter plots of VIN vs.\ SimSeed showing the distribution of all generations for all seeds on BACE.}
    \label{fig:scatter_per_seed_bace}
\end{figure}

\begin{figure}[!htbp]
    \centering
    \includegraphics[width=\textwidth]{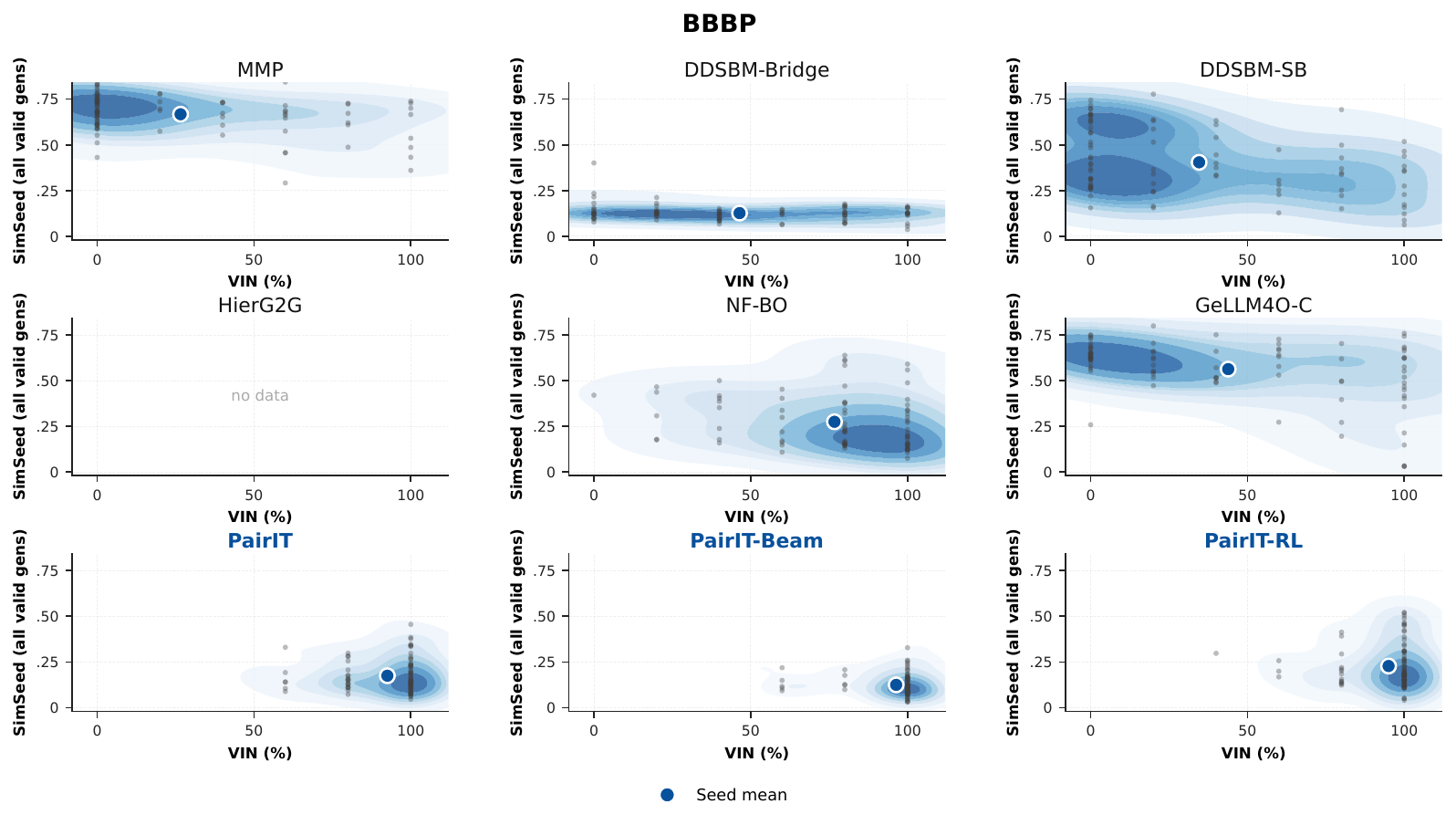}
    \caption{Scatter plots of VIN vs.\ SimSeed showing the distribution of all generations for all seeds on BBBP. HierG2G generations are excluded because they all contain duplicates of the seed.}
    \label{fig:scatter_per_seed_bbbp}
\end{figure}

\begin{figure}[!htbp]
    \centering
    \includegraphics[width=\textwidth]{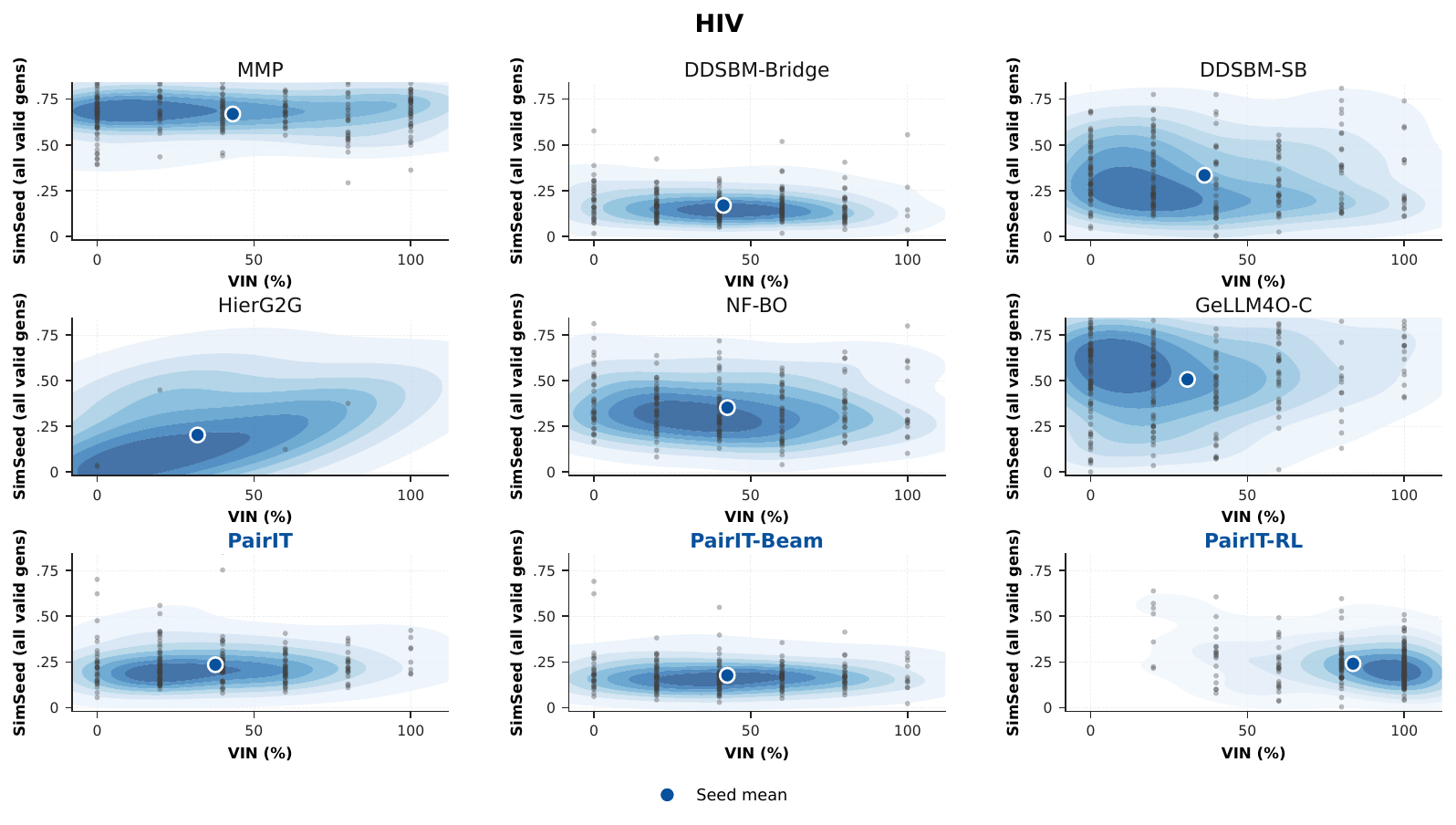}
    \caption{Scatter plots of VIN vs.\ SimSeed showing the distribution of all generations for all seeds on HIV. The majority of HierG2G generations contain duplicates of the seed.}
    \label{fig:scatter_per_seed_hiv}
\end{figure}

Figures~\ref{fig:scatter_per_seed_bace}--\ref{fig:scatter_per_seed_hiv} show the VIN--SimSeed trade-off
across test seeds. As a random-reference baseline, for each inactive test seed
we sample one molecule from the training set, either uniformly from all
molecules or from the active class. Mean Tanimoto similarity to the seed is
0.158/0.159 for BACE, 0.109/0.101 for BBBP, and 0.093/0.093 for HIV
(random molecule/random active, respectively). Thus, the similarities produced
by the generative methods remain well above the random-reference regime.

\FloatBarrier

\subsection{Failure-mode analysis}
\label{app:failure_modes}

\begin{figure}[!htbp]
    \centering
    \includegraphics[width=\textwidth]{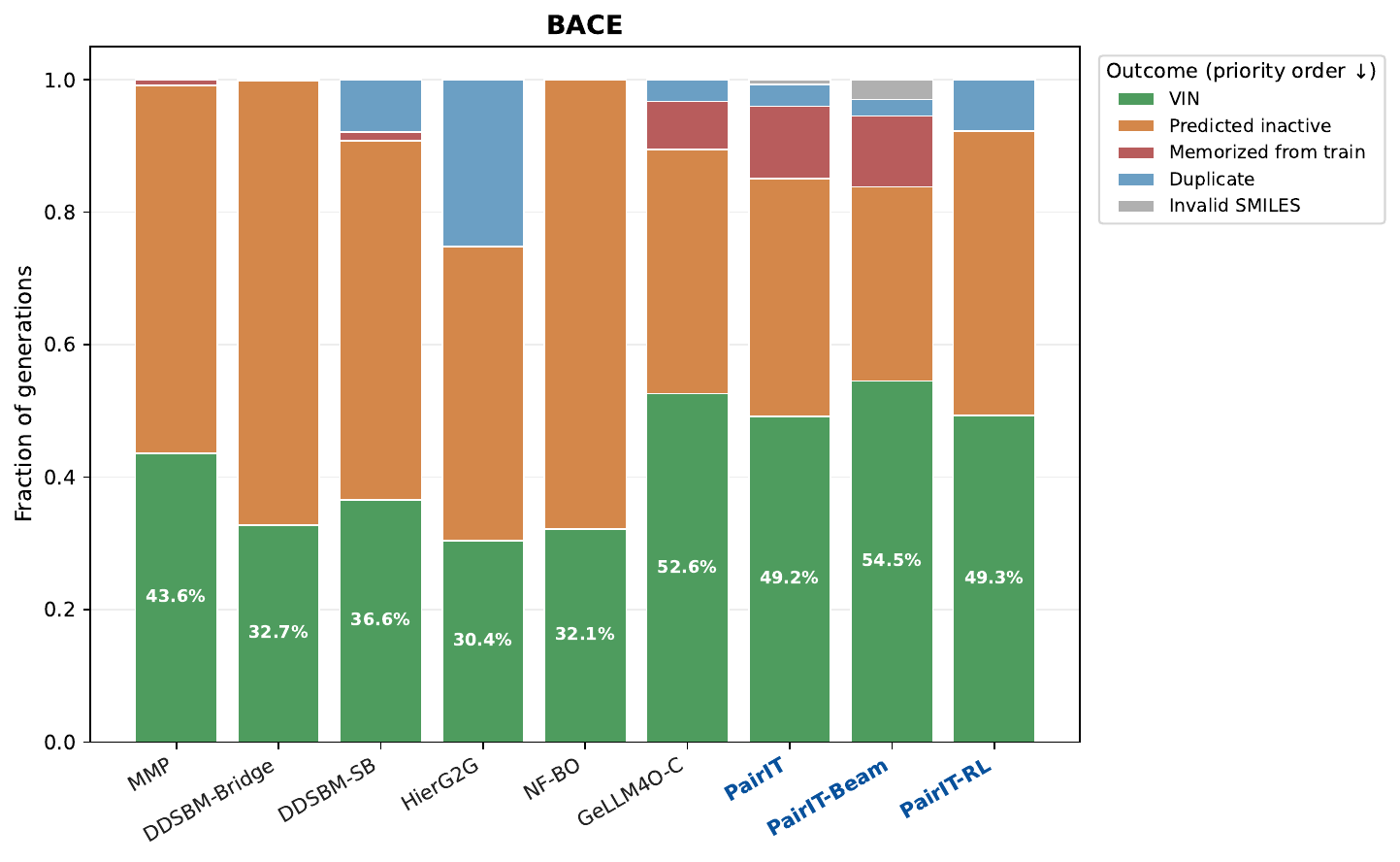}
    \caption{Failure-mode breakdown on MoleculeNet BACE. Bars show VIN success in green and the sources of error for generations not counted as VIN.}
    \label{fig:failure_bace}
\end{figure}

\begin{figure}[!htbp]
    \centering
    \includegraphics[width=\textwidth]{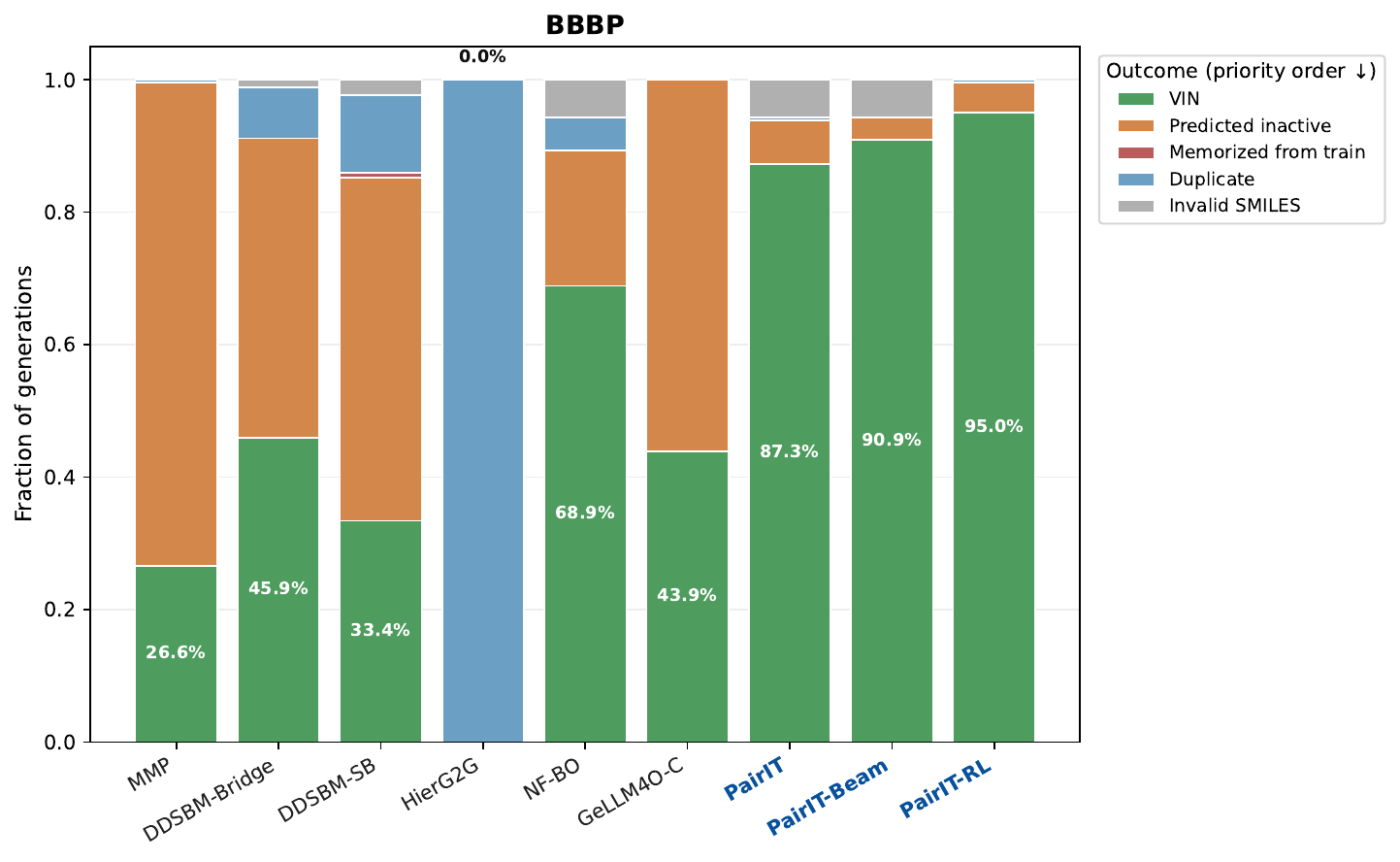}
\caption{Failure-mode breakdown on MoleculeNet BBBP. Bars show VIN success in green and the sources of error for generations not counted as VIN.}
    \label{fig:failure_bbbp}
\end{figure}

\begin{figure}[!htbp]
    \centering
    \includegraphics[width=\textwidth]{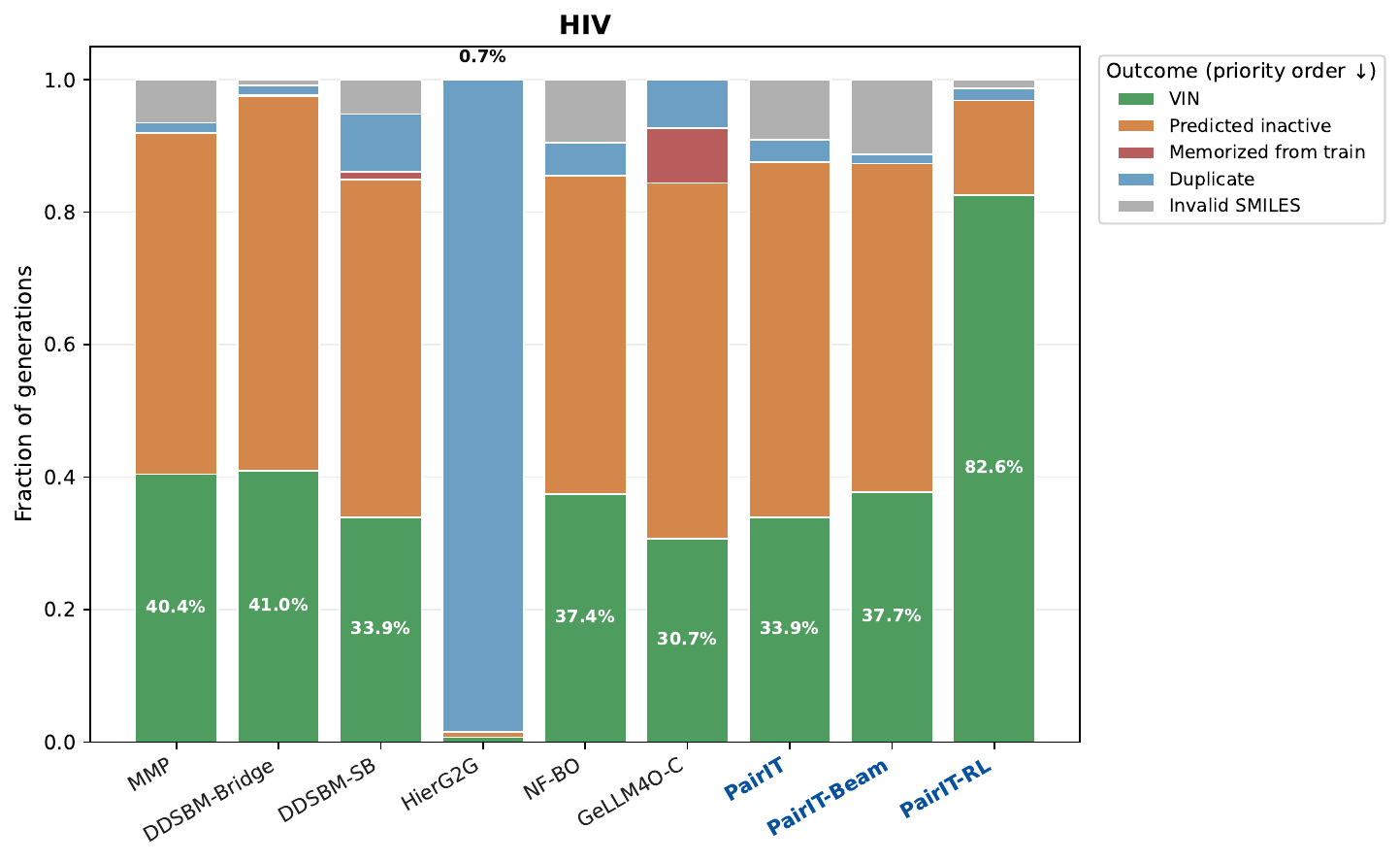}
    \caption{Failure-mode breakdown on MoleculeNet HIV. Bars show VIN success in green and the sources of error for generations not counted as VIN.}
    \label{fig:failure_hiv}
\end{figure}

Figures~\ref{fig:failure_bace}--\ref{fig:failure_hiv} show the sources of error for generated molecules that do not satisfy VIN. The green segment corresponds to VIN success; the remaining segments indicate why a retained generation was not counted as VIN. Categories are assigned using the priority order shown in the legend, so each generation is counted once.

\FloatBarrier

\subsection{Additional generation examples}
\label{app:more_generations}
\begin{figure}[!htbp]
    \centering
    \includegraphics[width=\textwidth]{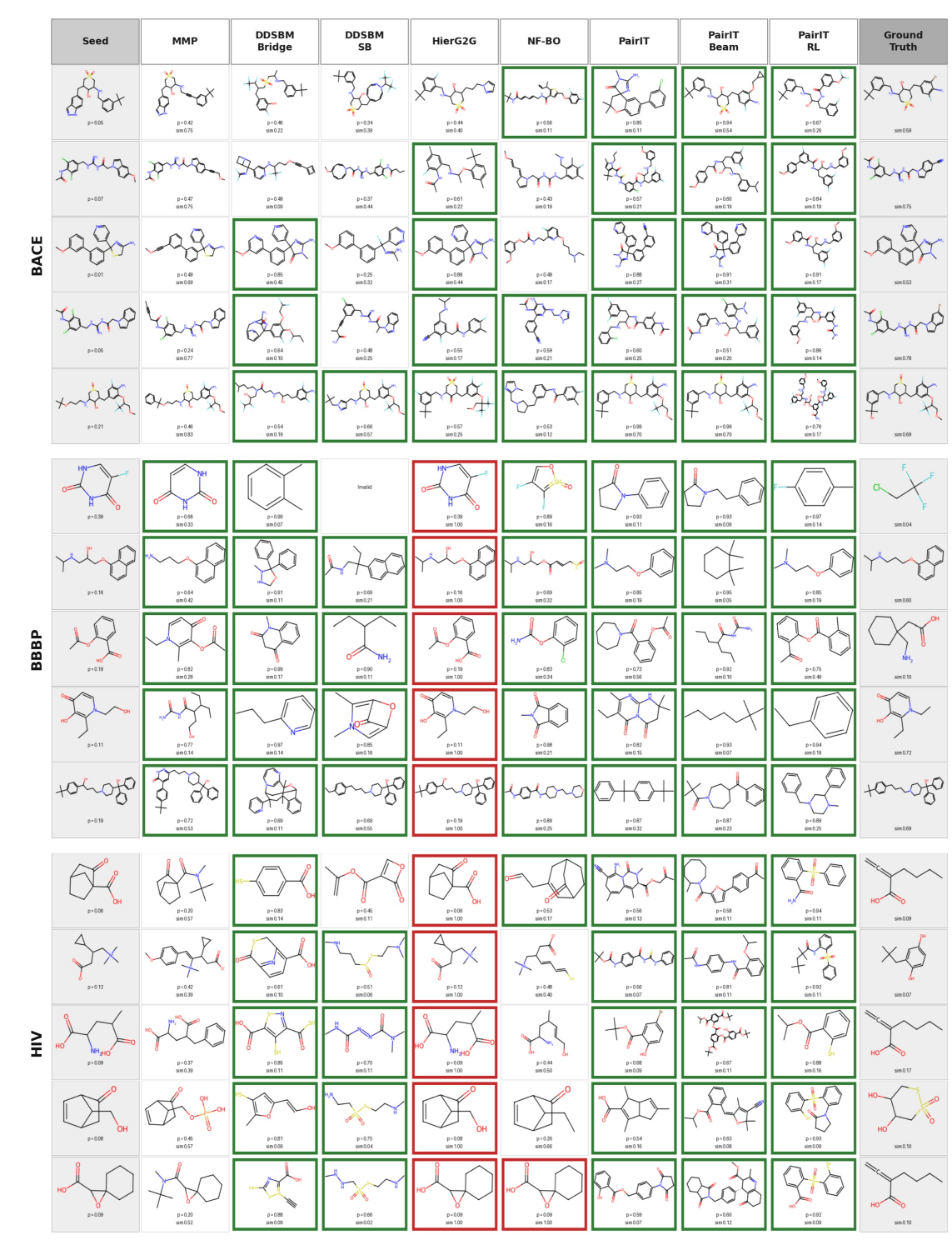}
        \caption{Additional example generations for MoleculeNet seeded optimization. Generations with predicted class probability greater than 0.5 are highlighted in green and seed duplicates in red.}
    \label{fig:seeded_baselines_app}
\end{figure}

Figure~\ref{fig:seeded_baselines_app} shows additional PairIT and PairIT-Beam generations on BACE, BBBP, and HIV.

\clearpage

\subsection{Additional low-data evaluation on hERG}
\label{app:herg}

We additionally evaluate on hERG from Therapeutics Data Commons
(TDC)~\citep{huang2021therapeutics,wang2016admet} as a distinct low-data toxicity
endpoint. hERG provides substantially sparser usable pair supervision than
BACE. MMPs are especially sparse, yielding only 41 blocker-to-non-blocker
training pairs, so we instead use local Tanimoto-neighbor pairs. Of 2,865
initial training pairs, 76 unique direction-improving pairs remain after
canonical-SMILES de-duplication and $\varepsilon$-filtering
(Table~\ref{tab:num_pairs}). This corresponds to approximately 0.18 usable
pairs per training molecule, similar to HIV and substantially below BACE
(0.83 pairs per molecule). Pairs are directed from blocker to non-blocker.

Table~\ref{tab:herg_results} reports the resulting seeded-generation
performance.

\begin{table}[t]
\centering
\footnotesize
\caption{Seeded molecular optimization on hERG. SimSeed measures Tanimoto
similarity to the input seed among VIN generations.}
\label{tab:herg_results}
\begin{tabular}{lccccc}
\toprule
Method & Valid \% & Improved \% & Novel \% & VIN \% & SimSeed \\
\midrule
MMP
& 91.76
& 10.77
& \textbf{100.00}
& 9.88
& \textbf{0.74} \\

DDSBM-Bridge
& \textbf{100.00}
& 19.76
& \textbf{100.00}
& 19.76
& 0.30 \\

DDSBM-SB
& 93.88
& \textbf{87.91}
& 97.74
& \textbf{80.71}
& 0.19 \\

HierG2G
& 99.06
& 81.67
& \underline{99.76}
& \textbf{80.71}
& 0.14 \\

NF-BO
& 98.59
& 22.30
& 99.52
& 21.88
& 0.47 \\

GeLLM4O-C
& \textbf{100.00}
& 29.18
& \textbf{100.00}
& 29.18
& \underline{0.60} \\

PairIT
& \textbf{100.00}
& 72.94
& \textbf{100.00}
& 72.94
& 0.20 \\

PairIT-Beam
& 94.12
& \underline{84.75}
& \textbf{100.00}
& \underline{79.76}
& 0.13 \\

PairIT-RL
& \underline{99.76}
& 40.19
& \underline{99.76}
& 40.00
& 0.39 \\
\bottomrule
\end{tabular}
\end{table}

Despite the sparse pair supervision, PairIT obtains 72.9\% VIN and
PairIT-Beam 79.8\%, with PairIT-Beam performing on par with the strongest
baselines. PairIT-RL does not improve over SFT on this task. hERG also
illustrates the importance of considering structural fidelity together with
property improvement: MMP and GeLLM4O-C remain substantially closer to the
input molecule but achieve much lower VIN, whereas the highest-VIN methods
have relatively low SimSeed. Thus, high VIN alone does not imply a localized
molecular improvement in this setting.

This issue is especially important for hERG, where improvement is defined as
blocker-to-non-blocker: methods can achieve high VIN while making large
structural departures from the seed. Indeed, HierG2G, DDSBM-SB, and
PairIT-Beam all achieve high VIN with SimSeed $<0.2$, illustrating that
property improvement and localized molecular editing need not coincide.

\FloatBarrier

\section{Pretraining and pair-construction ablations}

\subsection{Paired vs. single-molecule vs. no pretraining}
\label{app:pretrain_ablation}

To isolate the contribution of paired-molecule supervision during ZINC pretraining, we compare PairIT against single-molecule pretraining variants and against no pretraining at all.
  All variants share the same architecture (CLM-68M) and BACE fine-tuning pipeline; only the ZINC pretraining objective differs. Results are averaged across five replicate training seeds
  (four for Tanimoto $k{=}2$; two for the no-pretraining reference).

  \textit{Single-molecule (uniform)} refers to pretraining with single molecules from ZINC-250K (not pairs), sampled uniformly at random --- the model sees the full ZINC distribution.
  \textit{Single-molecule ($\varepsilon{=}1$ seeds)} refers to pretraining with single low-property molecules from ZINC-250K restricted to the subset that appears as a seed molecule in
  the $\varepsilon{=}1$ paired set (i.e.\ the first elements of the training pairs used for the paired variants). \textit{Tanimoto ($k$ nbr)} refers to pretraining with $k$-nearest
  neighbor pairs, where each seed is paired with its up to $k$ Tanimoto-nearest neighbors that satisfy $\Delta \log P > 1$; ``all nbr'' uses our default of up to 100 neighbors per seed.

 Table~\ref{tab:pretrain_pairs_vs_single} shows that pretraining on pairs of
similar molecules is central to PairIT; single-molecule pretraining from the
same dataset is less effective. Performance is not monotonic in the number of
neighbors, but more diverse ZINC-250K pairs modestly increase seed similarity
and expose the model to more unique molecules before early stopping,
motivating our default of up to 100 neighbors per seed.

Two failure modes are visible. Single-molecule (uniform) inflates Improved
(45.5\%) because the model emits seed-agnostic drug-like molecules that the
oracle scores as active, but SimSeed collapses to 0.20 and VIN drops to
40.0\%. Small-$k$ Tanimoto pairing (e.g., $k=1$) keeps generations close to
the seed (SimSeed 0.49), but produces enough near-duplicate canonicals that
fewer than five unique valid, improved, novel candidates remain to fill the
top-5 slots, giving 28.0\% VIN. Thus, pretraining with diverse pairs and early stopping is the key component.




\begin{table}[!htbp]
\centering
\footnotesize
\caption{Downstream BACE molecule-improvement with varying ZINC pretraining objectives. Valid, Improved, and Novel are computed over all 64 raw generations
per seed, while VIN and SimSeed are computed over the top-5 selected candidates
per seed.}
\label{tab:pretrain_pairs_vs_single}
\begin{tabular}{lccccc}
\toprule
Method & Valid \% & Improved \% & VIN \% & SimSeed & Novel \% \\
\midrule
\emph{No pretraining} & 40.9 & 25.3 & 20.0 & 0.42 & 36.3 \\
\midrule
Single-molecule (uniform)                 & 75.2 & \textbf{45.5} & 40.0 & 0.20 & 74.8 \\
Single-molecule ($\varepsilon{=}1$ seeds) & 69.3 & 37.0 & 44.0 & 0.21 & 70.0 \\
\midrule
Tanimoto (1 neighbor)   & 87.7 & 44.6 & 28.0 & 0.49 & 63.5 \\
Tanimoto (2 neighbor)   & 89.2 & 42.9 & 45.0 & 0.52 & 71.0 \\
Tanimoto (3 neighbor)   & 89.0 & 39.6 & 32.0 & 0.52 & 73.2 \\
Tanimoto (5 neighbor)   & 88.8 & 42.9 & 40.0 & 0.55 & 69.3 \\
Tanimoto (10 neighbor)  & 91.2 & 41.0 & 48.0 & 0.55 & 75.8 \\
Tanimoto (20 neighbor)  & \textbf{92.0} & 36.2 & 36.0 & \textbf{0.57} & \textbf{79.0} \\
Tanimoto (all neighbor) & 91.6 & 40.8 & \textbf{64.0} & 0.56 & 79.2 \\
\bottomrule
\end{tabular}
\end{table}

\begin{figure}[!htbp]
    \centering
    \includegraphics[width=\textwidth]{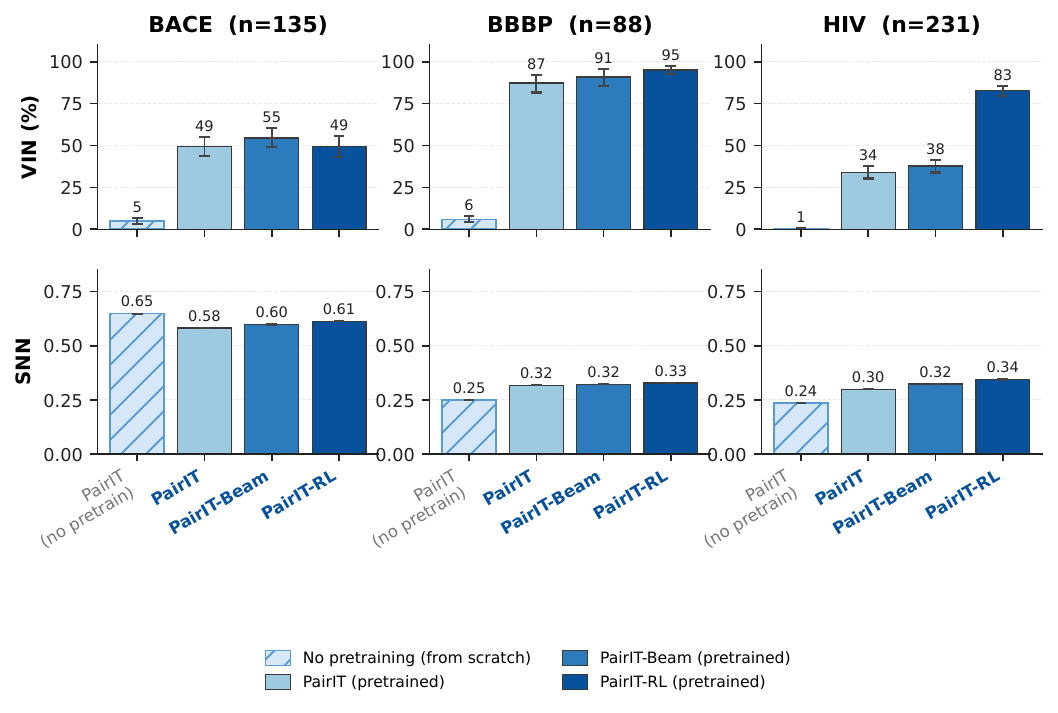}
    \caption{Effect of ZINC-250K pretraining on PairIT. Top row: VIN; bottom row: SNN computed over VIN molecules. The no-pretraining model uses the same architecture and MoleculeNet directed pairs as PairIT, but is trained from random initialization.}
    \label{fig:pairit_vs_no_pretraining}
\end{figure}

Figure~\ref{fig:pairit_vs_no_pretraining} compares the main pretrained PairIT variants with a PairIT model trained from random initialization using only the MoleculeNet directed pairs. The no-pretraining model uses the same 68M architecture and the same MoleculeNet fine-tuning data, but does not receive ZINC-250K auxiliary pretraining. Without pretraining, VIN collapses on all three MoleculeNet tasks, indicating that the small MoleculeNet pair sets alone are insufficient for learning valid, property-improving SMILES edits.

\begin{figure}[!htbp]
    \centering
    \includegraphics[width=\textwidth]{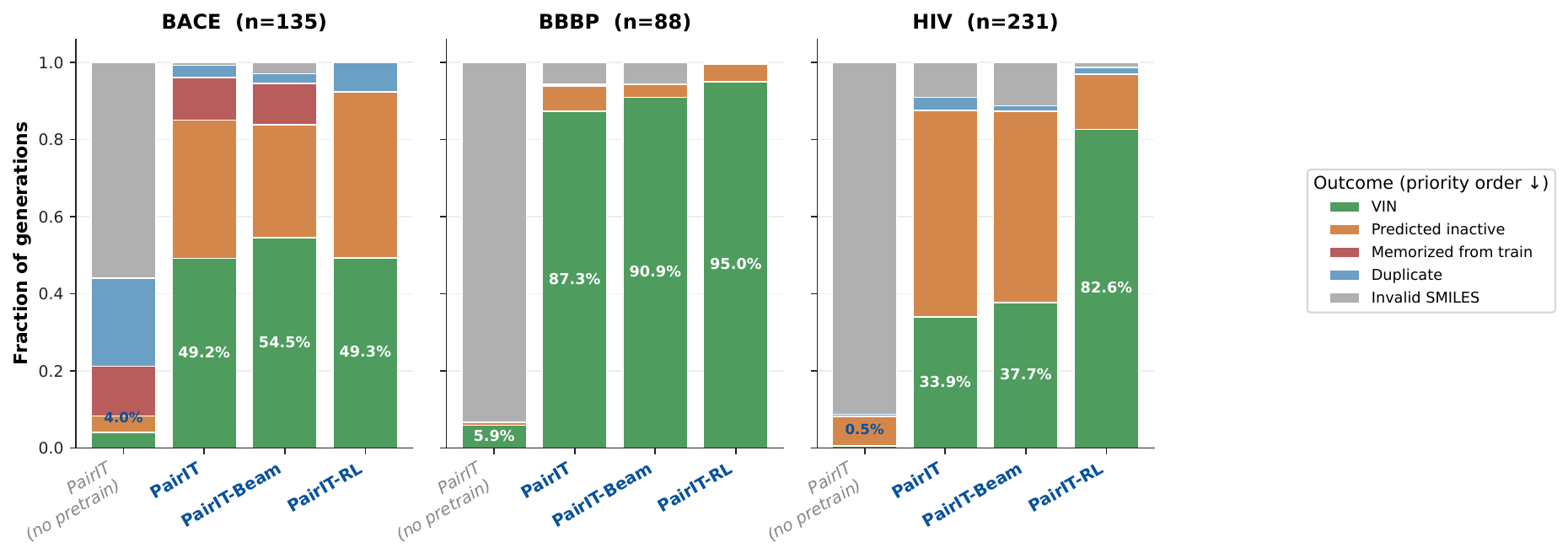}
    \caption{Failure-mode breakdown for the pretraining ablation. Without ZINC-250K pretraining, PairIT is dominated by invalid SMILES; pretrained PairIT variants rarely fail validity.}
    \label{fig:pretrain_failure_modes}
\end{figure}

Figure~\ref{fig:pretrain_failure_modes} shows that the main failure mode without pretraining is invalid SMILES, whereas the pretrained PairIT variants rarely fail validity. This suggests that pretraining on a moderate-size dataset is necessary for the model to learn the SMILES grammar.

\FloatBarrier

\subsection{Training dynamics and molecular coverage}
\label{app:training_dynamics}

\begin{figure}[!htbp]
    \centering
    \includegraphics[width=0.75\textwidth]{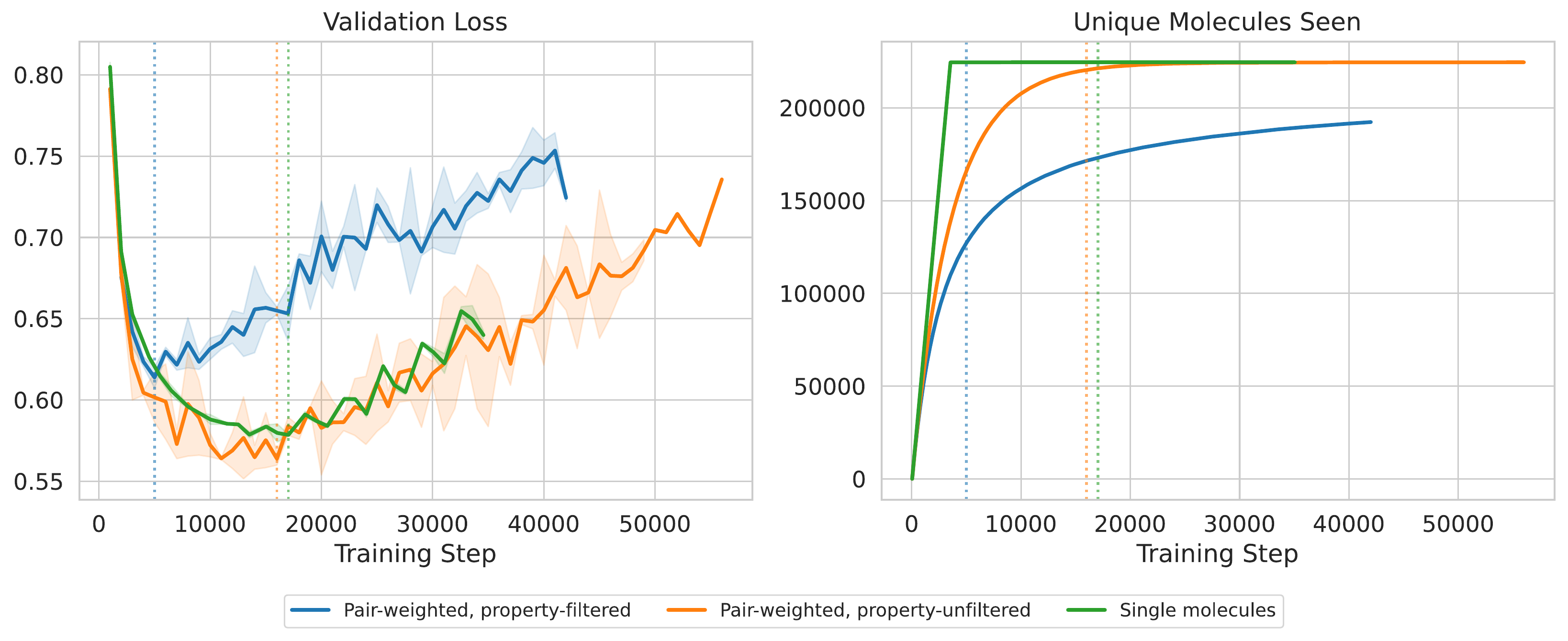}
    \caption{Variance in overfitting behavior, and the number of unique molecules seen at the time of overfitting, for different ZINC-based data distributions (pairs with and without property-based filtering, and unconditional single-molecule generation). Dotted lines indicate the point of early stopping, i.e. minimal validation loss.}
    \label{fig:overfitting-vs-unique}
\end{figure}

\FloatBarrier

\subsection{Validation loss and downstream performance}

\begin{figure}[!htbp]
    \centering
    \includegraphics[width=0.75\textwidth]{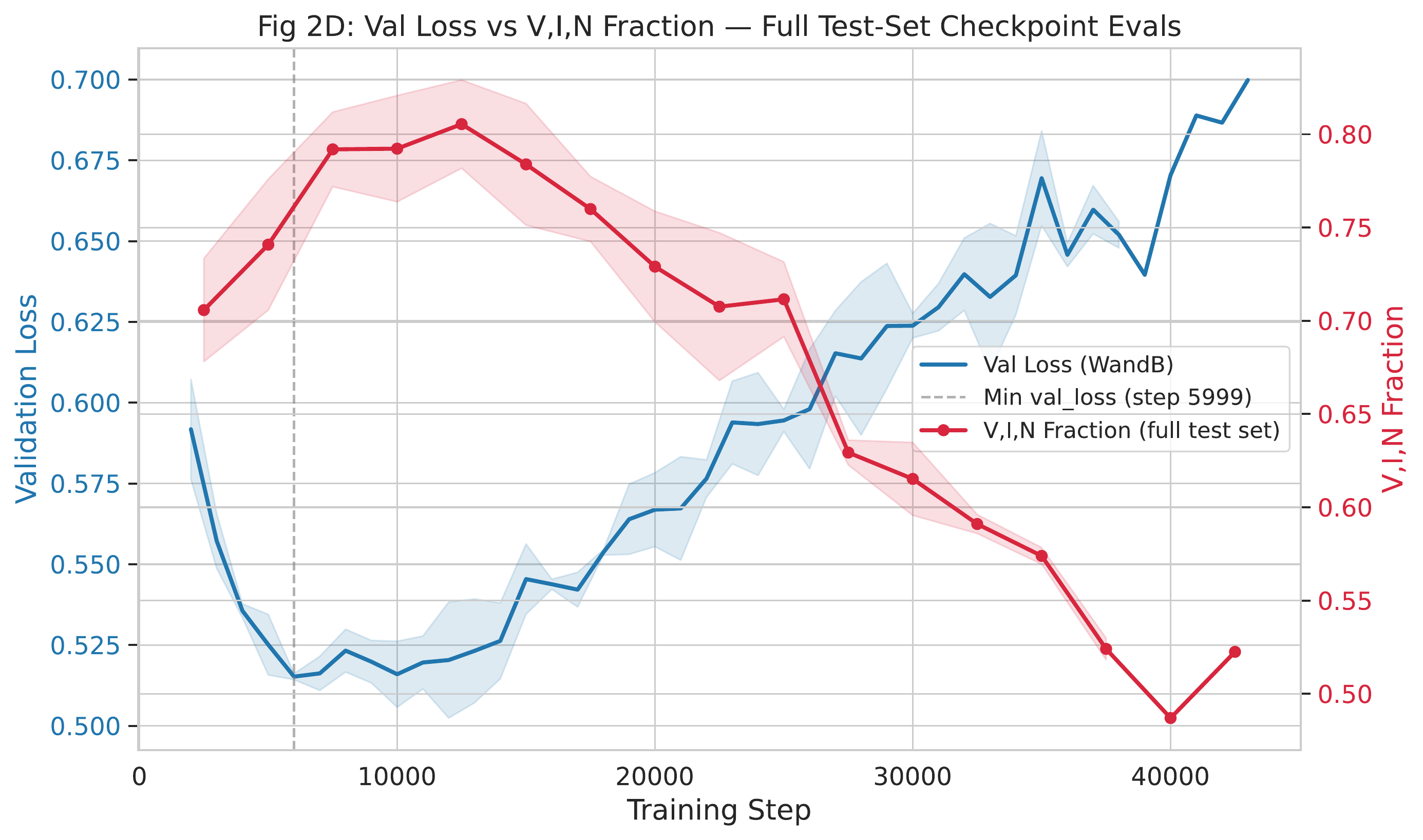}
    \caption{ZINC pretraining validation loss is well-correlated with chemical metrics, specifically its own generation rate of valid, logP-improved molecules, and novel molecules (VIN), justifying our early stopping criterion based on validation loss.}
    \label{fig:zinc-val-loss-and-zinc-vin}
\end{figure}

\begin{figure}
    \centering
    \includegraphics[width=0.75\textwidth]{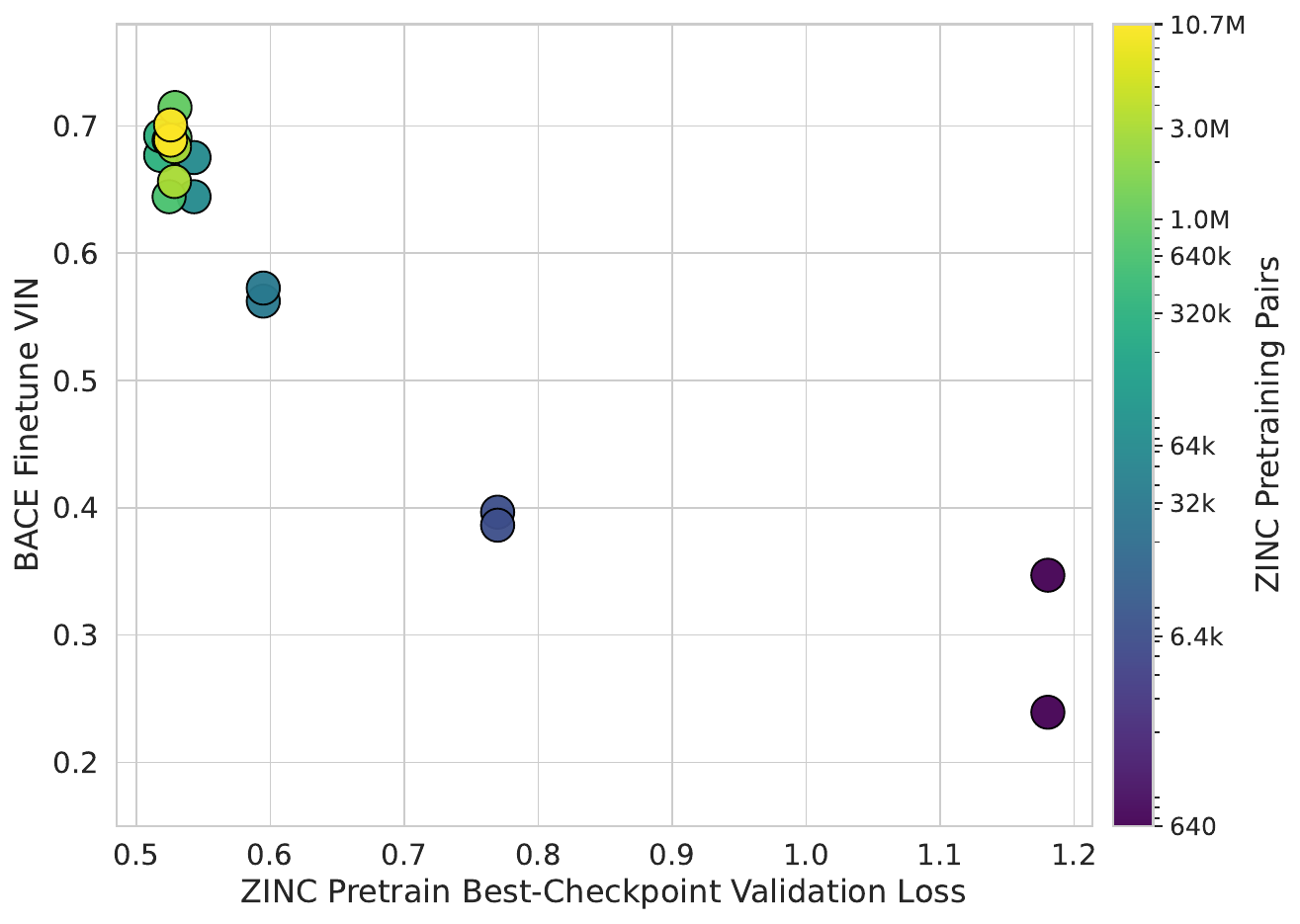}
    \caption{ZINC pretraining validation loss is well-correlated with downstream fine-tuned performance on BACE, further justifying our early stopping criterion based on validation loss. Here, each point is a separate checkpoint, pretrained on ZINC pair datasets of varying sizes and fine-tuned on the full BACE dataset.}
    \label{fig:zinc-val-loss-vs-bace-vin}
\end{figure}

\FloatBarrier

\subsection{Property filtering ablation}
We vary the property-filtering cutoff used to construct training pairs and find that more aggressive cutoffs induce correspondingly stronger property changes in the generated molecules, even without prompt conditioning.

\label{app:property_filtering}
\begin{figure}
    \centering
    \includegraphics[width=0.75\textwidth]{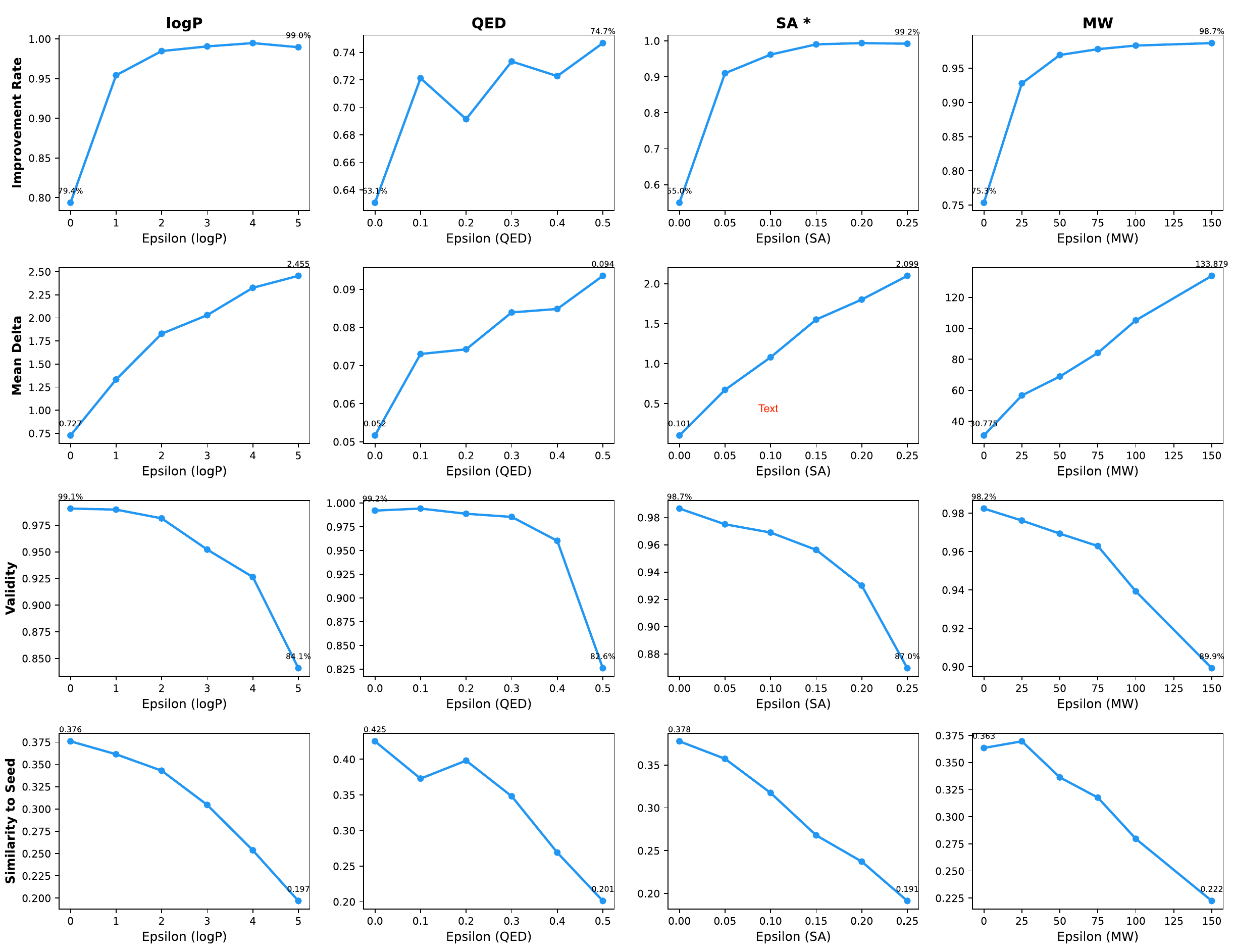}
    \caption{Filtering with more aggressive property cutoffs results in corresponding learned behavior. (This is in the basic setting without any prompting.)}
    \label{fig:data-filtering-to-behavior}
\end{figure}

\FloatBarrier

\subsection{ZINC-250K logP optimization}
\label{app:zinc_logp}

\begin{figure}[!htbp]
    \centering
    \includegraphics[width=\textwidth]{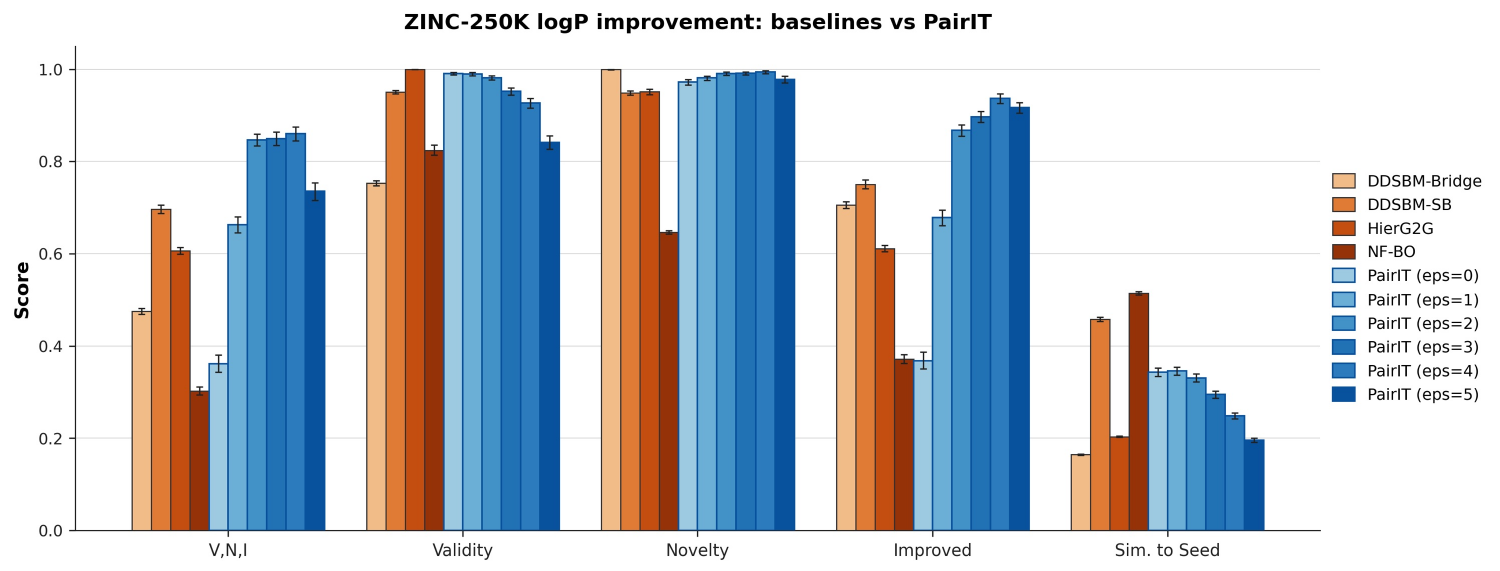}
    \caption{ZINC-250K logP optimization. Bars report joint valid--novel--improved success (V,N,I), validity, novelty, improvement, and SimSeed. PairIT variants sweep the pair-construction threshold $\varepsilon \in \{0,\ldots,5\}$; larger $\varepsilon$ corresponds to training on pairs with larger logP increases. Error bars are bootstrap 95\% confidence intervals over seeds.}
    \label{fig:zinc_logp_main}
\end{figure}

Figure~\ref{fig:zinc_logp_main} reports seeded logP optimization on ZINC-250K. PairIT variants differ by the pair-construction threshold $\varepsilon$, defined as the minimum logP increase between the seed and target molecules in a training pair. Increasing $\varepsilon$ makes the training pairs more strongly property-improving, which increases the Improved rate but decreases similarity to the seed. Overall, PairIT achieves the strongest joint valid--novel--improved performance at intermediate $\varepsilon$, while maintaining high validity and novelty.

\FloatBarrier

\subsection{Zero-shot logP generalization}
\label{app:zero_shot_logp}
\begin{figure}
    \centering
    \includegraphics[width=0.75\textwidth]{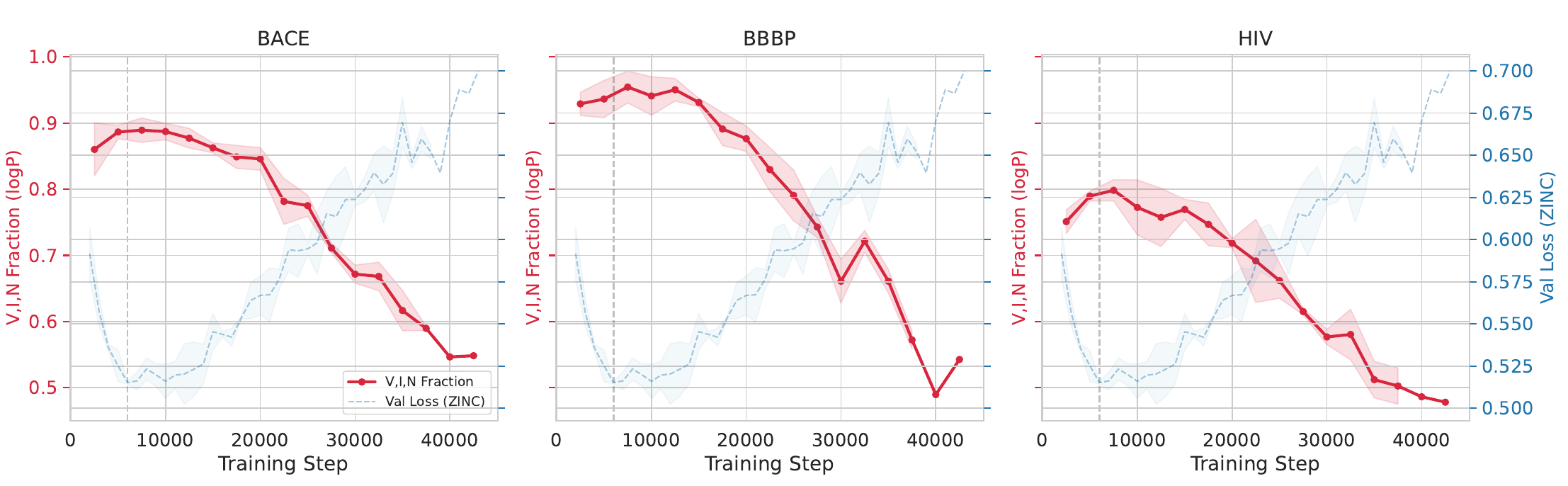}
    \caption{A model pretrained to optimize logP on ZINC transfers zero-shot to logP optimization on MoleculeNet. We track this ability over the course of training, and see that it correlates well with the pretraining validation loss.}
    \label{fig:zero-shot-logp-transfer}
\end{figure}

\FloatBarrier

\section{PairIT-RL}
\label{app:rl}

\subsection{Reward and training details}

\textbf{Reward Model} 
For each generation $g$ produced from seed $s$, the reward has a gated structure that assigns fixed penalties to degenerate outputs before scoring valid, non-trivial
  molecules:                                          
  
\[
R_{\mathrm{gated}}(g, s) =
\begin{cases}
-1.0 & \text{if } g \text{ is not a valid SMILES} \\
-0.5 & \text{if } g = s \text{ (exact copy of the seed)} \\
-0.5 & \text{if } g \in \mathcal{D}_{\mathrm{train}}
       \text{ (memorised from any training set)} \\
-0.5 & \text{if } \mathrm{Tan}(g, s) < \tau_{\mathrm{sim}}
       \text{ (below similarity gate)} \\
r_{\mathrm{obj}}(g,s) & \text{otherwise.}
\end{cases}
\]

\[
r_{\mathrm{obj}}(g,s)
=
\Delta_{\mathrm{property\ model}}(g, s)
+ \lambda_{\mathrm{act}}
  \mathbf{1}[p_{\mathrm{property\ model}}(g) > 0.5]
+ w_{\mathrm{sim}} \mathrm{Tan}(g, s).
\]

The first three gates enforce basic quality, assigning fixed penalties to invalid, copied, or memorized outputs rather than property-model rewards. The fourth gate (controlled by threshold $\tau_{\mathrm{sim}} \in \{0, 0.2\}$) optionally penalises generations that diverge too far from the seed before any positive reward is computed.

  For molecules that pass all gates, the reward combines three terms. $\Delta_{\mathrm{property model}}(g, s) = p_{\mathrm{property model}}(g) - p_{\mathrm{property model}}(s)$ is the change
  in predicted activity probability relative to the seed, rewarding improvement in the target property. $\lambda_{\mathrm{act}} \cdot \mathbf{1}[p_{\mathrm{property model}}(g)
  > 0.5]$ is a threshold bonus ($\lambda_{\mathrm{act}} = 2.0$) for generating molecules the property model classifies as active, providing a discrete incentive beyond the
  continuous $\Delta$ signal. $w_{\mathrm{sim}} \cdot \mathrm{Tan}(g, s)$ rewards structural similarity to the seed, where $w_{\mathrm{sim}}$ is the primary
  hyperparameter we sweep ($w_{\mathrm{sim}} \in \{1, 3, 5, 10\}$). This term serves both its obvious purpose — preserving structural fidelity — and an implicit one:
  keeping generations within the distribution where the activity property model was trained and can be trusted.

  For multi-property models, the total reward adds a prompt-adherence term:

  $$
  R(g, s) = R_{\mathrm{gated}}(g, s) + 2.0 \cdot R_{\mathrm{adhere}}(g, s)
  $$

  where $R_{\mathrm{adhere}}$ averages, over each prompted property, whether the generation changed that property in the prompted direction and by the prompted
  magnitude. The activity property is not prompted. It is driven entirely by the property model reward — so $R_{\mathrm{adhere}}$ covers only the nine RDKit descriptors (logP, QED, SA, MW, TPSA, HBD, HBA, rotatable bonds, fraction sp$^3$ carbons).

  The key point is that $w_{\mathrm{sim}}$ and $\tau_{\mathrm{sim}}$ are doing the work that KL regularisation alone is insufficient to do. The KL penalty
  (with $\beta_{\mathrm{KL}} = 0.01$) constrains the policy to stay near the SFT distribution in token space, but it doesn't directly constrain the output molecules to stay near the
  seed in chemical space. The Tanimoto terms provide that constraint explicitly, and the right operating point ($w_{\mathrm{sim}} = 5$, $\tau_{\mathrm{sim}} = 0.2$) is
  effectively a statement about how far from the seed the property model remains reliable.

\textbf{Training setup.}
PairIT-RL is initialized from the real-only SFT checkpoint and fine-tuned
with GRPO~\citep{shao2024deepseekmath} separately for each dataset. Common hyperparameters: temperature~1.0, $\beta_{\mathrm{KL}}=0.01$,
$\mathrm{lr}=5\times10^{-5}$, batch size~32, num\_gen~8, LoRA rank~16, 10 epochs.

\subsection{Similarity regularization and reward hacking.}
Without a similarity penalty ($w_{\rm sim}=0$), the policy collapses to
property model-fooling attractors at near-zero seed similarity, with BBBP VIN
dropping to~0\% and HIV VIN to~22\%.
Setting $w_{\rm sim}=5$ recovers seed similarity while maintaining high activity;
a hard penalty that rejects generations with $\mathrm{Tanimoto}<0.2$ works
similarly and trades a small amount of activity for greater similarity.

\subsection{Exploratory multi-property RL}
We additionally explored a multi-property RL variant initialized from the
10-property SFT model, using the activity reward above together with a
prompt-adherence term. We treat this as an exploratory extension rather than
part of the primary PairIT-RL evaluation; qualitative examples are shown in Appendix~\ref{app:rl_examples}.

\subsection{Representative Generations}
\label{app:rl_examples}
\subsubsection{Single-property RL}

\foreach \p in {1,...,2}{%
  \begin{figure}[H]
    \centering
    \includegraphics[page=\p, width=\linewidth]{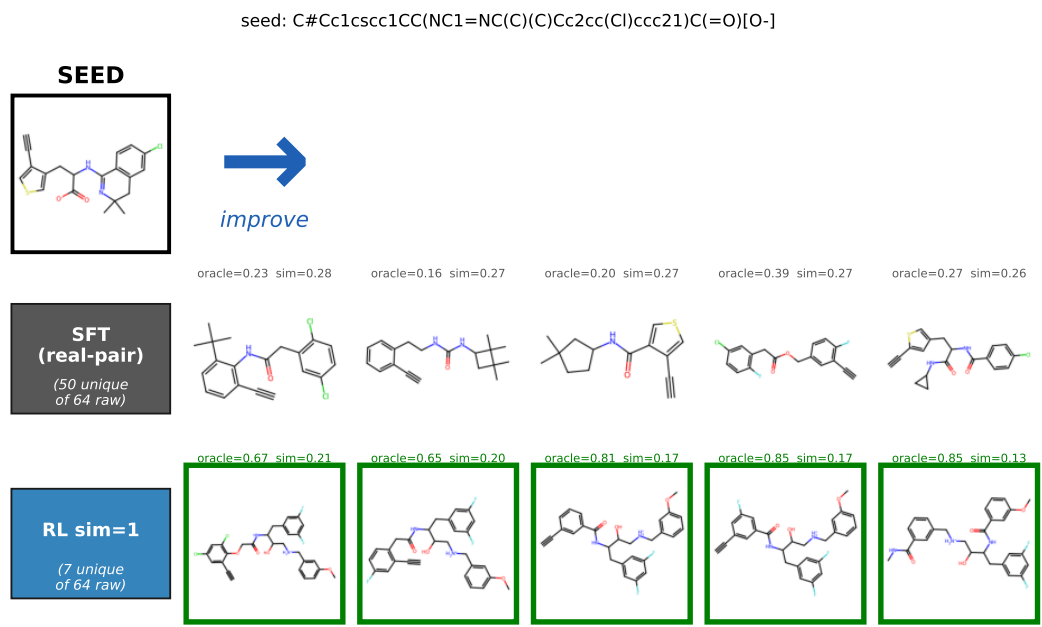}
    \caption{BACE single-property example, seed \p\ of 2.
    Top row: seed and ground-truth improved partner from the test pair.
    Each subsequent row: top-5 deduplicated generations for one method.
    Box color: green = V$\cdot$N$\cdot$I (valid $\wedge$ novel $\wedge$ improved);
    orange = V$\cdot$I but training-memorized; red = V$\cdot$I but seed-copy.}
  \end{figure}%
}

\foreach \p in {1,...,2}{%
  \begin{figure}[H]
    \centering
    \includegraphics[page=\p, width=\linewidth]{Figures/singleprop_examples_bbbp.pdf}
    \caption{BBBP single-property example, seed \p\ of 2.}
  \end{figure}%
}

\foreach \p in {1,...,2}{%
  \begin{figure}[H]
    \centering
    \includegraphics[page=\p, width=\linewidth]{Figures/singleprop_examples_hiv.pdf}
    \caption{HIV single-property example, seed \p\ of 2.}
  \end{figure}%
}

\subsubsection{Multi-property RL}

\foreach \p in {1,...,2}{%
  \begin{figure}[H]
    \centering
    \includegraphics[page=\p, width=\linewidth]{Figures/multiprop_examples_bace.pdf}
    \caption{BACE multi-property example, seed \p\ of 2.
    Methods compared: SFT, RL \texttt{sim=1}, RL \texttt{sim=3},
    RL \texttt{sim=5,gate=0.2}, RL \texttt{sim=10}.}
  \end{figure}%
}

\foreach \p in {1,...,2}{%
  \begin{figure}[H]
    \centering
    \includegraphics[page=\p, width=\linewidth]{Figures/multiprop_examples_bbbp.pdf}
    \caption{BBBP multi-property example, seed \p\ of 2.}
  \end{figure}%
}

\foreach \p in {1,...,2}{%
  \begin{figure}[H]
    \centering
    \includegraphics[page=\p, width=\linewidth]{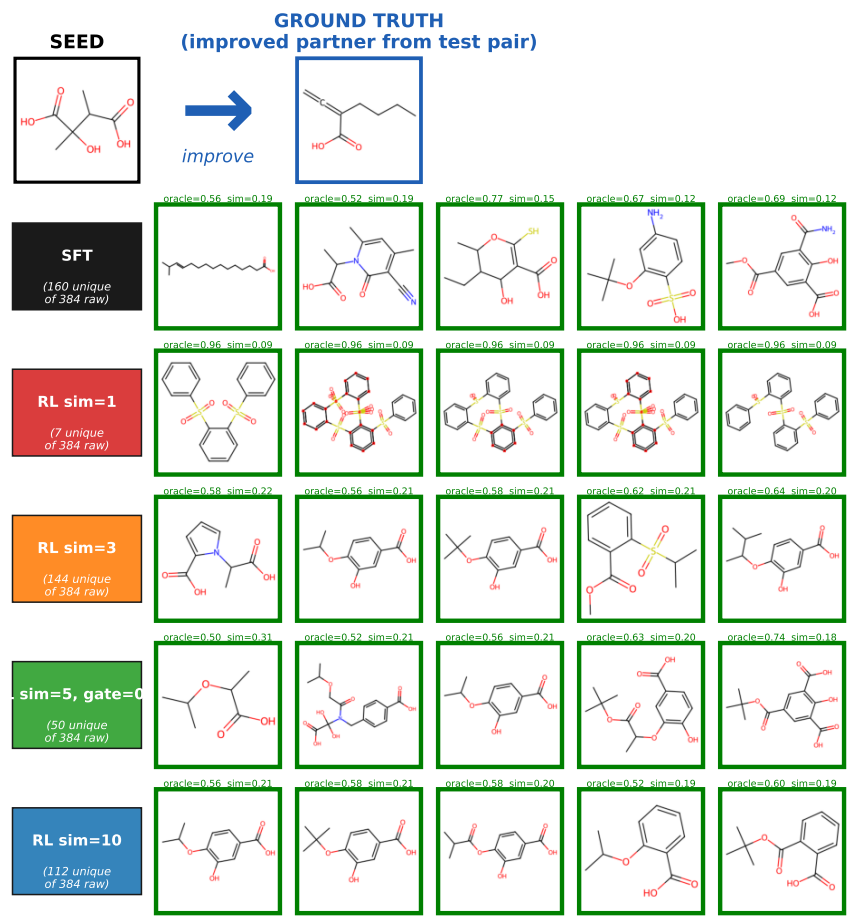}
    \caption{HIV multi-property example, seed \p\ of 2.}
  \end{figure}%
}

\clearpage
\section{Prompt-conditioned property control}
\label{app:prompting}

\subsection{Prompt tokens}
\label{sec:tokens}

At inference time, the prompt tokens and seed SMILES are provided as context,
and generation begins after \texttt{<sep>}. The prompt vocabulary consists of
a start-of-prompt marker \texttt{<pc>}, a property identifier such as
\texttt{<p\_logp>}, and a property-specific signed delta-bin token encoding the
desired change. The seed property value is not included in the prompt; direction
and approximate magnitude are encoded jointly by the sign and magnitude of the
delta bin.

\subsection{Single-property control}\label{app:single}
\begin{table}[t]
\centering
\caption{Single-property controllability on MoleculeNet after fine-tuning. VIN is the fraction of valid, novel generations whose property moves in the prompted direction. MSE is computed relative to the midpoint of the prompted delta bin.}
\label{tab:single_prop_control}
\begin{tabular}{llcc}
\toprule
Property & Dataset & VIN\% & MSE \\
\midrule
\multirow{3}{*}{QED ($\uparrow$ only)}
  & BACE & 52 & 0.041 \\
  & BBBP & 80 & 0.034 \\
  & HIV  & 77 & 0.034 \\
\midrule
\multirow{3}{*}{logP (bidirectional)}
  & BACE & 61 & 1.82 \\
  & BBBP & 69 & 4.09 \\
  & HIV  & 70 & 2.34 \\
\bottomrule
\end{tabular}
\end{table}

\begin{figure}[!htbp]
\centering
\includegraphics[width=\textwidth]{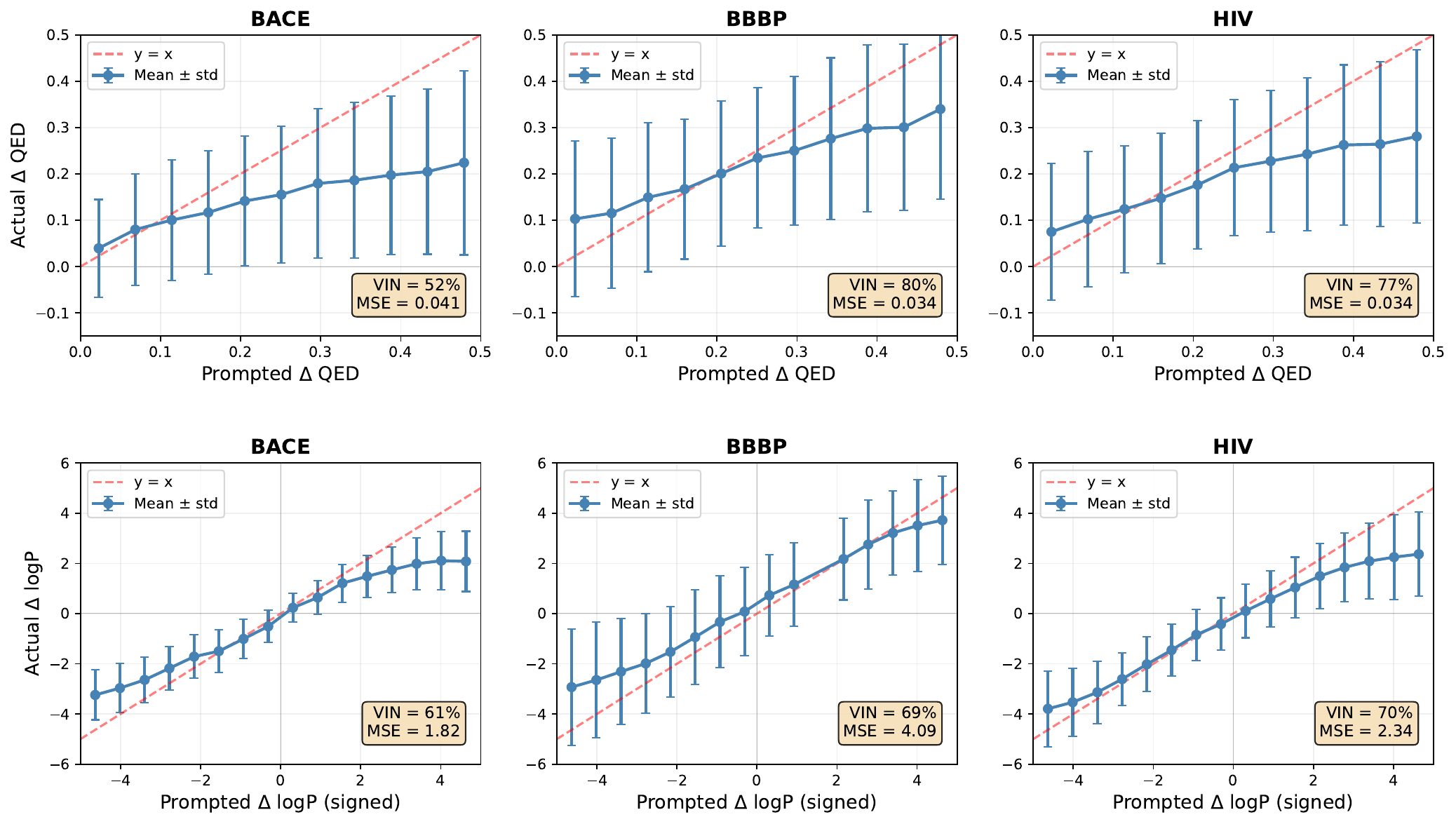}
\caption{Single-property controllability on MoleculeNet for both QED and logP. Top: QED, upward prompts over $[0,0.5]$. Bottom: logP, bidirectional prompts over $[-5,5]$. Error bars show standard deviation across 5 generations per seed per bin. Realized changes scale with prompted magnitude in both cases, indicating coarse magnitude control. The QED-only version of the top row appears in the main body as Figure~\ref{fig:single_prop_control}.}
\label{fig:single_prop_control_full}
\end{figure}

\FloatBarrier

\subsection{Two-property control on ZINC-250K}
\label{app:multi}
\begin{figure}[!htbp]
    \centering
    \includegraphics[width=\textwidth]{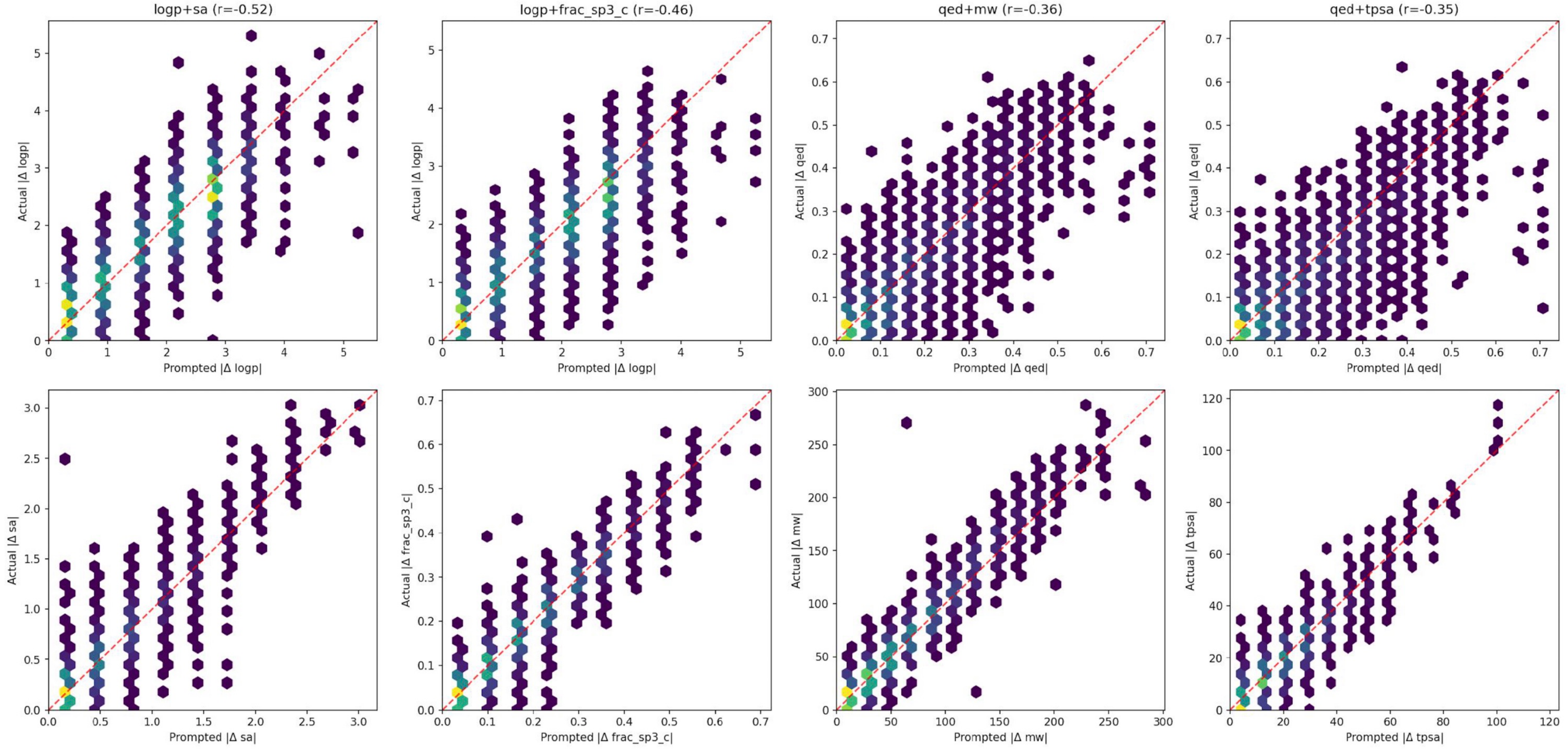}
    \caption{Prompted versus realized absolute property change under multi-property conditioning for four selected low-correlation property pairs from ZINC-250K. Each panel corresponds to one property within a pair and plots the prompted absolute delta on the x-axis against the realized absolute delta of the generated molecule on the y-axis; the dashed diagonal indicates perfect agreement. Across all pairs, larger requested edits produce larger realized changes, showing that approximate magnitude control is retained for both properties under simultaneous conditioning, although dispersion increases for large requested deltas and bounded properties such as QED show some saturation.}
    \label{fig:multi_property_delta}
\end{figure}

\FloatBarrier

\subsection{Two-property control on MoleculeNet}
\label{app:two_property_control}

The figures below show the full scatter plots for selected low-correlation two-property control and joint QED+logP control. In both cases, both properties are prompted simultaneously using signed delta bins derived from matched test pairs. Each point shows a valid generation; points are not filtered by whether the other property
also moves correctly.

We selected MW/HBD, TPSA/logP, and HBA/logP as weakly correlated descriptor
pairs using Pearson correlations of molecular property values across the
pooled ZINC training and validation molecules ($n=249{,}455$; $r=-0.005$,
$-0.162$, and $+0.056$, respectively), and used them for BACE, BBBP, and HIV,
respectively.

For each held-out matched test pair, we compute the requested property changes
as reference-target minus seed values and convert them to signed-delta-bin
tokens. Generation is conditioned only on the prompt tokens and seed SMILES;
the reference-target SMILES is not provided to the model. The two-property
evaluations use 51 BACE, 14 BBBP, and 46 HIV test pairs with inactive seeds and
active reference targets. We generate 64 outputs per pair with the same prompt,
without top-five selection. Results pool generated attempts across pairs;
prompts follow the observed pair distribution rather than a balanced bin sweep.

For each descriptor, VIN is the fraction of all generated attempts with a
nonzero raw requested change that are valid, novel relative to the task's
training molecules, and move in the requested direction. Combined VIN includes attempts with at least one applicable requested change and requires every applicable criterion
to succeed, together with validity and novelty. A zero requested descriptor
change imposes no success condition, and pairs with both descriptor changes
zero are excluded from Combined VIN. For bioactivity co-control, success means
an inactive seed yields a generation with train-only random-forest activity
probability greater than 0.5. The scatter plots show valid generations with a
nonzero request for the displayed property, without requiring success on the
other property. Their MSE pools squared errors against the raw matched-pair
change across valid generations with finite values and nonzero target changes;
novelty is not required. The per-pair medians are display summaries and do not
enter MSE.

For the randomized-prompt control, we independently sample each property's
signed-bin index uniformly from 00--15, with replacement, while keeping the
property identities and prompt-block order fixed. No separate direction token
is sampled. One reproducible randomized assignment is fixed per ordered
seed--target pair and shared by its 64 generations. The control uses the same
trained checkpoint as the correctly prompted arm. Both arms are scored against
the original requested changes.

\begin{figure}[!htbp]
\centering
\includegraphics[width=\textwidth]{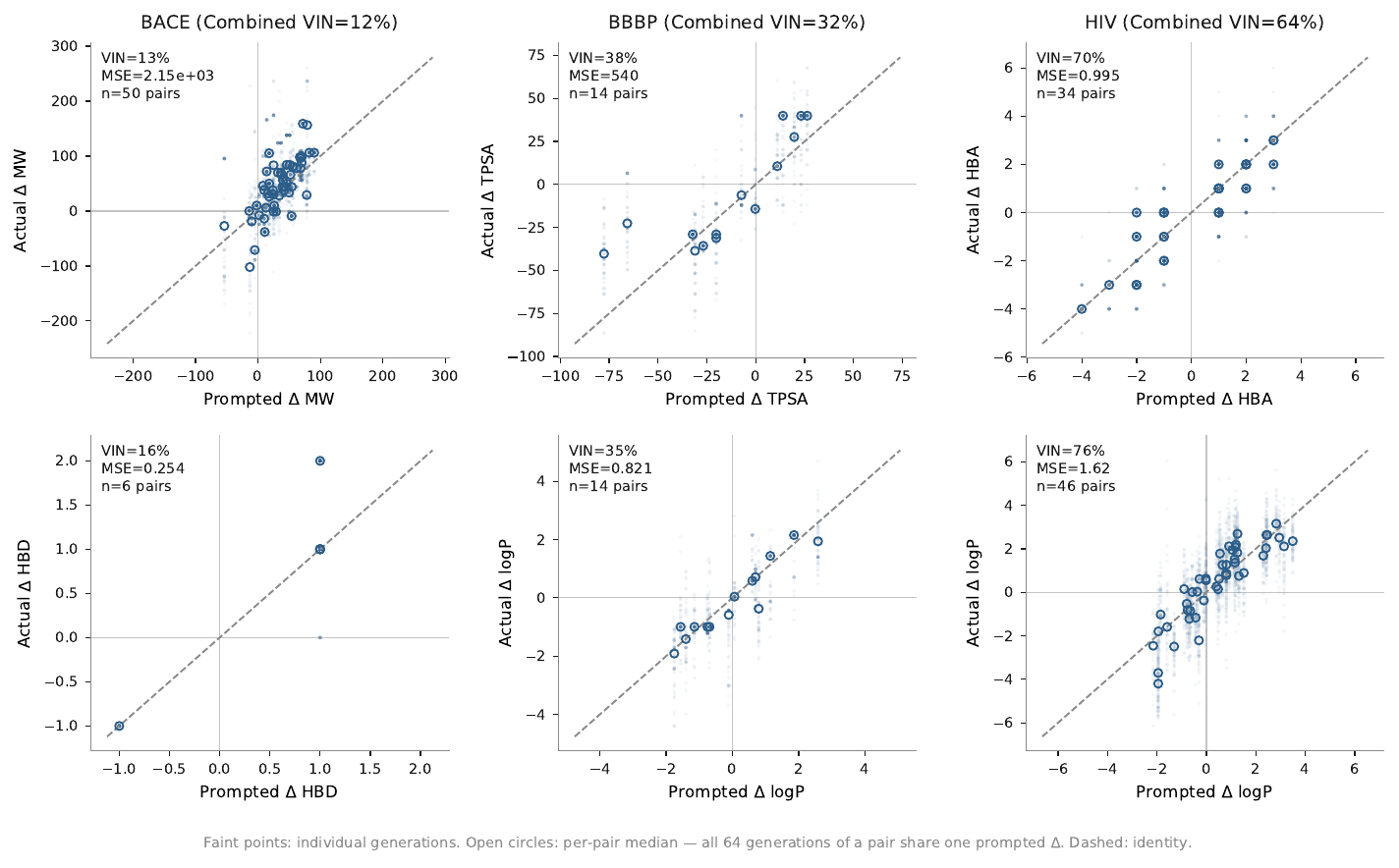}
\caption{Selected low-correlation two-property control on MoleculeNet. Each column is one
dataset; the top and bottom rows show the two prompted properties. Column titles
report combined VIN\% (valid, novel outputs satisfying every nonzero requested direction, divided by all attempts with at least one nonzero request).
The model achieves 12--64\% combined VIN, demonstrating independent control of
selected property pairs with low empirical correlation.}
\label{fig:uncorr_2prop}
\end{figure}

\begin{table}[htbp]
\centering
\caption{Two-property controllability for selected low-correlation property pairs on
MoleculeNet, using 64 raw generations per matched test pair without top-five
selection. Per-property VIN divides valid, novel, direction-matching outputs
by all attempts with a nonzero requested change for that property. Combined
VIN divides valid, novel outputs satisfying every nonzero requested direction
by all attempts with at least one nonzero request. Invalid and non-novel
attempts remain denominator failures; zero-delta properties impose no condition.}
\label{tab:vin_results}
\begin{tabular}{llccc}
\toprule
Dataset & Props & Combined VIN\% & Prop1 VIN\% & Prop2 VIN\% \\
\midrule
BACE & MW+HBD    & 12 & 13 & 16 \\
BBBP & TPSA+logP & 32 & 38 & 35 \\
HIV  & HBA+logP  & 64 & 70 & 76 \\
\bottomrule
\end{tabular}
\end{table}

\begin{figure}[!htbp]
\centering
\includegraphics[width=\textwidth]{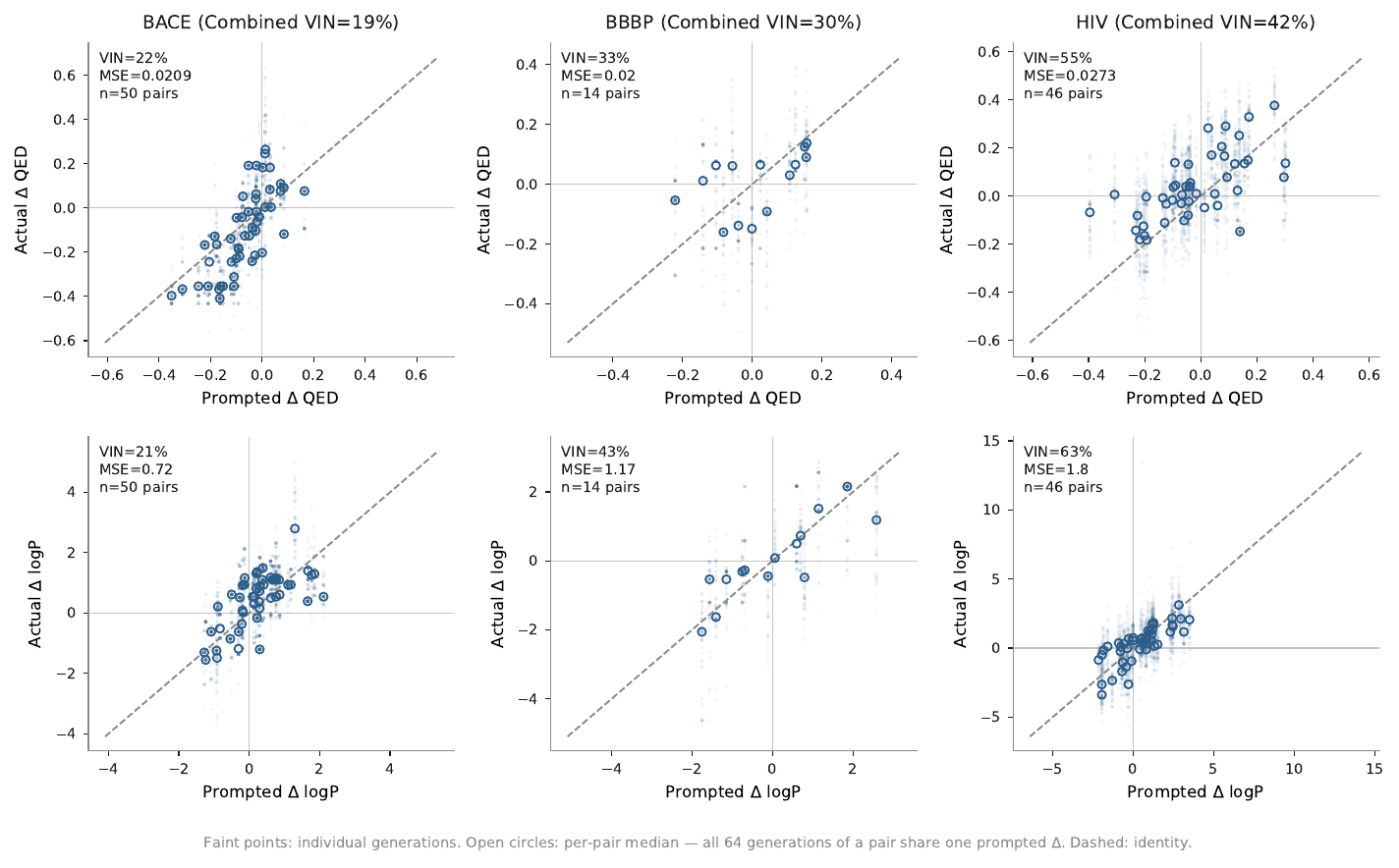}
\caption{Joint QED+logP control, with both properties prompted using natural
signed deltas from matched test pairs. The model achieves 19--42\% combined VIN
(valid, novel outputs satisfying every nonzero requested direction, divided by all attempts with at least one nonzero request). QED+logP is
a harder joint-control setting than uncorrelated descriptor pairs.}
\label{fig:qed_logp_joint}
\end{figure}

\begin{table}[htbp]
\centering
\caption{Two-property controllability for joint QED and logP control on
MoleculeNet,
using signed-bin requests from held-out matched pairs and 64 raw generations
per pair without top-five selection. Per-property VIN requires validity,
novelty, and the correct direction, using all attempts with a nonzero requested
change for that property as its denominator. Combined VIN
uses all attempts with at least one nonzero request and requires validity,
novelty, and agreement with every nonzero requested direction. Zero-delta
properties impose no condition.}
\label{tab:vin_qed_logp}
\begin{tabular}{lccc}
\toprule
Dataset & Combined VIN\% & QED VIN\% & logP VIN\% \\
\midrule
BACE & 19 & 22 & 21 \\
BBBP & 30 & 33 & 43 \\
HIV  & 42 & 55 & 63 \\
\bottomrule
\end{tabular}
\end{table}

\FloatBarrier

\subsection{Many-property control}
\label{app:multi_prop_full}

We train a single model on all nine computable RDKit properties simultaneously
(logP, QED, SA, MW, TPSA, HBD, HBA, rotatable bonds, and fraction
sp\(^3\) carbons) plus a tenth \emph{target} prompt field. During ZINC pretraining, the target prompt field mirrors a randomly chosen real property for each training example, without committing it to a fixed property during pretraining. Property dropout (rate \(=0.3\)) is applied during pretraining, randomly omitting properties from the prompt so the model can be conditioned on only a subset of properties at inference time.

At MoleculeNet fine-tuning, the target prompt field is assigned to the dataset-specific
bioactivity property (\texttt{Class} for BACE, \texttt{p\_np} for BBBP, and
\texttt{HIV\_active} for HIV). The target-slot change is the reference
target's label minus the seed's label, discretized with the activity bin range.
The model is fine-tuned with this reassigned target meaning and the
dataset-specific bin limits. Every retained training pair is used in both
orderings, with all signed changes recomputed after reversal, and all ten
property blocks remain included.

\begin{table}[htbp]
\centering
\caption{Full multi-property prompted model evaluated with all 10 properties on
MoleculeNet test pairs. Each property is prompted with the signed delta bin corresponding to the
ground-truth pair. Active\% is the fraction of valid, novel generations
predicted active (class~1) by the property model; property model~Up\% is the fraction for which
the property model score moves in the correct direction relative to the seed. SimSeed is the mean Tanimoto similarity to the input seed.}
\label{tab:hero_full}
\begin{tabular}{lcccccc}
\toprule
Dataset & VIN\% & Valid\% & Novel\% & Active\% & Property-model Up\% & SimSeed \\
\midrule
BACE & 21.7 & 95.8 & 40.9 & 53.0 & 60.4 & 0.491 \\
BBBP & 83.0 & 99.8 & 91.6 & 90.6 & 72.2 & 0.290 \\
HIV  & 12.3 & 92.3 & 89.4 & 13.8 & 47.9 & 0.162 \\
\bottomrule
\end{tabular}
\end{table}
For the ten-property model's direction-adherence results in, we use
only held-out pairs with an inactive seed and an activity-label request of
$0\!\to\!1$. For each of the nine non-activity descriptors, we calculate the
fraction of valid generations with a nonzero raw requested change whose
realized change has the same sign. We pool eligible valid generations
within each descriptor and then take the unweighted mean of the nine rates.
These are raw generations without top-five selection.

Full multi-property context substantially helps BBBP, where RDKit descriptors
are highly correlated with bioactivity (e.g., TPSA: \(r=-0.50\), QED:
\(r=+0.48\)). 

\begin{table}[htbp]
\centering
\caption{Full multi-property prompted model evaluated with the target bioactivity
property plus logP, SA, and QED (three-property subset) on MoleculeNet test
pairs.}
\label{tab:hero_subset}
\begin{tabular}{lcccccc}
\toprule
Dataset & VIN\% & Valid\% & Novel\% & Active\% & property model Up\% & SimSeed \\
\midrule
BACE & 25.5 & 95.8 & 61.3 & 41.4 & 67.1 & 0.401 \\
BBBP & 79.3 & 90.0 & 90.0 & 88.1 & 95.2 & 0.818 \\
HIV  & 19.8 & 84.6 & 84.6 & 23.4 & 75.5 & 0.841 \\
\bottomrule
\end{tabular}
\end{table}

The logP+SA+QED subset yields 25.5\% VIN on BACE, 79.3\% on BBBP,
and 19.8\% on HIV, compared with 21.7\%, 83.0\%, and 12.3\%,
respectively, under the full 10-property prompt. Thus, reducing the prompt
to this subset improves VIN on BACE and HIV while modestly reducing it on
BBBP, consistent with the broader property-subset sensitivity in
Table~\ref{tab:9c3_summary}.

\begin{table}[htbp]
\centering
\caption{Property-subset sensitivity for the full multi-property prompted model.
A single trained model is evaluated across all \(\binom{9}{3}=84\) combinations
of three RDKit properties prompted alongside the target bioactivity property
(84 evaluation runs per dataset, not 84 trained models).}
\label{tab:9c3_summary}
\begin{tabular}{lcccc}
\toprule
Dataset & Mean VIN\% & Best VIN\% & Worst VIN\% & Spread (pp) \\
\midrule
BACE & \(25.1 \pm 2.2\) & 30.4 & 21.1 & 9.3 \\
BBBP & \(72.0 \pm 6.3\) & 85.0 & 57.1 & 27.9 \\
HIV  & \(13.0 \pm 3.8\) & 25.0 & 5.5  & 19.5 \\
\bottomrule
\end{tabular}
\end{table}

Property-subset selection has a significant impact, especially on BBBP
(27.9~pp spread) and HIV (19.5~pp spread). The best HIV combination (logP +
QED + TPSA, 25.0\%) is \(4.5\times\) better than the worst (QED + rotatable
bonds + fraction sp\(^3\) carbons, 5.5\%). Consistently poor subsets contain
rotatable bonds and fraction sp\(^3\) carbons across all datasets; consistently
strong subsets contain logP and/or SA for BACE and HIV, and SA, HBA, and TPSA
for BBBP, reflecting the high correlation of these descriptors with blood--brain
barrier permeability.

\begin{table}[htbp]
\centering
\caption{Mean property delta (generated minus seed) for the full multi-property
prompted model under 10-property prompting.}
\label{tab:hero_deltas}
\begin{tabular}{lrrr}
\toprule
Property & BACE & BBBP & HIV \\
\midrule
logP           & +0.022 & -0.037 & +0.709 \\
QED            & -0.032 & +0.018 & +0.033 \\
SA             & +0.126 & -0.132 & -0.595 \\
MW             & +17.78 & +3.11 & +18.56 \\
TPSA           & +6.20 & -7.09 & -15.51 \\
HBD            & -0.041 & -0.354 & -0.719 \\
HBA            & +0.423 & -0.921 & -0.125 \\
Rot. bonds     & +0.661 & +0.312 & +0.306 \\
Frac. sp\(^3\) & +0.057 & +0.200 & -0.006 \\
\bottomrule
\end{tabular}
\end{table}

BBBP generations systematically decrease TPSA (\(-7.09\)) and HBD
  (\(-0.354\)), consistent with the strong negative correlation between these
properties and BBB permeability (TPSA: \(r=-0.50\), HBD: \(r=-0.47\)). The
model appears to learn the structure--activity relationship through
multi-property prompting.

\FloatBarrier

\end{document}